\documentclass{article} 
\usepackage{iclr2023_conference,times}

\usepackage{amsmath,amsfonts,bm}

\def\eqref#1{equation~\ref{#1}}

\def\1{\bm{1}}

\DeclareMathAlphabet{\mathsfit}{\encodingdefault}{\sfdefault}{m}{sl}
\SetMathAlphabet{\mathsfit}{bold}{\encodingdefault}{\sfdefault}{bx}{n}

\DeclareMathOperator*{\argmax}{arg\,max}

\usepackage[utf8]{inputenc} 
\usepackage[T1]{fontenc}    
\usepackage{hyperref}       
\usepackage{url}            
\usepackage{booktabs}       
\usepackage{amsfonts}       
\usepackage{nicefrac}       
\usepackage{microtype}      
\usepackage{xcolor}         

\usepackage[linesnumbered, ruled]{algorithm2e}
\usepackage{amsmath}
\usepackage{amssymb}
\usepackage{bm}
\usepackage{bbm}
\usepackage[scaled=0.9]{beramono}
\usepackage{chngcntr}
\usepackage{dsfont}
\usepackage{glossaries}
\usepackage{graphicx}
\usepackage{multirow}
\usepackage[super]{nth}
\usepackage{siunitx}
\usepackage{subcaption}

\usepackage[outdir=./figures/]{epstopdf}

\SetAlCapSkip{1em}
\SetKwInput{KwInput}{Input}
\SetKwInput{KwOutput}{Output}

\SetCommentSty{mycommfont}

\usepackage[capitalize]{cleveref}

\newcommand{\eg}{e.g., }
\newcommand{\ie}{i.e., }
\newcommand{\etc}{etc.}

\newcommand{\cf}{cf.\ }
\newcommand{\wrt}{w.r.t.\ }
\newcommand{\iid}{i.i.d.\ }
\newcommand{\versus}{vs.\ }

\newcommand{\binaryaddition}{\texttt{Binary Addition}}

\newcommand{\duplicatestring}{\texttt{Duplicate String}}
\newcommand{\evenpairs}{\texttt{Even Pairs}}
\newcommand{\modulararithmeticbrackets}{\texttt{Modular Arithmetic}}
\newcommand{\modulararithmeticsimple}{\texttt{Modular Arithmetic (Simple)}}
\newcommand{\oddsfirst}{\texttt{Odds First}}
\newcommand{\paritycheck}{\texttt{Parity Check}}
\newcommand{\reversestring}{\texttt{Reverse String}}
\newcommand{\stackmanipulation}{\texttt{Stack Manipulation}}
\newcommand{\solveequation}{\texttt{Solve Equation}}
\newcommand{\computesqrt}{\texttt{Compute Sqrt}}
\newcommand{\missingduplicate}{\texttt{Missing Duplicate}}
\newcommand{\cyclenavigation}{\texttt{Cycle Navigation}}
\newcommand{\binarymultiplication}{\texttt{Binary Multiplication}}
\newcommand{\bucketsort}{\texttt{Bucket Sort}}

\newcommand{\push}{\textsc{push}}
\newcommand{\pusha}{\textsc{push $a$}}
\newcommand{\pushb}{\textsc{push $b$}}
\newcommand{\pop}{\textsc{pop}}
\newcommand{\noop}{\textsc{no-op}}
\newcommand{\writeleft}{\textsc{write-left}}
\newcommand{\writeright}{\textsc{write-right}}
\newcommand{\writestay}{\textsc{write-stay}}
\newcommand{\jumpleft}{\textsc{jump-left}}
\newcommand{\jumpright}{\textsc{jump-right}}

\newcommand{\parameters}{\theta}
\newcommand{\model}{p_\parameters}

\newacronym{regular}{R}{regular}
\newacronym{cf}{CF}{context-free}
\newacronym{cs}{CS}{context-sensitive}
\newacronym{re}{RE}{recursively enumerable}
\newacronym{dcf}{DCF}{deterministic context-free}
\newacronym{ndcf}{NDCF}{nondeterministic context-free}
\newacronym{fsa}{FSA}{finite-state automaton}
\newacronym{pda}{PDA}{push-down automaton}
\newacronym{npda}{NPDA}{nondeterministic push-down automaton}
\newacronym{lba}{LBA}{linear bounded automaton}
\newacronym{tm}{TM}{Turing machine}
\newacronym{utm}{UTM}{universal Turing machine}

\title{Neural Networks and the Chomsky Hierarchy}

\author{Gr{\'{e}}goire Del\'etang$^{*1}$
Anian Ruoss$^{*1}$
Jordi Grau-Moya$^1$
Tim Genewein$^1$
Li Kevin Wenliang$^1$
\And 
Elliot Catt$^1$
Chris Cundy$^{\dagger2}$
Marcus Hutter$^1$
Shane Legg$^1$
Joel Veness$^1$
Pedro A.\ Ortega$^\dagger$}

\iclrfinalcopy

\begin{document}

\maketitle

\def\thefootnote{*}\footnotetext{Equal contribution. Correspondence to \{gdelt, anianr\}@deepmind.com.}\def\thefootnote{\arabic{footnote}}
\def\thefootnote{1}\footnotetext{DeepMind. $^2$Stanford University. $^\dagger$Work performed while the author was at DeepMind.}\def\thefootnote{\arabic{footnote}}

\begin{abstract}
Reliable generalization lies at the heart of safe ML and AI.
However, understanding when and how neural networks generalize remains one of the most important unsolved problems in the field.
In this work, we conduct an extensive empirical study (\num{20910} models, 15 tasks) to investigate whether insights from the theory of computation can predict the limits of neural network generalization in practice.
We demonstrate that grouping tasks according to the Chomsky hierarchy allows us to forecast whether certain architectures will be able to generalize to out-of-distribution inputs.
This includes negative results where even extensive amounts of data and training time never lead to any non-trivial generalization, despite models having sufficient capacity to fit the training data perfectly.
Our results show that, for our subset of tasks, RNNs and Transformers fail to generalize on non-regular tasks, LSTMs can solve regular and counter-language tasks, and only networks augmented with structured memory (such as a stack or memory tape) can successfully generalize on context-free and context-sensitive tasks.
\end{abstract}
\section{Introduction}
\label{sec:introduction}

Statistical learning theory is the most widely used theory of generalization in practical machine learning, justifying empirical risk minimization and estimating the generalization error via a test set~\citep{vapnik1998statistical}.
However, its central assumption that training and test data are independent and identically distributed (i.i.d.) is violated for many problems of interest (distribution shifts, continual learning, \etc).
An example of such a non-\iid setting is testing generalization on sequence prediction problems, where an agent is trained with sequences of length $\ell \le N$  and tested with arbitrarily longer sequences $\ell \gg N$.
This problem is of particular importance since it subsumes all computable problems~\citep{dawid1984present, rich2007automata, sipser1997introduction, solomonoff2009algorithmic, solomonoff2010algorithmic}.
Central to sequence prediction is \emph{inductive inference}, which consists of deriving a general rule from a finite set of concrete instances and using this rule to make predictions.
For example, in \emph{program induction}~\citep{goldberg1989genetic, gomez2008accelerated, holland992adaptation, liang2013learning, nording1997evolutionary, solomonoff1964formalI, solomonoff1964formalII, wineberg1994representation}, the goal is to obtain a model that correctly identifies the underlying data-generating process given examples of input-output sequences.
Then, if the model is correct, it can produce results in accordance with the generative process for previously unseen input sequences.

The key challenge of inductive inference (as opposed to deduction) is that it does not allow selecting one hypothesis with certainty among the ones that fit the data.
For instance, the sequence $2, 4, 6, 8$ has infinitely many possible continuations.
Thus, any principle that selects a particular continuation requires additional assumptions that are independent of the data, \ie inductive biases~\citep{mitchell1980need}.
In machine learning, the network architecture, training mechanisms (\eg gradient descent), and initial distributions over parameters all generate their corresponding inductive biases.
This has led to a vast number of approaches for designing inductive biases via architectural and training protocol changes (see \citet{battaglia2018relational} for an overview).
However, the problem is that stronger inductive biases generally come at the cost of decreasing the universality of a model, and thus finding a good balance between the two is one of the biggest challenges in the contemporary literature.

\begin{figure}
    \begin{center}
        \includegraphics[width=0.75\columnwidth]{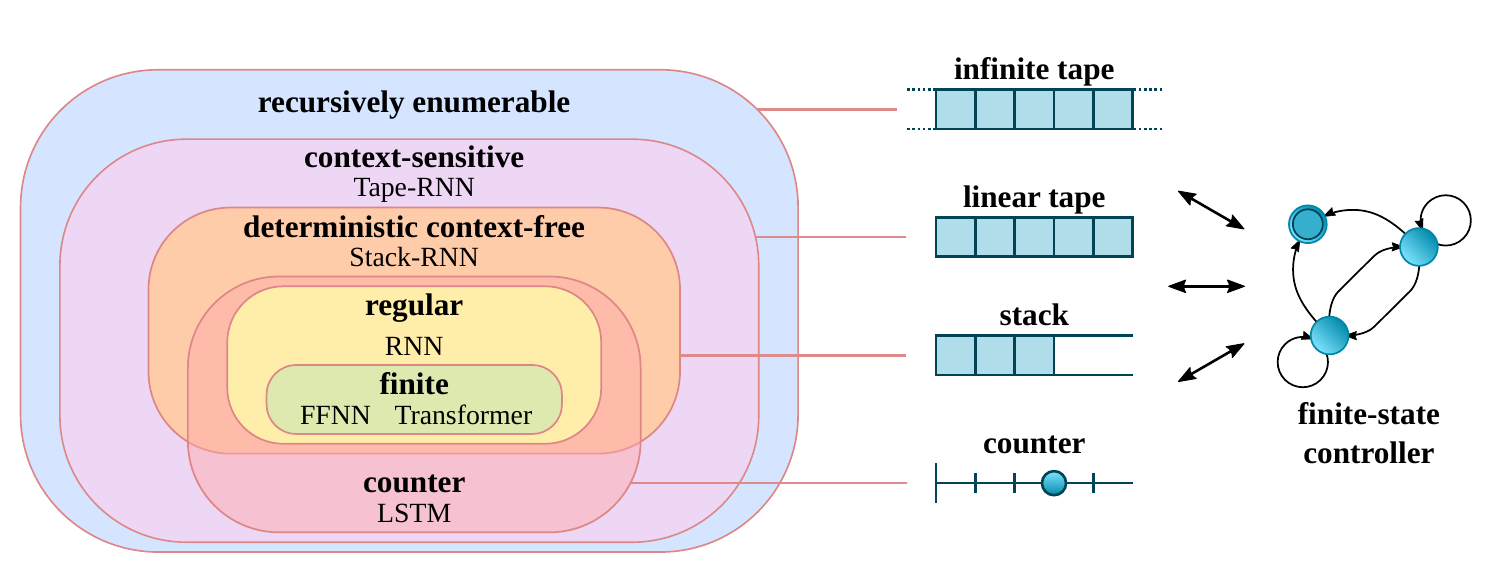}
    \end{center}
    \caption{
        Formal language classes and their correspondence with neural network architectures.
        \emph{Left}: Our empirical evaluation locates the architectures on the hierarchy of formal language classes.
        \emph{Right}: Each formal language class is associated with a minimal computational model (automaton) to recognize or generate the language (see \cref{sec:background}).
        All automata have a finite-state controller at their core, in addition to increasingly restrictive memory access as we descend the hierarchy.
    }
    \label{fig:overview}
\end{figure}

Even if a neural architecture is theoretically universal or Turing complete, gradient-based training, which cannot exhaustively search the parameter space, can impede finding the right solution and thus practically render the model non-universal.
Therefore, both architectural and training limitations impact which sequence prediction problems a model can solve in practice.
In formal language theory, the Chomsky hierarchy~\citep{chomsky1956three} classifies such (sequence prediction) problems by increasing complexity.
This hierarchy is associated with an equivalent hierarchy of models (automata) that can solve different problem classes~\citep{savage1998models, sipser1997introduction}.
Lower-level automata have restrictive memory models and can only solve lower-level problems, while Turing machines with infinite memory and unrestricted memory access lie on top of the hierachy and can solve all computable problems.
However, unlike for classical automata, a unified placement of neural architectures on the Chomsky hierarchy has not yet been practically established, which is precisely the goal of our work.

\paragraph{This work}

We conduct an extensive empirical study with the aim of discovering how neural network models used for program induction relate to the idealized computational models defined by the Chomsky hierarchy \emph{in practice} (see \cref{fig:overview} for a summary of our findings).
We investigate whether the theoretical limitations of certain neural models hold in practice when trained with gradient-based methods.
For example, previous work has theoretically argued that RNNs are Turing complete~\citep{siegelmann1994analog}.
However, more recent theoretical analyses~\citep{ackermann2020survey, merrill2019sequential, weiss2018practical} showed that RNNs lie much lower on the Chomsky hierarchy.
To complement these theoretical analyses, we conduct a large-scale empirical evaluation on sequence prediction problems.
We make the following main contributions:

\begin{itemize}
    \item We conduct an extensive generalization study (\num{20910} models, 15 tasks) of state-of-the-art neural network architectures (RNN, LSTM, Transformer) and memory-augmented networks (Stack-RNN, Tape-RNN) on a battery of sequence-prediction tasks spanning the entire Chomsky hierarchy that can be practically tested with finite-time computation.
    \item We open-source a length generalization benchmark  (\url{https://github.com/deepmind/neural_networks_chomsky_hierarchy}) that is out of reach for state-of-the-art sequence prediction models and allows us to pinpoint the failure modes of these architectures.
    \item We show how increasing amounts of training data do not enable generalization on our tasks higher up in the hierarchy for some architectures (under sufficient capacity to perfectly learn the training data) potentially implying hard limitations for scaling laws~\citep{kaplan2020scaling}.
    \item We demonstrate how augmenting architectures with differentiable structured memory (\eg with a stack or a tape) can enable them to solve tasks higher up the hierarchy.
\end{itemize}
\section{Related work}
\label{sec:related-work}

\paragraph{Learning formal languages}

A long line of work has empirically investigated whether common machine learning architectures, including RNNs~\citep{elman1990finding}, GRUs~\citep{cho2014learning}, SCNs~\citep{giles1992learning, pollack1991induction}, LSTMs~\citep{hochreiter1997long}, and Transformers~\citep{vaswani2017attention}, are capable of learning formal languages.
The main insights are: These networks can learn simple counting languages~\citep{holldobler1997designing, steijvers1996recurrent, wiles1995learning,rodriguez1997recurrent} and the Dyck-1 language~\citep{skachkova2018closing, suzgun2019lstm}.
To learn more advanced languages, RNNs and LSTMs require exponential memory in terms of the input length~\citep{sennhauser2018evaluating}, as they lack an external memory structure.
They are capable of learning simple context-sensitive languages, such as $a^nb^nc^n$, but in a very limited way, \ie they generalize only to lengths close to those seen during training~\citep{boden2000context, boden2002learning, gers2001lstm}.
Similarly, Transformers cannot learn Dyck-$n$ languages for $n > 1$ and long sequences~\citep{ebrahimi2020how}.
Concurrent work also studies length generalization on synthetic reasoning tasks (akin to learning formal languages) but for pretrained large language models~\citep{anil2022exploring, zhang2022unveiling}.
Thus, while prior work has investigated a single architecture on a restricted set of tasks under different experimental conditions, we provide a unified experimental protocol that spans all the levels of the Chomsky hierarchy for a wide range of models.

\paragraph{Neural networks and the Chomsky hierarchy}

It was theoretically shown that RNNs and Transformers are Turing complete~\citep{chen2018recurrent, perez2019turing, perez2021attention, siegelmann1994analog}.
However, these results are impractical as they rely on an unbounded number of recurrent steps and on arbitrary numerical precision.
Thus, more recent work~\citep{ackermann2020survey, bhattamishra2020ability, hahn2020theoretical, hao2022formal, korsky2019computational, merrill2019sequential, merrill2020formal, merrill2022logprecision, weiss2018practical} has refined these theoretical analyses by considering linear computation steps and logarithmic precision, showing that: (i) RNNs and GRUs can, in theory, recognize regular languages, and (ii) LSTMs are strictly more powerful since they can learn a counting mechanism (\ie are k-counter machines).
Moreover, it was theoretically shown that Transformers are not well-aligned with the Chomsky hierarchy since they cannot recognize certain regular languages (\eg periodic finite-state languages), while being able to learn some counter languages (\eg Shuffle-Dyck and $n$-ary Boolean expressions).
A different approach proposed a computational model to capture the Transformer operations and used it to show which tasks could conceivably be learned by a Transformer (histograms, sorting, Dyck languages)~\citep{weiss2021thinking}.
However, this approach only upper-bounds the capabilities of a model and does not provide any insight on whether gradient-based methods will find parameters that can solve a task in practice, which is precisely the goal of our work.
In that sense, our work complements the above studies by investigating how well gradient-based learning can exploit the inductive biases of common machine learning architectures to recognize languages on different levels of the Chomsky hierarchy.

\paragraph{Memory-augmented networks}

A popular approach for augmenting neural architectures with external memory is to equip a recurrent network with read and write access to differentiable memory structures, such as deterministic stacks~\citep{das1992learning, das1992using, grefenstette2015learning, hao2018context, joulin2015inferring, mali2021recognizing, mozer1992connectionist, stogin2020provably, sun1993neural, suzgun2019memory, yogatama2018memory}, nondeterministic stacks~\citep{dusell2020learning, dusell2022learning}, random access memory~\citep{danihelka2016associative, kurach2016neural}, and memory matrices~\citep{graves2014neural, graves2016hybrid, gulcehre2018dynamic, yang2017lie}.
Such memory-augmented networks are capable of recognizing complex languages, including tasks like copying or reversing strings, but they can only perform one computation step per input token.
Thus, to solve superlinear tasks (\eg binary multiplication), memory-augmented networks also need to be able to perform a linear number of steps per token~\citep{freivalds2017improving, kaiser2015neural, price2016extensions}.
In a different effort, reinforcement learning has been applied to interact with discrete memory interfaces, making deterministic memory operations amenable to gradient-based training~\citep{zaremba2015reinforcement, zaremba2016learning}.
Finally, prior work also proposed networks with read-only memory access~\citep{sukhbaatar2015end, weston2014memory} or spatiotemporal connections~\citep{kalchbrenner2015grid}.
We also consider memory-augmented networks and find that memory structures are necessary to generalize on sequence prediction problems higher up the Chomsky hierarchy.
\section{Background}
\label{sec:background}

We now provide the necessary background on how to link sequence prediction and formal language theory, allowing us to associate neural architectures and the Chomsky hierarchy.
As we want to evaluate the capability of networks to generalize to sequences of unseen length, we need a way of generating arbitrarily long sequences that all exhibit the same structure.
To that end, we will consider sequences as elements of infinite formal languages, which are generated by a grammar.

\paragraph{Formal language}

A formal language $L$ is a set of words, where each word is a finite string of symbols drawn from a finite alphabet $\Sigma$.
For example, with a binary alphabet $\Sigma=\{0,1\}$ we can describe the following infinite languages:
$A = \{ 0^n1^m~|~n,m>0 \}$, $B = \{ w~|~w~\mathrm{is~a~palindrome} \}$, $C = \{ ww~|~w~\mathrm{is~a~word} \}$,
and $D = \{ w~|~w~\mathrm{describes~a~terminating~Turing~machine} \}$.

\paragraph{Generative grammar}

Languages of infinite cardinality can be generated with a finite alphabet and a finite generative grammar.
A formal grammar~\citep{chomsky1956three} is defined by a finite set of terminal symbols, a finite set of non-terminal symbols, a distinctive start symbol, and a finite set of production rules that map from non-terminal symbols to other symbols.
The set of all words that can be formed by such a grammar is its corresponding formal language.
For example, the binary palindrome language $B$ can be generated by the set of rules: $P \rightarrow \epsilon \mid 0 \mid 1 \mid 0P0 \mid 1P1$, with start and nonterminal symbol $P$ and terminal symbols $\{0, 1\}$.
The Chomsky hierarchy arises from categorizing the allowed relationships between terminals and non-terminals in the grammar (see \cref{tab:chomsky-hierarchy}).
For example, $B$ is not regular, since the right-hand side contains a variable enclosed by two terminal symbols, but context-free, since the left-hand side consists of a single nonterminal symbol.

\begin{table}
    \caption{
        Chomsky hierarchy grammar types, their corresponding minimal automata and memory structures required for language recognition or generation, their production rules, and the corresponding example languages from \cref{sec:background}.
        The production rules symbols are defined as: $a$ terminal; $A,B$ non-terminal; $\alpha, \beta, \gamma$ string of terminals and/or non-terminals ($\alpha, \beta$ can be empty, but not $\gamma$).
    }
    \begin{center}
        \resizebox{\columnwidth}{!}{\begin{tabular}{@{}lllll@{}}
    \toprule
    \textbf{Grammar type (low $\to$ high)} & \textbf{Automaton} & \textbf{Memory} & \textbf{Production rules} & \textbf{Example} \\
    \midrule
    \Gls*{regular}  & \Gls*{fsa}    & Automaton state                                   & $A \rightarrow a \mid aB$ & A \\
    \Gls*{cf}       & \Gls*{pda}    & $+$ infinite stack (only top entry accessible)    & $A \rightarrow \alpha$         & B \\
    \Gls*{cs}       & \Gls*{lba}    & $+$ bounded tape (all entries accessible)         & $\alpha A \beta \rightarrow \alpha\gamma\beta$ & C \\
    \Gls*{re}       & \Gls*{tm}     & $+$ infinite tape (all entries accessible)        & $\gamma \rightarrow \alpha$ & D \\
    \bottomrule
\end{tabular}}
    \end{center}
    \label{tab:chomsky-hierarchy}
\end{table}

\paragraph{Recognizing a language}

An automaton recognizes a language if it accepts all strings in the language and no others.
For example, a (deterministic) finite-state automaton \acrfull*{fsa} can be defined as a 5-tuple $(Q, \Sigma, \delta, q_0, F)$, consisting of a finite set of states $Q$, the finite alphabet $\Sigma$, a transition function $\delta : Q \times \Sigma \rightarrow Q$, an initial state $q_0 \in Q$, and a set of accept states $F \subseteq Q$.
Thus, an \gls*{fsa} accepts a string $w = a_1 a_2 \ldots a_n$ over the alphabet $\Sigma$ if there exists a sequence of states $r_0, r_1, \ldots, r_n \in Q$ such that $r_0 = q_0$, $r_{i + 1} = \delta(r_i, a_{i+1})$ for $i = 0, \ldots, n - 1$, and $r_n \in F$.
More complex automata, such as the \acrfull*{pda}, additionally have access to an external memory structure.
The Chomsky hierarchy classifies languages based on the automaton that can recognize it (see \cref{tab:chomsky-hierarchy}).
All automata in the hierarchy have a finite state controller at their core, with increasingly flexible memory structures.
Language classes generated by higher grammar types subsume languages generated by those lower in the hierarchy, \eg all \acrlong*{regular} languages are \gls*{cf}, all \gls*{cf} languages are \gls*{cs}, etc.
For technical reasons, we differentiate \gls*{dcf} and \gls*{ndcf} languages, since nondeterministic push-down automata (PDAs) can recognize some \gls*{cf} languages that deterministic \gls*{pda}s cannot.
We also consider (real-time) counter languages, which are recognized by finite automata with one or more (infinite) counters (or, equivalently, a \gls*{pda} with one or more stacks and a single stack symbol).
Counter languages are a proper superset of \acrlong*{regular} languages and a proper subset of \gls*{cs} languages (see \cref{fig:overview}).

\paragraph{Transduction \versus recognition}

In practice, learning to recognize a formal language by training a classifier on words and non-words is hard because there is, in general, no formal construction for generating a sufficient but finite set of negative examples.
For this reason, we consider language transduction tasks, where the inputs and outputs form two individual formal languages, and the goal is to learn a deterministic function that maps a word from one language to the other.
Concretely, a deterministic finite transducer, defined as a 6-tuple $(Q, \Sigma_I, \Sigma_O, \delta, q_0, F)$, differs from a finite-state automaton in that its transition function $\delta : Q \times (\Sigma_I \cup \{\varnothing\}) \rightarrow (\Sigma_O \cup \{\varnothing\}) \times Q$ can output a symbol from the output alphabet $\Sigma_O$ (or the dummy token $\varnothing$) at every step.
Thus, a transducer maps $w_I = a_1 a_2 \ldots a_n$ to $w_O = b_1 b_2 \ldots b_n$ if there exists a sequence of states $r_0, r_1, \ldots, r_n \in Q$ such that $r_0 = q_0$, $(b_{i + 1}, r_{i + 1}) = \delta(r_i, a_{i+1})$ for $i = 0, \ldots, n - 1$, and $r_n \in F$.
The transducer outputs $b_1 = \ldots = b_k = \varnothing$ while reading input word $w'_I = a_1 \ldots a_k \in \Sigma_I$ and outputs $w'_O = b_{k+1} \ldots b_n \in \Sigma_O$ while reading dummy tokens $a_{k+1} = \ldots = a_n = \varnothing$.
Transducers therefore define a deterministic function $f: \Sigma_I^\ast \rightarrow \Sigma_O^\ast$, given by $f(w'_I) = w'_O$, between two formal languages, which is precisely what we want to learn.
Thus, strictly speaking, we do not consider a hierarchy over formal languages but over language transduction tasks.
Importantly, the machines required are identical to
those in the Chomsky hierarchy, \ie \gls*{fsa}s, \gls*{pda}s, \gls*{lba}s, and \gls*{tm}s (see \cref{tab:chomsky-hierarchy}), except that they can now also output values at each transition (\ie the memory augmentations stay the same).
We further discuss the transition from recognition to transduction, as well as prior approaches, in \cref{sec:experimental-details:recognition-vs-transduction}.

\section{Methods}
\label{sec:methods}

\paragraph{Problem setup}

For each task we define input and output languages $L_I$ and $L_O$, with their corresponding alphabets $\Sigma_I$ and $\Sigma_O$.
The goal of a task is to learn the deterministic function $f$ that maps words $\bm{x} \in L_I$ to words $\bm{y} \in L_O$, \ie language transduction.
To that end, we consider models $p_\theta(\cdot | \bm{x})$ as conditional distributions over next possible sequence continuations (with parameters $\theta$).
We one-hot encode the tokens in both languages $L_I$ and $L_O$ as vectors of length $|\Sigma_I|$ and $|\Sigma_O|$, respectively, denoting the corresponding $j$th entries of the $i$th one-hot tokens $x_{ij}$ and $y_{ij}$.
We use gradient-based training to minimize the average cross-entropy loss, where $C(\bm{x}, \bm{y}) := - \frac{1}{|\bm{y}|}\sum_{i} y_i^T \log p_\theta(\cdot|\bm{x})$ denotes the per-example loss (see \cref{alg:pipeline}).
Note that we do not use auto-regressive models, but rather append $|\bm{y}|$ empty dummy tokens to the input.
Thus, the model will know when the input sequence $\bm{x}$ has terminated by reading the first empty token.  
We compute the per-sequence accuracy as the percentage of correctly predicted tokens, \ie $A(\bm{x}, \bm{y}) := \frac{1}{|\bm{y}|}\sum_{i}\mathds{1} \left[(\argmax_j y_{ij}) = (\argmax_j p_\theta(\cdot|\bm{x})_j)\right]$.
Moreover, we compute the overall performance score as the per-sequence accuracy averaged over all sequences of unseen length.
We run our experiments over 10 different network parameter initialization seeds and report the maximum score instead of the average, as this better indicates if an architecture is capable of learning the right algorithm \emph{at all} (we provide the means and variances in \cref{sec:additional-experiments}).

\paragraph{Tasks}

For each level of the Chomsky hierarchy, we define various sequence prediction tasks ranging from modular arithmetic (\gls*{regular}) to binary addition (\gls*{cs}).
We place the most emphasis on \gls*{cs} tasks as they are currently out of reach for state-of-the-art models and thus provide an ideal benchmark to guide future research on architecture design.
We list the tasks in \cref{tab:scores} and provide a full formalization in \cref{tab:tasks}.
Although seemingly simple, these tasks are well-known examples in the theory of computation and concisely demonstrate the key properties of the grammars in the Chomsky hierarchy, \ie counting, repetition, long-distance dependencies, and hierarchy.
Note that the levels of our tasks are independent of the levels of $L_I$ or $L_O$ but correspond to the automata that can solve the corresponding transduction tasks.
For example, for the \reversestring{} task, $L_I = L_O = \{a, b\}^*$, which are both \acrlong*{regular}.
However, learning the function $f$ that maps $w \in L_I$ to $reverse(w) \in L_O$ requires a deterministic push-down transducer, thus rendering the task \acrlong*{dcf}.

\paragraph{Architectures}

We consider a wide range of neural network architectures, including both state-of-the-art and memory-augmented models (full details in \cref{sec:experimental-details:models}).
Our goal is to have at least one model per level of the Chomsky hierarchy.
To that end, we use a standard RNN as a controller and augment it with two kinds of differentiable memory structures: a deterministic stack and a bounded tape.
The stack and tape have elements in $\mathbb{R}^d$ with $d = 8$, and we ensure that the stack and tape sizes are large enough to simulate infinite memory during training and testing.
The Stack-RNN~\citep{joulin2015inferring} can perform any linear combination of \push{}, \pop{}, and \noop{} operations on the stack.
The Tape-RNN, which is inspired by the Baby-NTM~\citep{suzgun2019memory}, can perform any linear combination of 
\writeleft{}, \writeright{}, \writestay{}, \jumpright{}, \jumpleft{} (details in \cref{sec:experimental-details}).
In addition to these memory-augmented models, we also evaluate LSTMs~\citep{hochreiter1997long} and Transformers~\citep{vaswani2017attention}.
We compute the Transformer output from the encoder only model that overlooks the whole input sequence.
We consider five different positional encodings: none, classical $\sin$/$\cos$~\citep{vaswani2017attention}, the rotary positional encodings (RoPE) \citep{su2021roformer}, ALiBi~\citep{press2021train}, and the relative positional encodings of Transformer-XL~\citep{dai2019transformer}.
The latter two have been shown to allow extrapolation to longer sequences on some language modeling tasks.
All models receive the tokens $\bm{x}$ as an input and are expected to produce the output $\bm{y}$ (\ie we do not feed the output back to the input unlike sequence-to-sequence models~\citep{sutskever2014sequence}).
We study CNNs, Stack-LSTMs, NDStack-RNNs~\citep{dusell2020learning, dusell2022learning}, and autoregressive versions of our models in \cref{sec:additional-experiments}.

\paragraph{Data generation}

Instead of using fixed-size datasets, we define training and test distributions from which we continually sample sequences.
We define the maximum training sequence length as the \emph{training range} $N$, with $N = 40$.
Every mini-batch is generated by first sampling a length $\ell$ from the uniform distribution $\mathcal{U}(1, N)$ and then sampling sequences of length $\ell$ from the task's generative grammar.
For testing, we sample the sequence length $\ell$ from $\mathcal{U}(N+1, M)$, with $M=500$.

We provide an open-source implementation of our models, tasks, and training and evaluation suite at \url{https://github.com/deepmind/neural_networks_chomsky_hierarchy}.

\section{Results}
\label{sec:results}

In this section, we provide the results of our empirical evaluation.
In \cref{sec:results:main-result} we relate the networks to the Chomsky hierarchy, in \cref{sec:results:analysis} we analyze the algorithms learned by the networks, and in \cref{sec:results:lstms,sec:results:transformers} we discuss the performance of LSTMs and Transformers in more detail.

\begin{table}
  \caption{
    Score (in percentage, see \cref{sec:methods}), \ie accuracy averaged over all test lengths and maximized over 10 random seeds (and other hyperparameters, see \cref{sec:experimental-details}).
    We consider a model to generalize successfully (bold) if its $\mathrm{score} \geq 90\%$.
    The random accuracy is $50\%$ (except for \cyclenavigation{}~(\gls*{regular}), \bucketsort{} (\gls*{cs}), and the two modular arithmetic tasks where it is $20\%$).
    We observe that, in general, RNNs with more permissive memory structures can solve more challenging tasks.
    We denote permutation-invariant tasks with $\dagger$, counting tasks with $\star$, tasks requiring a nondeterministic controller with $\circ$, and tasks requiring superlinear running time with $\times$.
  }
  \label{tab:scores}
  \begin{center}
    \resizebox{\columnwidth}{!}{\begin{tabular}{@{}llrrrrr@{}}
    \toprule
    \textbf{Level} & \textbf{Task}  & \textbf{RNN} & \textbf{Stack-RNN} & \textbf{Tape-RNN} & \textbf{Transformer} & \textbf{LSTM} \\
    \midrule
    \multirow{4}{*}{\Gls*{regular}}
    & \evenpairs & \textbf{100.0} & \textbf{100.0} & \textbf{100.0} & \textbf{96.4} & \textbf{100.0} \\
    & \modulararithmeticsimple & \textbf{100.0} & \textbf{100.0} & \textbf{100.0} & 24.2 & \textbf{100.0} \\
    & \paritycheck$^\dagger$ & \textbf{100.0} & \textbf{100.0} & \textbf{100.0} & 52.0 & \textbf{100.0} \\
    & \cyclenavigation$^\dagger$ & \textbf{100.0} & \textbf{100.0} & \textbf{100.0} & 61.9 & \textbf{100.0} \\
    \\
    \multirow{4}{*}{\Gls*{dcf}}
    & \stackmanipulation & 56.0 & \textbf{100.0} & \textbf{100.0} & 57.5 & 59.1 \\
    & \reversestring & 62.0 & \textbf{100.0} & \textbf{100.0} & 62.3 & 60.9 \\
    & \modulararithmeticbrackets & 41.3 & \textbf{96.1} & \textbf{95.4} & 32.5 & 59.2 \\
    & \solveequation$^\circ$ & 51.0 & 56.2 & 64.4 & 25.7 & 67.8 \\
    \\
    \multirow{7}{*}{\Gls*{cs}}
    & \duplicatestring & 50.3 & 52.8 & \textbf{100.0} & 52.8 & 57.6 \\
    & \missingduplicate & 52.3 & 55.2 & \textbf{100.0} & 56.4 & 54.3 \\
    & \oddsfirst & 51.0 & 51.9 & \textbf{100.0} & 52.8 & 55.6 \\
    & \binaryaddition & 50.3 & 52.7 & \textbf{100.0} & 54.3 & 55.5 \\
    & \binarymultiplication$^\times$ & 50.0 & 52.7 & 58.5 & 52.2 & 53.1 \\
    & \computesqrt & 54.3 & 56.5 & 57.8 & 52.4 & 57.5 \\
    & \bucketsort$^{\dagger\bigstar}$ & 27.9 & 78.1 & 70.7 & \textbf{91.9} & \textbf{99.3} \\
    \bottomrule
\end{tabular}
 }
  \end{center}
\end{table}

\begin{figure}
    \begin{center}
        \begin{subfigure}[b]{0.315\textwidth}
            \begin{center}
                \includegraphics[width=\textwidth]{{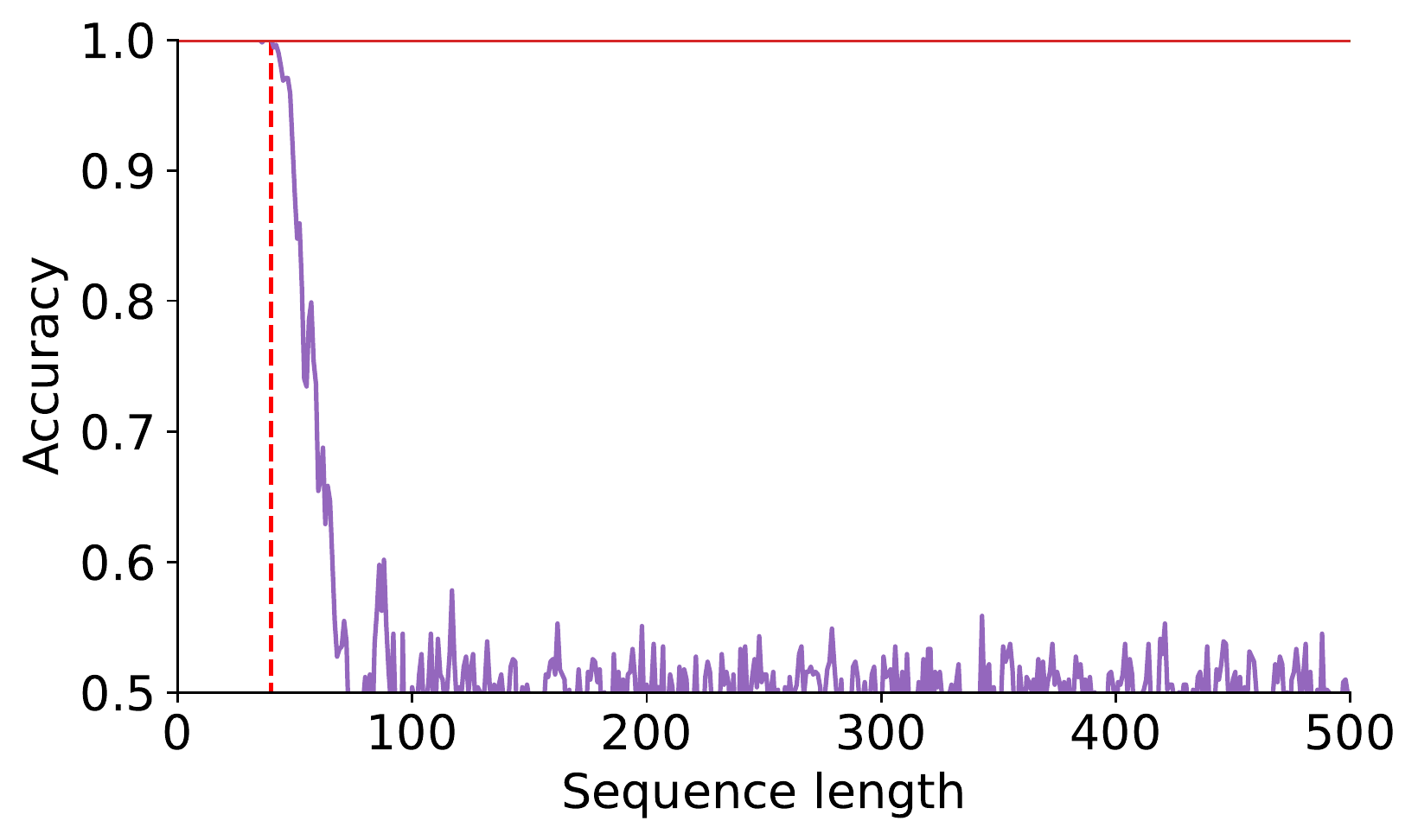}}
            \end{center}
            \caption{\paritycheck{} (\gls*{regular})}
        \end{subfigure}
        \hspace{0.01\textwidth}
        \begin{subfigure}[b]{0.315\textwidth}
            \begin{center}
                \includegraphics[width=\textwidth]{{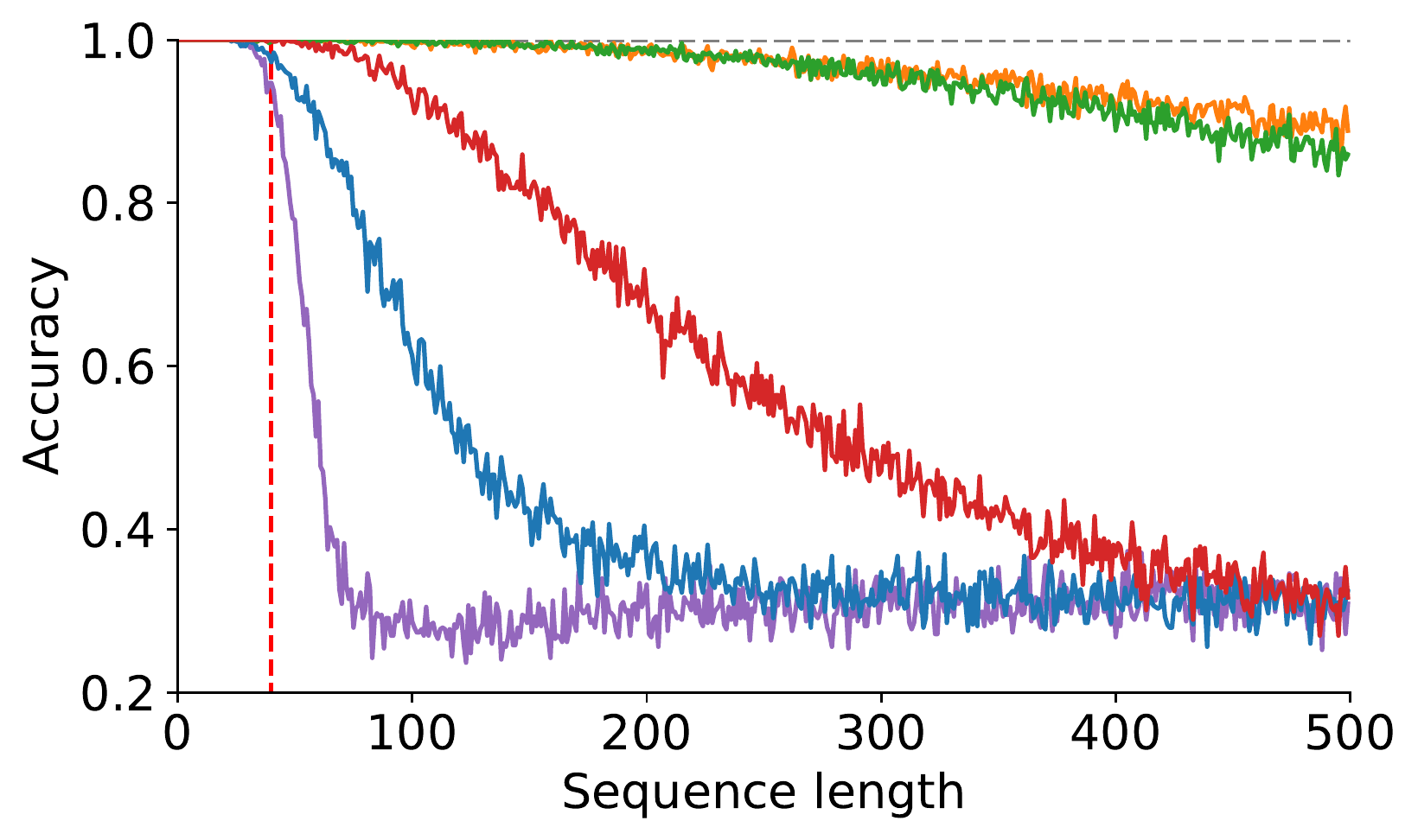}}
            \end{center}
            \caption{\modulararithmeticbrackets{} (\gls*{dcf})}
        \end{subfigure}
        \hspace{0.01\textwidth}
        \begin{subfigure}[b]{0.315\textwidth}
            \begin{center}
                \includegraphics[width=\textwidth]{{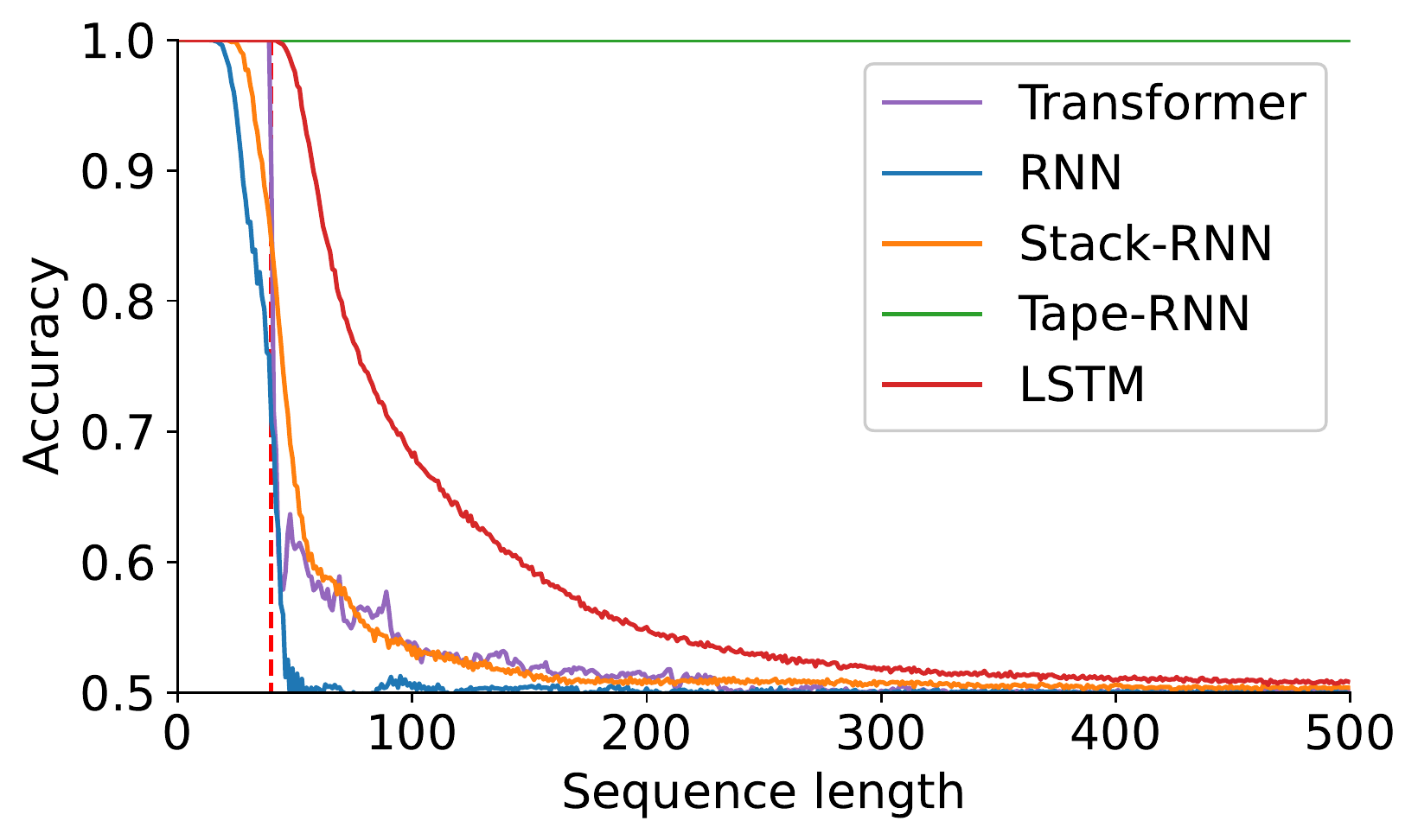}}
            \end{center}
            \caption{\duplicatestring{} (\gls*{cs})}
        \end{subfigure}
    \end{center}
    \caption{
        Performance curves on three tasks.
        The dashed vertical red line is the training range, meaning that sequences to the right have not been seen during training and thus measure generalization.
    }
    \label{fig:scores}
\end{figure}

\subsection{Main result}
\label{sec:results:main-result}

We evaluate all architectures (defined in \cref{sec:methods}) on all tasks (defined in \cref{tab:tasks}) and show the generalization accuracy averaged over the segment $\mathcal{U}(N+1, M)$ (as described in \cref{sec:methods}) in \cref{tab:scores}.
Furthermore, \cref{fig:scores} displays the accuracy per sequence length for the best-performing model per architecture on three sample tasks (see \cref{sec:additional-experiments} for the remaining tasks).
As the test sequences have not been seen during training, we can gauge whether the architectures can learn the ``correct'' algorithm.
We observe that the networks generally match the computational models associated with the Chomsky hierarchy: RNNs can solve tasks up to the \acrlong*{regular} level, Stack-RNNs up to the \gls*{dcf} level, and Tape-RNNs up to the CS level.
However, the match is not perfect, as some architectures cannot solve tasks at their supposed level, \eg Stack-RNN cannot solve \solveequation{} (\gls*{dcf}) and the Tape-RNN cannot solve the (challenging) \binarymultiplication{}~(\gls*{cs}) and \computesqrt{} (\gls*{cs}) tasks.
This may be due to shortcomings in the architecture, the training protocol, or the particular difficulty of the tasks.
For example, \solveequation{}~(\gls*{dcf}) is context-free but requires a nondeterministic \gls*{fsa} controller because it has to compute multiple modular arithmetic expressions for the different values of the unknown variable $x$ and compare their results to the actual solution of the equation, which is only provided at the end of the input sequence.
Similarly, \binarymultiplication{}~(\gls*{cs}) is a context-sensitive task but has a quadratic complexity, which is infeasible for the architectures we investigate.
Conversely, some architectures show non-trivial performance on tasks beyond their supposed level, such as the Stack-RNN on \bucketsort{}~(\gls*{cs}) (more details in \cref{sec:additional-experiments}).
Finally, we observe that Transformers and LSTMs are not well-aligned with the Chomsky hierarchy: Transformers fail on \acrlong*{regular} tasks, while LSTMs can solve (counting) tasks more difficult than \acrlong*{regular}.
We discuss both cases in detail below (\cref{sec:results:lstms,sec:results:transformers}).

\subsection{Analysis of learned program structure}
\label{sec:results:analysis}

\begin{figure}
    \begin{center}
        \begin{subfigure}[b]{0.24\textwidth}
            \includegraphics[width=\textwidth]{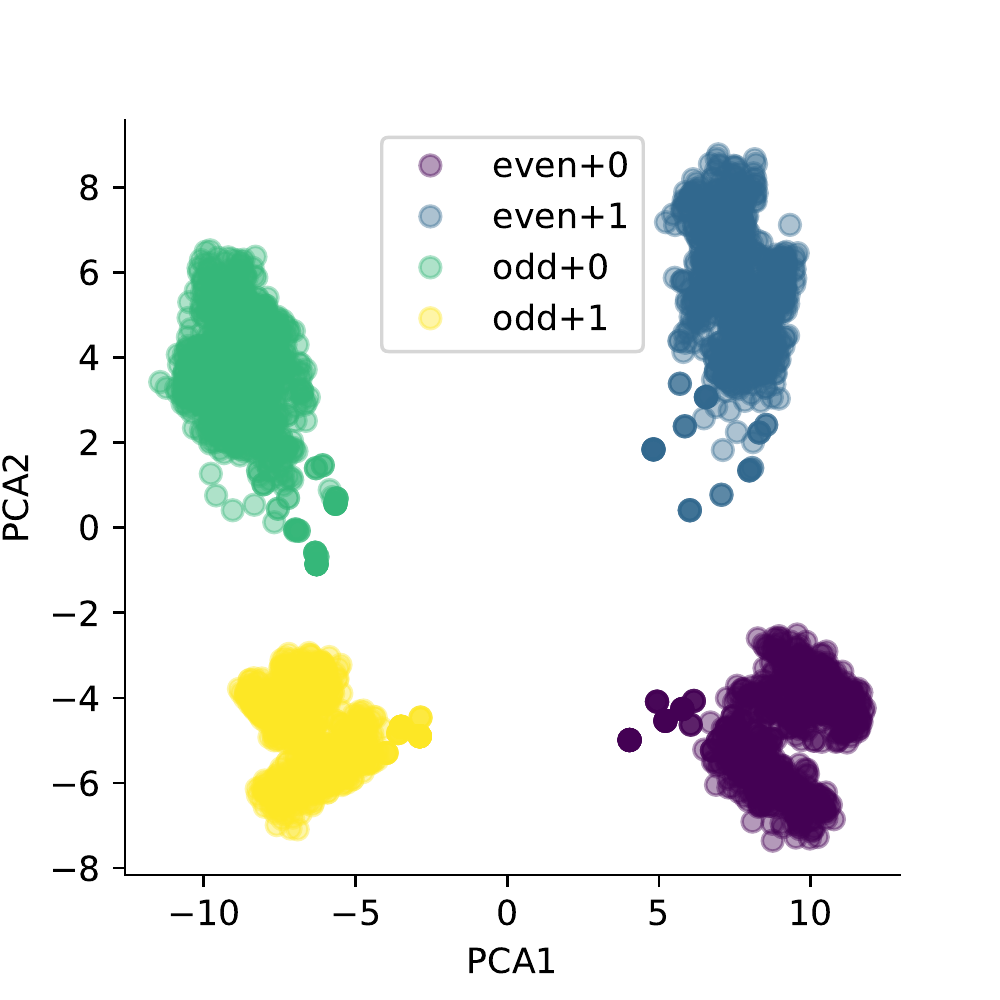}
            \caption{Internal RNN states on \paritycheck{} (\gls*{regular}), colored by the previous state and the current input.}
            \label{fig:analysis-rnn:states}
        \end{subfigure}
        \hspace{0.01\textwidth}
        \begin{subfigure}[b]{0.27\textwidth}
            \includegraphics[width=\textwidth]{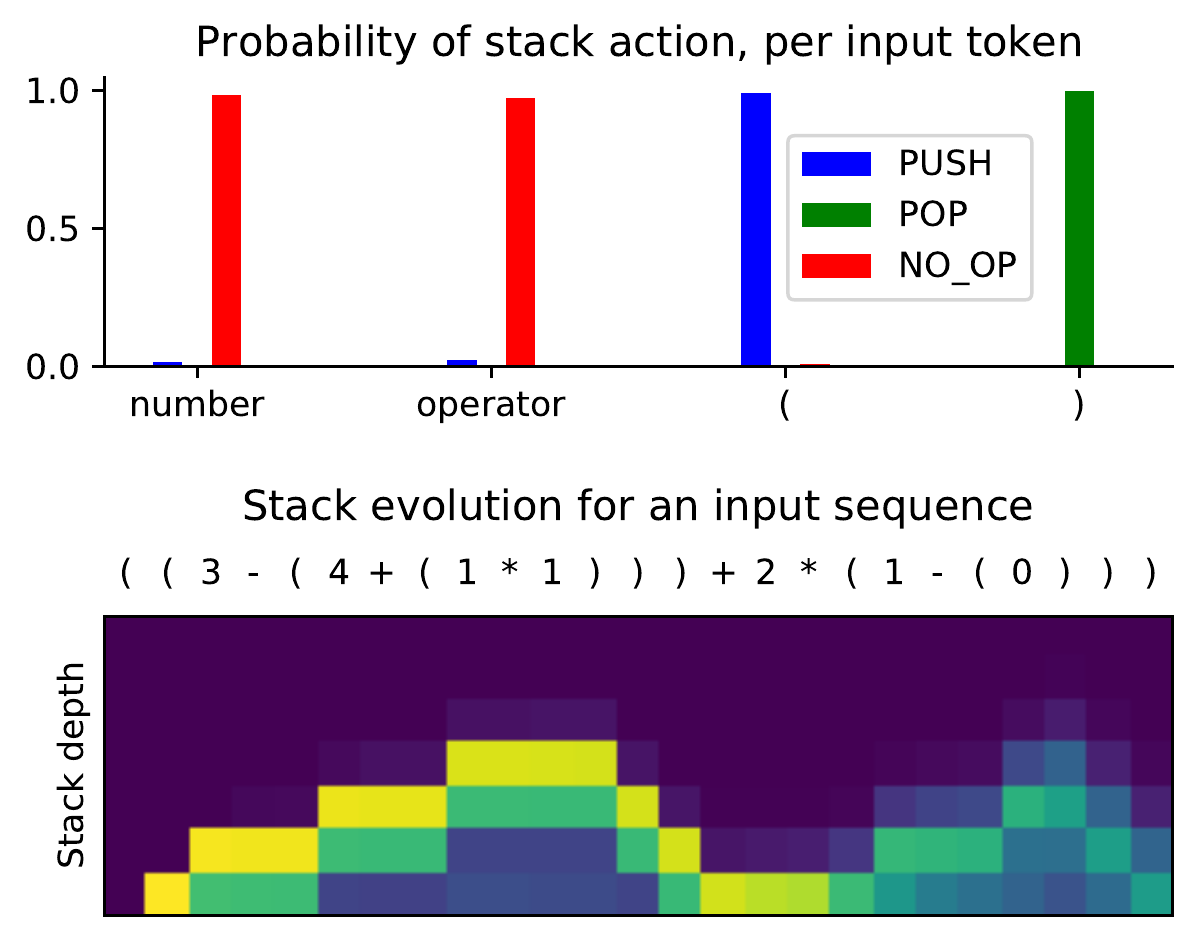}
            \caption{Stack-RNN action probabilities (\textit{above}) and stack values (\textit{below}) on the \modulararithmeticbrackets{} (\gls*{dcf}) task.}
            \label{fig:analysis-rnn:stack}
        \end{subfigure}
        \hspace{0.01\textwidth}
        \begin{subfigure}[b]{0.42\textwidth}
            \includegraphics[width=\textwidth]{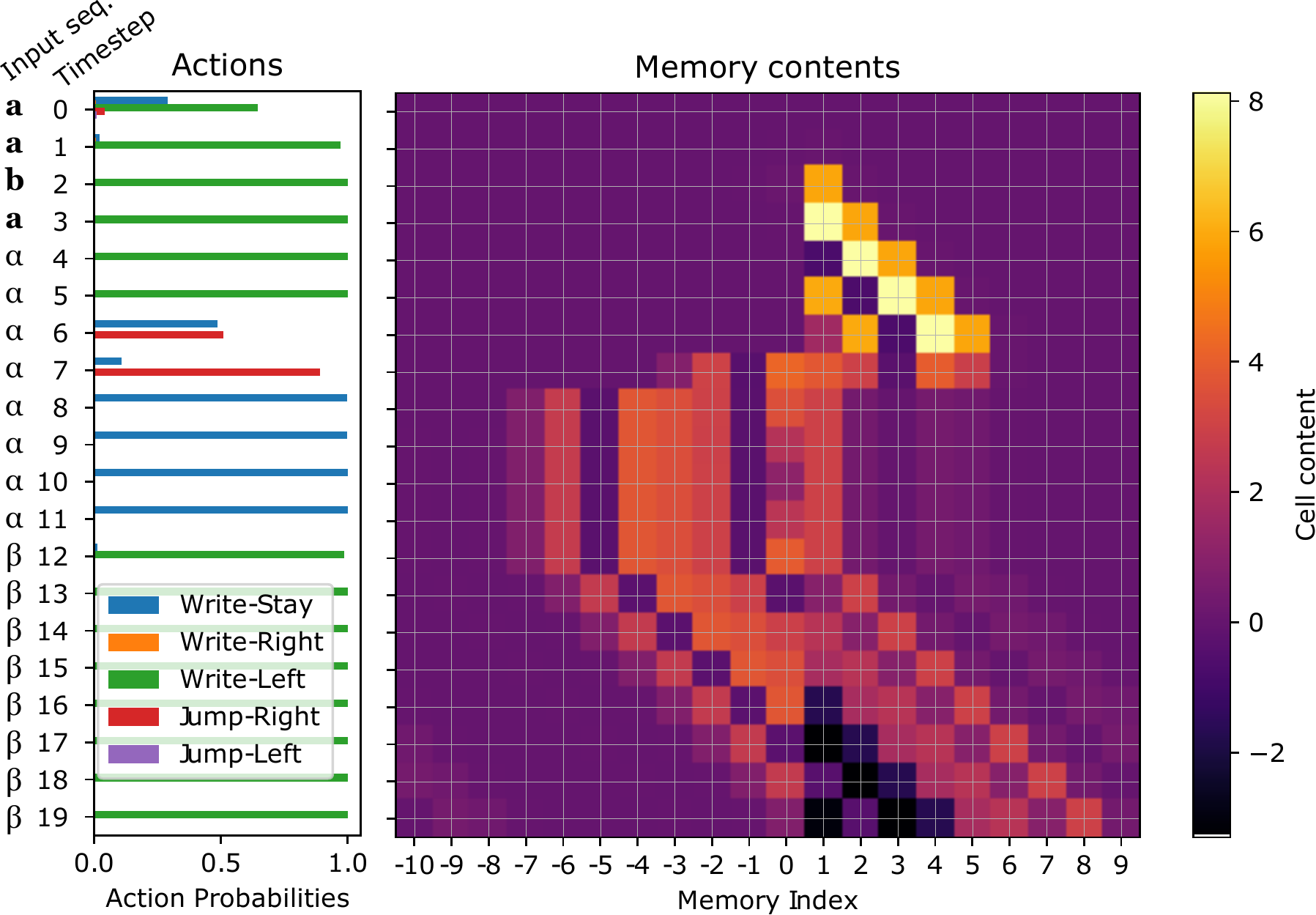}
            \caption{Tape-RNN action probabilities and memory values on \duplicatestring{} (\gls*{cs}).}
            \label{fig:analysis-rnn:tape}
        \end{subfigure}
    \end{center}
    \caption{
        Analysis of the internal representations and memory dynamics of an RNN on a \acrlong*{regular} task, a Stack-RNN on a \gls*{dcf} task, and a Tape-RNN on a \gls*{cs} task.
        The reverse-engineered network mechanisms are indicative of the correct algorithms that solve the corresponding tasks.
    }
    \label{fig:analysis-rnn}
\end{figure}

To provide further evidence supporting the claim that some of the networks are able to learn the right algorithm, we now provide an analysis of solutions learned by some of our networks on our tasks, by investigating internal state representations and memory-update dynamics.

\paragraph{\Acrlong*{regular} tasks}

According to the Chomsky hierarchy, we expect networks that solve regular tasks to simulate a \acrlong*{fsa}.
We investigate this hypothesis by analyzing the internal states of the RNN controller and checking whether they are finite (or form a finite number of clusters).
\Cref{fig:analysis-rnn:states} shows a scatter plot of a PCA projection of the internal RNN states when trained on \paritycheck{}~(\gls*{regular}).
We observe that the states form four clusters, each representing one of the four values of the tuple (subsequence result, last token seen).
In principle, one can solve this task with only two states, corresponding to the subsequence result: If the state is 1 and the input token is 1, move to state 0, otherwise stay in state 1, and finally output the state itself.
However, due to the problem symmetry, the network also learns to compute the parity of the number of zeros in the string, therefore adding another two states to the computation.
We also investigate whether and how Stack-RNNs use their stack for this task, which is unnecessary, in principle.
Indeed, the Stack-RNNs only use \noop{} or \pop{} (which is equivalent to \noop{} in the absence of \push{}) actions, indicating that the RNN controller uses only a \acrlong*{fsa} and does not rely on the stack (see \cref{sec:additional-experiments:analysis} for more details).

\paragraph{\Acrlong*{dcf} tasks}

On \gls*{dcf} tasks, we anticipate Stack-RNNs to use their stack.
Indeed, on \reversestring{} (\gls*{dcf}), the networks learn the expected algorithm: \push{} until an empty token appears, and then \pop{} until the stack is empty.
Moreover, the internal state of the controller seems to be largely guided by the last input token and not the last state (details in \cref{sec:additional-experiments:analysis}).
On \modulararithmeticbrackets{} (\gls*{dcf}), which requires a stack to remember past results when entering a new bracket, the networks use the stack exactly as expected.
\Cref{fig:analysis-rnn:stack} shows that the Stack-RNN learns to solve this task by: (i) pushing the last result when a bracket is opened, (ii) not operating on the stack and only using its internal controller when a number or an operator is seen, and (iii) popping the last result from the stack when a bracket is closed.
The top plot reports the probability of each action given an input, and the bottom plot shows the stack evolution along the sequence.
On the bottom plot, the x-axis is the timestep and the y-axis is the cell position in the stack (top of the stack on the bottom). Each cell is colored according to the first principal component of its actual content.

\paragraph{\Acrlong*{cs} tasks}

In contrast to the RNN and Stack-RNN, the reverse-engineered network mechanisms for the Tape-RNN are less interpretable.
Neverthless, \cref{fig:analysis-rnn:tape} shows that the Tape-RNN has learned an algorithm that is indicative of the data-generating grammar of the \duplicatestring{} (\gls*{cs}) task.
The input to the network consists of the sequence $a, a, b, a$ of length $\ell = 4$, followed by $2\ell$ $\alpha$-symbols that the RNN can use to compute and manipulate the tape, followed by $2\ell$  $\beta$-symbols used to output the result (\ie the sequence $a, a, b, a, a, a, b, a$ of length $2\ell$).
The panels show the action probabilities and memory contents over time.
We observe that up to time $t=5$ the RNN writes on the memory with \writeleft{}.
Then, on time-step $6$ the RNN mixes \jumpright{} and \writestay{} to duplicate the input string in the memory, before moving back to the initial position with \jumpright{} on time-step $7$.
After this, the RNN idles using \writestay{} until it reaches the first $\beta$-symbol, whereupon it starts outputting the result using \writeleft{} actions.

\subsection{LSTMs}
\label{sec:results:lstms}

LSTMs are similar to RNNs as they also carry a hidden state and are unrolled over a sequence of inputs.
However, prior work showed that LSTMs are strictly more powerful than RNNs because they can also solve counting tasks~\citep{boden2000context, boden2002learning, gers2001lstm, merrill2019sequential, weiss2018practical}.
Indeed, \Cref{tab:scores} provides further evidence for this claim since LSTMs are capable of solving \bucketsort{} (\gls*{cs}) almost perfectly, even though the task is \acrlong*{cs}.
That is, counting tasks can reside on levels higher than \acrlong*{regular} in the Chomsky hierarchy, since every Turing machine can, in theory, be simulated by a two-counter machine~\citep{minsky1967computation}.
However, finding an architecture that solves non-regular tasks with counters via gradient descent is more difficult than when the controller has access to a more permissive memory structure, such as a stack or a tape.

\subsection{Transformers}
\label{sec:results:transformers}

Unlike the other architectures, Transformers are \emph{permutation invariant} \wrt the position of a token in a sequence because they rely on a global attention mechanism.
As a result, Transformers are more scalable since all tokens can be processed in parallel, however, at the cost of reduced expressivity (\ie permutation invariance).
Similar to real-world problems, most of our tasks are not permutation invariant, \eg reversing a string depends on the tokens' position.
To overcome this problem, Transformers are generally augmented with positional encodings, which are added to the tokens to simulate some notion of position.
Consequently, the augmented tokens intrinsically contain position information, \ie the token $x$ at position 0 will no longer be considered the same as the same token $x$ at position 10.
As described in \cref{sec:methods}, we evaluate five encodings (none, $\sin$/$\cos$, RoPE, ALiBi, and the relative positional encoding from Transformer-XL) and report the best-performing variant.

\begin{figure}
 \begin{center}
     \begin{subfigure}[b]{0.49\textwidth}
        \begin{center}
            \includegraphics[width=0.65\textwidth]{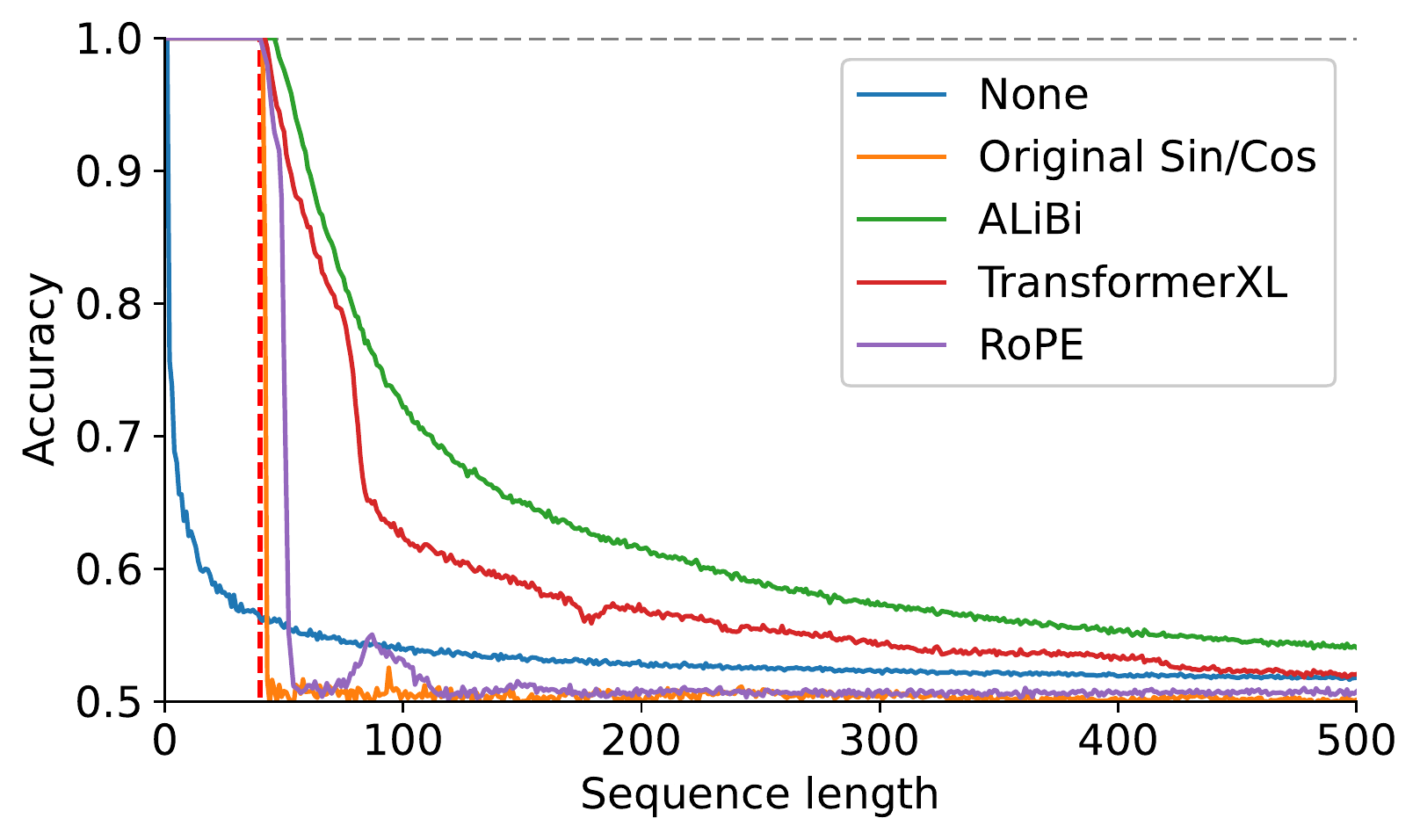}
        \end{center}
        \caption{Accuracy per length by positional encoding.}
        \label{fig:analysis-transformers:accuracy}
     \end{subfigure}
     \hspace{0.005\textwidth}
     \begin{subfigure}[b]{0.49\textwidth}
        \begin{center}
            \includegraphics[width=0.65\textwidth]{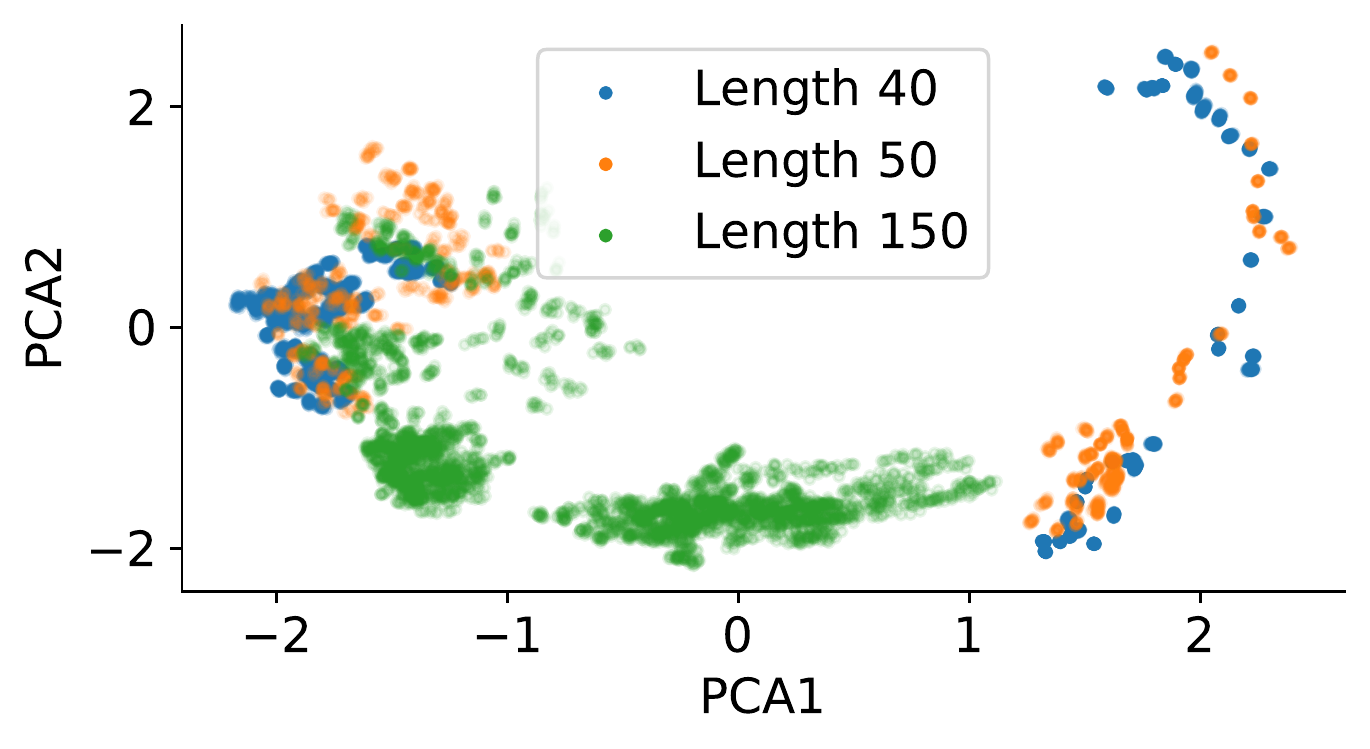}
        \end{center}
        \caption{First layer activations (PCA) by sequence length.}
        \label{fig:analysis-transformers:pca}
     \end{subfigure}
 \end{center}
 \caption{
    Transformers on \reversestring{} (\gls*{dcf}).
    Without positional encodings, Transformers are permutation invariant, and with encodings, positions are out-of-distribution for longer sequences. 
  }
 \label{fig:analysis-transformers}
\end{figure}

Indeed, \Cref{tab:scores} shows that Transformers are most successful on permutation-invariant tasks: They solve \bucketsort{}~(\gls*{cs}) and demonstrate non-trivial generalization on \cyclenavigation{}~(\gls*{regular}).
Moreover, they solve \evenpairs{} (\gls*{regular}), which is not permutation-invariant but only requires checking whether the first and last tokens are equal (\ie all other positions are irrelevant).
However, for all other tasks, Transformers fail to generalize, regardless of their positional encodings (see \cref{fig:analysis-transformers:accuracy} for the different positional encodings on \reversestring{}~(\gls*{dcf}) and \cref{fig:scores-transformer-encodings} for the remaining tasks).
In particular, Transformers are unable to solve all permutation-invariant tasks (\ie \paritycheck{}~(\gls*{regular}), which is a well-known failure mode~\citep{chiangparitycheck}).
We hypothesize that this poor performance is due to the positional encodings, which take on new values when the sequences grow longer (even for relative encodings), meaning that the network activations will be out-of-distribution.
Indeed, \cref{fig:analysis-transformers:pca} shows that the 2D PCA of the first layer activations is similar for sequences of length 40 (\ie the training range, in blue) and 50 (in orange), but completely different for length 150 (in green).
All our other models, which do not rely on global self-attention, do not suffer from this effect, as they only have access to a local view of the input token (and, potentially, a local memory reading).

\section{Discussion}
\label{sec:discussion}

In this paper, we considered generalization from a sequence prediction viewpoint, using formal language theory to establish the computational complexity of each task.
Naturally, we had to choose a maximum test sequence length for our evaluations, which theoretically renders all of our input and output languages finite.
However, our sequences share the structure of an infinite language, allowing us to measure generalization by testing on input sequences that are significantly longer than those seen during training (and are thus out-of-distribution).
Therefore, observing successful generalization in our setting is strong evidence that the network has learned the ``correct'' algorithm.
However, for some tasks where we consider generalization to be successful we still see a slight degradation of the accuracy as the test length increases.
This is a result of implementing finite state machines and memory updates with a neural network trained via SGD: Even slight numerical inaccuracies in state-transitions or memory dynamics might accumulate when increasing the test sequence length. 

Our experiments indicate that RNNs, LSTMs, and Transformers, admittedly powerful architectures, are fairly limited in terms of their ability to generalize to longer inputs.
However, this result must be contrasted with their ability to generalize and detect/extrapolate patterns for fixed size inputs or fixed size context windows.
Transformers, for instance, are capable of learning complex and highly structured generalization patterns, but they cannot overcome the limitation of not having an extendable memory (which only becomes apparent when probed as in our experiments).
This might imply hard limits for scaling laws~\citep{kaplan2020scaling}: Even significantly increasing the amount of training data and the size of a Transformer are insufficient for it to ``climb the Chomsky hiearchy''. 

\section{Conclusion}
\label{sec:conclusion}

We leveraged the theory of computation to better understand how and why neural networks generalize on algorithmic sequence prediction tasks.
Our extensive empirical evaluation demonstrates that there is a model hierarchy on the tasks we investigated, which are representative of the different levels of the Chomsky hierarchy.
In particular, we showed that state-of-the-art architectures, such as LSTMs and Transformers, cannot solve seemingly simple tasks, such as duplicating a string, when evaluated on sequences that are significantly longer than those seen during training.
Moreover, we showed that models interacting with an external memory structure, such as a stack or a finite tape, can climb the Chomsky hierarchy, indicating a promising direction for improvements in architecture design.

\section{Ethics Statement \& Limitations}

Although our results are consistent with prior theoretical analyses~\citep{ackermann2020survey, merrill2019sequential, weiss2018practical}, and suggest that we can group neural architectures according to the Chomsky hierarchy, our claims are limited to our empirical study.
In particular, we cannot guarantee that no tasks exist higher in the hierarchy (\cref{fig:overview}) that an architecture could solve.
Similarly, we cannot guarantee that no tasks exist lower or on the same level of the hierarchy that an architecture cannot solve.
Moreover, without extracting and analyzing the automata implemented by the networks, we cannot strictly claim that our architectures generalize to arbitrary-length inputs, as we only test up to a maximum length.
Finally, our results are \wrt our precise experimental setting, \ie if an architecture fails to generalize we cannot guarantee that no weight configuration would solve the task; we were simply unable to find such a configuration with our training protocol.

\subsubsection*{Acknowledgments}

We thank Ann He, Chris Dyer, Markus Kunesch, R{\'{o}}bert Csord{\'{a}}s, Tom McGrath, and Zhengdong Wang for their helpful feedback and insightful conversations.

\bibliography{references}

\begin{thebibliography}{94}
\providecommand{\natexlab}[1]{#1}
\providecommand{\url}[1]{\texttt{#1}}
\expandafter\ifx\csname urlstyle\endcsname\relax
  \providecommand{\doi}[1]{doi: #1}\else
  \providecommand{\doi}{doi: \begingroup \urlstyle{rm}\Url}\fi

\bibitem[Ackerman \& Cybenko(2020)Ackerman and Cybenko]{ackermann2020survey}
Joshua Ackerman and George Cybenko.
\newblock A survey of neural networks and formal languages.
\newblock \emph{CoRR}, 2020.

\bibitem[Anil et~al.(2022)Anil, Wu, Andreassen, Lewkowycz, Misra, Ramasesh,
  Slone, Gur{-}Ari, Dyer, and Neyshabur]{anil2022exploring}
Cem Anil, Yuhuai Wu, Anders Andreassen, Aitor Lewkowycz, Vedant Misra, Vinay~V.
  Ramasesh, Ambrose Slone, Guy Gur{-}Ari, Ethan Dyer, and Behnam Neyshabur.
\newblock Exploring length generalization in large language models.
\newblock \emph{CoRR}, 2022.

\bibitem[Babuschkin et~al.(2020)Babuschkin, Baumli, Bell, Bhupatiraju, Bruce,
  Buchlovsky, Budden, Cai, Clark, Danihelka, Fantacci, Godwin, Jones, Hennigan,
  Hessel, Kapturowski, Keck, Kemaev, King, Martens, Mikulik, Norman, Quan,
  Papamakarios, Ring, Ruiz, Sanchez, Schneider, Sezener, Spencer, Srinivasan,
  Stokowiec, and Viola]{babuschkin2020deepmind}
Igor Babuschkin, Kate Baumli, Alison Bell, Surya Bhupatiraju, Jake Bruce, Peter
  Buchlovsky, David Budden, Trevor Cai, Aidan Clark, Ivo Danihelka, Claudio
  Fantacci, Jonathan Godwin, Chris Jones, Tom Hennigan, Matteo Hessel, Steven
  Kapturowski, Thomas Keck, Iurii Kemaev, Michael King, Lena Martens, Vladimir
  Mikulik, Tamara Norman, John Quan, George Papamakarios, Roman Ring, Francisco
  Ruiz, Alvaro Sanchez, Rosalia Schneider, Eren Sezener, Stephen Spencer,
  Srivatsan Srinivasan, Wojciech Stokowiec, and Fabio Viola.
\newblock The {D}eep{M}ind {JAX} {E}cosystem, 2020.
\newblock URL \url{http://github.com/deepmind}.

\bibitem[Bahdanau et~al.(2015)Bahdanau, Cho, and Bengio]{bahdanau2015neural}
Dzmitry Bahdanau, Kyunghyun Cho, and Yoshua Bengio.
\newblock Neural machine translation by jointly learning to align and
  translate.
\newblock In \emph{3rd International Conference on Learning Representations},
  2015.

\bibitem[Bansal et~al.(2022)Bansal, Schwarzschild, Borgnia, Emam, Huang,
  Goldblum, and Goldstein]{bansal2022endtoend}
Arpit Bansal, Avi Schwarzschild, Eitan Borgnia, Zeyad Emam, Furong Huang, Micah
  Goldblum, and Tom Goldstein.
\newblock End-to-end algorithm synthesis with recurrent networks: Logical
  extrapolation without overthinking.
\newblock \emph{CoRR}, 2022.

\bibitem[Battaglia et~al.(2018)Battaglia, Hamrick, Bapst, Sanchez{-}Gonzalez,
  Zambaldi, Malinowski, Tacchetti, Raposo, Santoro, Faulkner,
  G{\"{u}}l{\c{c}}ehre, Song, Ballard, Gilmer, Dahl, Vaswani, Allen, Nash,
  Langston, Dyer, Heess, Wierstra, Kohli, Botvinick, Vinyals, Li, and
  Pascanu]{battaglia2018relational}
Peter~W. Battaglia, Jessica~B. Hamrick, Victor Bapst, Alvaro
  Sanchez{-}Gonzalez, Vin{\'{\i}}cius~Flores Zambaldi, Mateusz Malinowski,
  Andrea Tacchetti, David Raposo, Adam Santoro, Ryan Faulkner, {\c{C}}aglar
  G{\"{u}}l{\c{c}}ehre, H.~Francis Song, Andrew~J. Ballard, Justin Gilmer,
  George~E. Dahl, Ashish Vaswani, Kelsey~R. Allen, Charles Nash, Victoria
  Langston, Chris Dyer, Nicolas Heess, Daan Wierstra, Pushmeet Kohli,
  Matthew~M. Botvinick, Oriol Vinyals, Yujia Li, and Razvan Pascanu.
\newblock Relational inductive biases, deep learning, and graph networks.
\newblock \emph{CoRR}, 2018.

\bibitem[Bhattamishra et~al.(2020)Bhattamishra, Ahuja, and
  Goyal]{bhattamishra2020ability}
Satwik Bhattamishra, Kabir Ahuja, and Navin Goyal.
\newblock On the ability and limitations of transformers to recognize formal
  languages.
\newblock In \emph{Proceedings of the 2020 Conference on Empirical Methods in
  Natural Language Processing}, 2020.

\bibitem[Bod{\'{e}}n \& Wiles(2000)Bod{\'{e}}n and Wiles]{boden2000context}
Mikael Bod{\'{e}}n and Janet Wiles.
\newblock Context-free and context-sensitive dynamics in recurrent neural
  networks.
\newblock \emph{Connect. Sci.}, 2000.

\bibitem[Bod{\'{e}}n \& Wiles(2002)Bod{\'{e}}n and Wiles]{boden2002learning}
Mikael Bod{\'{e}}n and Janet Wiles.
\newblock On learning context-free and context-sensitive languages.
\newblock \emph{{IEEE} Trans. Neural Networks}, 2002.

\bibitem[Bradbury et~al.(2018)Bradbury, Frostig, Hawkins, Johnson, Leary,
  Maclaurin, Necula, Paszke, Vander{P}las, Wanderman-{M}ilne, and
  Zhang]{bradbury2018jax}
James Bradbury, Roy Frostig, Peter Hawkins, Matthew~James Johnson, Chris Leary,
  Dougal Maclaurin, George Necula, Adam Paszke, Jake Vander{P}las, Skye
  Wanderman-{M}ilne, and Qiao Zhang.
\newblock {JAX}: composable transformations of {P}ython+{N}um{P}y programs,
  2018.
\newblock URL \url{http://github.com/google/jax}.

\bibitem[Chen et~al.(2018)Chen, Gilroy, Maletti, May, and
  Knight]{chen2018recurrent}
Yining Chen, Sorcha Gilroy, Andreas Maletti, Jonathan May, and Kevin Knight.
\newblock Recurrent neural networks as weighted language recognizers.
\newblock In \emph{Proceedings of the 2018 Conference of the North American
  Chapter of the Association for Computational Linguistics: Human Language
  Technologies}, 2018.

\bibitem[Chiang \& Cholak(2022)Chiang and Cholak]{chiangparitycheck}
David Chiang and Peter Cholak.
\newblock Overcoming a theoretical limitation of self-attention, 2022.
\newblock URL \url{https://arxiv.org/abs/2202.12172}.

\bibitem[Cho et~al.(2014)Cho, van Merrienboer, G{\"{u}}l{\c{c}}ehre, Bahdanau,
  Bougares, Schwenk, and Bengio]{cho2014learning}
Kyunghyun Cho, Bart van Merrienboer, {\c{C}}aglar G{\"{u}}l{\c{c}}ehre, Dzmitry
  Bahdanau, Fethi Bougares, Holger Schwenk, and Yoshua Bengio.
\newblock Learning phrase representations using {RNN} encoder-decoder for
  statistical machine translation.
\newblock In \emph{Proceedings of the 2014 Conference on Empirical Methods in
  Natural Language Processing}, 2014.

\bibitem[Chomsky(1956)]{chomsky1956three}
Noam Chomsky.
\newblock Three models for the description of language.
\newblock \emph{{IRE} Trans. Inf. Theory}, 1956.

\bibitem[Dai et~al.(2019)Dai, Yang, Yang, Carbonell, Le, and
  Salakhutdinov]{dai2019transformer}
Zihang Dai, Zhilin Yang, Yiming Yang, Jaime~G. Carbonell, Quoc~Viet Le, and
  Ruslan Salakhutdinov.
\newblock Transformer-xl: Attentive language models beyond a fixed-length
  context.
\newblock In \emph{Proceedings of the 57th Conference of the Association for
  Computational Linguistics}, 2019.

\bibitem[Danihelka et~al.(2016)Danihelka, Wayne, Uria, Kalchbrenner, and
  Graves]{danihelka2016associative}
Ivo Danihelka, Greg Wayne, Benigno Uria, Nal Kalchbrenner, and Alex Graves.
\newblock Associative long short-term memory.
\newblock In \emph{Proceedings of the 33nd International Conference on Machine
  Learning}, 2016.

\bibitem[Das et~al.(1992{\natexlab{a}})Das, Giles, and Sun]{das1992learning}
Sreerupa Das, C.~Lee Giles, and Guo-Zheng Sun.
\newblock Learning context-free grammars: Capabilities and limitations of a
  recurrent neural network with an external stack memory.
\newblock In \emph{Fourteenth Annual Conference of the Cognitive Science
  Society}, 1992{\natexlab{a}}.

\bibitem[Das et~al.(1992{\natexlab{b}})Das, Giles, and Sun]{das1992using}
Sreerupa Das, C.~Lee Giles, and Guo{-}Zheng Sun.
\newblock Using prior knowledge in a {NNPDA} to learn context-free languages.
\newblock In \emph{Advances in Neural Information Processing Systems 5},
  1992{\natexlab{b}}.

\bibitem[Dawid(1984)]{dawid1984present}
A.~Philip Dawid.
\newblock Present position and potential developments: Some personal views:
  Statistical theory: The prequential approach.
\newblock \emph{Journal of the Royal Statistical Society. Series A (General)},
  1984.

\bibitem[DuSell \& Chiang(2020)DuSell and Chiang]{dusell2020learning}
Brian DuSell and David Chiang.
\newblock Learning context-free languages with nondeterministic stack rnns.
\newblock In \emph{Proceedings of the 24th Conference on Computational Natural
  Language Learning}, 2020.

\bibitem[DuSell \& Chiang(2022)DuSell and Chiang]{dusell2022learning}
Brian DuSell and David Chiang.
\newblock Learning hierarchical structures with differentiable nondeterministic
  stacks.
\newblock In \emph{The Tenth International Conference on Learning
  Representations}, 2022.

\bibitem[Ebrahimi et~al.(2020)Ebrahimi, Gelda, and Zhang]{ebrahimi2020how}
Javid Ebrahimi, Dhruv Gelda, and Wei Zhang.
\newblock How can self-attention networks recognize dyck-n languages?
\newblock In \emph{Findings of the Association for Computational Linguistics},
  2020.

\bibitem[Elman(1990)]{elman1990finding}
Jeffrey~L. Elman.
\newblock Finding structure in time.
\newblock \emph{Cogn. Sci.}, 1990.

\bibitem[Freivalds \& Liepins(2017)Freivalds and
  Liepins]{freivalds2017improving}
Karlis Freivalds and Renars Liepins.
\newblock Improving the neural {GPU} architecture for algorithm learning.
\newblock \emph{CoRR}, 2017.

\bibitem[Gers \& Schmidhuber(2001)Gers and Schmidhuber]{gers2001lstm}
Felix~A. Gers and J{\"{u}}rgen Schmidhuber.
\newblock {LSTM} recurrent networks learn simple context-free and
  context-sensitive languages.
\newblock \emph{{IEEE} Trans. Neural Networks}, 2001.

\bibitem[Giles et~al.(1992)Giles, Miller, Chen, Chen, Sun, and
  Lee]{giles1992learning}
C.~Lee Giles, Clifford~B. Miller, Dong Chen, Hsing{-}Hen Chen, Guo{-}Zheng Sun,
  and Yee{-}Chun Lee.
\newblock Learning and extracting finite state automata with second-order
  recurrent neural networks.
\newblock \emph{Neural Comput.}, 1992.

\bibitem[Goldberg(1989)]{goldberg1989genetic}
David~E. Goldberg.
\newblock \emph{Genetic Algorithms in Search Optimization and Machine
  Learning}.
\newblock Addison-Wesley, 1989.

\bibitem[Gomez et~al.(2008)Gomez, Schmidhuber, and
  Miikkulainen]{gomez2008accelerated}
Faustino~J. Gomez, J{\"{u}}rgen Schmidhuber, and Risto Miikkulainen.
\newblock Accelerated neural evolution through cooperatively coevolved
  synapses.
\newblock \emph{J. Mach. Learn. Res.}, 2008.

\bibitem[Graves(2016)]{graves2016adaptive}
Alex Graves.
\newblock Adaptive computation time for recurrent neural networks.
\newblock \emph{CoRR}, 2016.

\bibitem[Graves et~al.(2014)Graves, Wayne, and Danihelka]{graves2014neural}
Alex Graves, Greg Wayne, and Ivo Danihelka.
\newblock Neural turing machines.
\newblock \emph{CoRR}, 2014.

\bibitem[Graves et~al.(2016)Graves, Wayne, Reynolds, Harley, Danihelka,
  Grabska{-}Barwinska, Colmenarejo, Grefenstette, Ramalho, Agapiou, Badia,
  Hermann, Zwols, Ostrovski, Cain, King, Summerfield, Blunsom, Kavukcuoglu, and
  Hassabis]{graves2016hybrid}
Alex Graves, Greg Wayne, Malcolm Reynolds, Tim Harley, Ivo Danihelka, Agnieszka
  Grabska{-}Barwinska, Sergio~Gomez Colmenarejo, Edward Grefenstette, Tiago
  Ramalho, John~P. Agapiou, Adri{\`{a}}~Puigdom{\`{e}}nech Badia, Karl~Moritz
  Hermann, Yori Zwols, Georg Ostrovski, Adam Cain, Helen King, Christopher
  Summerfield, Phil Blunsom, Koray Kavukcuoglu, and Demis Hassabis.
\newblock Hybrid computing using a neural network with dynamic external memory.
\newblock \emph{Nat.}, 2016.

\bibitem[Grefenstette et~al.(2015)Grefenstette, Hermann, Suleyman, and
  Blunsom]{grefenstette2015learning}
Edward Grefenstette, Karl~Moritz Hermann, Mustafa Suleyman, and Phil Blunsom.
\newblock Learning to transduce with unbounded memory.
\newblock In \emph{Advances in Neural Information Processing Systems 28}, 2015.

\bibitem[G{\"{u}}l{\c{c}}ehre et~al.(2018)G{\"{u}}l{\c{c}}ehre, Chandar, Cho,
  and Bengio]{gulcehre2018dynamic}
{\c{C}}aglar G{\"{u}}l{\c{c}}ehre, Sarath Chandar, Kyunghyun Cho, and Yoshua
  Bengio.
\newblock Dynamic neural turing machine with continuous and discrete addressing
  schemes.
\newblock \emph{Neural Comput.}, 2018.

\bibitem[Hahn(2020)]{hahn2020theoretical}
Michael Hahn.
\newblock Theoretical limitations of self-attention in neural sequence models.
\newblock \emph{Trans. Assoc. Comput. Linguistics}, 2020.

\bibitem[Hao et~al.(2018)Hao, Merrill, Angluin, Frank, Amsel, Benz, and
  Mendelsohn]{hao2018context}
Yiding Hao, William Merrill, Dana Angluin, Robert Frank, Noah Amsel, Andrew
  Benz, and Simon Mendelsohn.
\newblock Context-free transductions with neural stacks.
\newblock In \emph{Proceedings of the Workshop: Analyzing and Interpreting
  Neural Networks for NLP}, 2018.

\bibitem[Hao et~al.(2022)Hao, Angluin, and Frank]{hao2022formal}
Yiding Hao, Dana Angluin, and Robert Frank.
\newblock Formal language recognition by hard attention transformers:
  Perspectives from circuit complexity.
\newblock \emph{CoRR}, 2022.

\bibitem[Hennigan et~al.(2020)Hennigan, Cai, Norman, and
  Babuschkin]{hennigan2020haiku}
Tom Hennigan, Trevor Cai, Tamara Norman, and Igor Babuschkin.
\newblock {H}aiku: {S}onnet for {JAX}, 2020.
\newblock URL \url{http://github.com/deepmind/dm-haiku}.

\bibitem[Hessel et~al.(2020)Hessel, Budden, Viola, Rosca, Sezener, and
  Hennigan]{hessel2020optax}
Matteo Hessel, David Budden, Fabio Viola, Mihaela Rosca, Eren Sezener, and Tom
  Hennigan.
\newblock Optax: composable gradient transformation and optimisation, in jax!,
  2020.
\newblock URL \url{http://github.com/deepmind/optax}.

\bibitem[Hochreiter \& Schmidhuber(1997)Hochreiter and
  Schmidhuber]{hochreiter1997long}
Sepp Hochreiter and J{\"{u}}rgen Schmidhuber.
\newblock Long short-term memory.
\newblock \emph{Neural Comput.}, 1997.

\bibitem[Holland(1992)]{holland992adaptation}
John~H. Holland.
\newblock \emph{Adaptation in Natural and Artificial Systems: An Introductory
  Analysis with Applications to Biology, Control, and Artificial Intelligence}.
\newblock {MIT} Press, 1992.

\bibitem[H{\"{o}}lldobler et~al.(1997)H{\"{o}}lldobler, Kalinke, and
  Lehmann]{holldobler1997designing}
Steffen H{\"{o}}lldobler, Yvonne Kalinke, and Helko Lehmann.
\newblock Designing a counter: Another case study of dynamics and activation
  landscapes in recurrent networks.
\newblock In \emph{Advances in Artificial Intelligence, 21st Annual German
  Conference on Artificial Intelligence}, 1997.

\bibitem[Joulin \& Mikolov(2015)Joulin and Mikolov]{joulin2015inferring}
Armand Joulin and Tom{\'{a}}s Mikolov.
\newblock Inferring algorithmic patterns with stack-augmented recurrent nets.
\newblock In \emph{Advances in Neural Information Processing Systems 28}, 2015.

\bibitem[Kaiser \& Sutskever(2016)Kaiser and Sutskever]{kaiser2015neural}
Lukasz Kaiser and Ilya Sutskever.
\newblock Neural gpus learn algorithms.
\newblock In \emph{4th International Conference on Learning Representations},
  2016.

\bibitem[Kalchbrenner et~al.(2016)Kalchbrenner, Danihelka, and
  Graves]{kalchbrenner2015grid}
Nal Kalchbrenner, Ivo Danihelka, and Alex Graves.
\newblock Grid long short-term memory.
\newblock In \emph{4th International Conference on Learning Representations},
  2016.

\bibitem[Kaplan et~al.(2020)Kaplan, McCandlish, Henighan, Brown, Chess, Child,
  Gray, Radford, Wu, and Amodei]{kaplan2020scaling}
Jared Kaplan, Sam McCandlish, Tom Henighan, Tom~B. Brown, Benjamin Chess, Rewon
  Child, Scott Gray, Alec Radford, Jeffrey Wu, and Dario Amodei.
\newblock Scaling laws for neural language models.
\newblock \emph{CoRR}, 2020.

\bibitem[Kingma \& Ba(2015)Kingma and Ba]{kingma2015adam}
Diederik~P. Kingma and Jimmy Ba.
\newblock Adam: {A} method for stochastic optimization.
\newblock In \emph{3rd International Conference on Learning Representations},
  2015.

\bibitem[Korsky \& Berwick(2019)Korsky and Berwick]{korsky2019computational}
Samuel~A. Korsky and Robert~C. Berwick.
\newblock On the computational power of rnns.
\newblock \emph{CoRR}, 2019.

\bibitem[Kurach et~al.(2016)Kurach, Andrychowicz, and
  Sutskever]{kurach2016neural}
Karol Kurach, Marcin Andrychowicz, and Ilya Sutskever.
\newblock Neural random-access machines.
\newblock In \emph{4th International Conference on Learning Representations},
  2016.

\bibitem[Liang et~al.(2013)Liang, Jordan, and Klein]{liang2013learning}
Percy Liang, Michael~I. Jordan, and Dan Klein.
\newblock Learning dependency-based compositional semantics.
\newblock \emph{Comput. Linguistics}, 2013.

\bibitem[Mali et~al.(2021)Mali, Ororbia, Kifer, and Giles]{mali2021recognizing}
Ankur~Arjun Mali, Alexander Ororbia, Daniel Kifer, and C.~Lee Giles.
\newblock Recognizing long grammatical sequences using recurrent networks
  augmented with an external differentiable stack.
\newblock In \emph{Proceedings of the 15th International Conference on
  Grammatical Inference}, 2021.

\bibitem[Merrill(2019)]{merrill2019sequential}
William Merrill.
\newblock Sequential neural networks as automata.
\newblock \emph{CoRR}, 2019.

\bibitem[Merrill \& Sabharwal(2022)Merrill and
  Sabharwal]{merrill2022logprecision}
William Merrill and Ashish Sabharwal.
\newblock Log-precision transformers are constant-depth uniform threshold
  circuits.
\newblock \emph{CoRR}, 2022.

\bibitem[Merrill et~al.(2020)Merrill, Weiss, Goldberg, Schwartz, Smith, and
  Yahav]{merrill2020formal}
William Merrill, Gail Weiss, Yoav Goldberg, Roy Schwartz, Noah~A. Smith, and
  Eran Yahav.
\newblock A formal hierarchy of {RNN} architectures.
\newblock In \emph{Proceedings of the 58th Annual Meeting of the Association
  for Computational Linguistics}, 2020.

\bibitem[Minsky(1967)]{minsky1967computation}
Marvin~L. Minsky.
\newblock \emph{Computation: Finite and Infinite Machines}.
\newblock Prentice-Hall, Inc., 1967.

\bibitem[Mitchell(1980)]{mitchell1980need}
Tom~M. Mitchell.
\newblock The need for biases in learning generalizations.
\newblock Technical report, Rutgers University, 1980.

\bibitem[Mozer \& Das(1992)Mozer and Das]{mozer1992connectionist}
Michael Mozer and Sreerupa Das.
\newblock A connectionist symbol manipulator that discovers the structure of
  context-free languages.
\newblock In \emph{Advances in Neural Information Processing Systems 5}, 1992.

\bibitem[Nordin(1997)]{nording1997evolutionary}
Peter Nordin.
\newblock \emph{Evolutionary program induction of binary machine code and its
  applications}.
\newblock PhD thesis, Dortmund University of Technology, 1997.

\bibitem[P{\'{e}}rez et~al.(2019)P{\'{e}}rez, Marinkovic, and
  Barcel{\'{o}}]{perez2019turing}
Jorge P{\'{e}}rez, Javier Marinkovic, and Pablo Barcel{\'{o}}.
\newblock On the turing completeness of modern neural network architectures.
\newblock In \emph{7th International Conference on Learning Representations},
  2019.

\bibitem[P{\'{e}}rez et~al.(2021)P{\'{e}}rez, Barcel{\'{o}}, and
  Marinkovic]{perez2021attention}
Jorge P{\'{e}}rez, Pablo Barcel{\'{o}}, and Javier Marinkovic.
\newblock Attention is turing-complete.
\newblock \emph{J. Mach. Learn. Res.}, 2021.

\bibitem[Pollack(1991)]{pollack1991induction}
Jordan~B. Pollack.
\newblock The induction of dynamical recognizers.
\newblock \emph{Mach. Learn.}, 1991.

\bibitem[Power et~al.(2022)Power, Burda, Edwards, Babuschkin, and
  Misra]{power2022grokking}
Alethea Power, Yuri Burda, Harrison Edwards, Igor Babuschkin, and Vedant Misra.
\newblock Grokking: Generalization beyond overfitting on small algorithmic
  datasets.
\newblock \emph{CoRR}, 2022.

\bibitem[Press et~al.(2021)Press, Smith, and Lewis]{press2021train}
Ofir Press, Noah~A. Smith, and Mike Lewis.
\newblock Train short, test long: Attention with linear biases enables input
  length extrapolation.
\newblock \emph{CoRR}, 2021.

\bibitem[Price et~al.(2016)Price, Zaremba, and Sutskever]{price2016extensions}
Eric Price, Wojciech Zaremba, and Ilya Sutskever.
\newblock Extensions and limitations of the neural {GPU}.
\newblock \emph{CoRR}, 2016.

\bibitem[Rich(2007)]{rich2007automata}
Elaine Rich.
\newblock \emph{Automata, Computability and Complexity}.
\newblock Prentice-Hall, 2007.

\bibitem[Rodriguez \& Wiles(1997)Rodriguez and Wiles]{rodriguez1997recurrent}
Paul Rodriguez and Janet Wiles.
\newblock Recurrent neural networks can learn to implement symbol-sensitive
  counting.
\newblock In \emph{Advances in Neural Information Processing Systems 10}, 1997.

\bibitem[Savage(1998)]{savage1998models}
John~E. Savage.
\newblock \emph{Models of computation - exploring the power of computing}.
\newblock Addison-Wesley, 1998.

\bibitem[Sennhauser \& Berwick(2018)Sennhauser and
  Berwick]{sennhauser2018evaluating}
Luzi Sennhauser and Robert~C. Berwick.
\newblock Evaluating the ability of lstms to learn context-free grammars.
\newblock In \emph{Proceedings of the Workshop: Analyzing and Interpreting
  Neural Networks for NLP}, 2018.

\bibitem[Siegelmann \& Sontag(1994)Siegelmann and Sontag]{siegelmann1994analog}
Hava~T. Siegelmann and Eduardo~D. Sontag.
\newblock Analog computation via neural networks.
\newblock \emph{Theor. Comput. Sci.}, 1994.

\bibitem[Sipser(1997)]{sipser1997introduction}
Michael Sipser.
\newblock \emph{Introduction to the theory of computation}.
\newblock {PWS} Publishing Company, 1997.

\bibitem[Skachkova et~al.(2018)Skachkova, Trost, and
  Klakow]{skachkova2018closing}
Natalia Skachkova, Thomas~Alexander Trost, and Dietrich Klakow.
\newblock Closing brackets with recurrent neural networks.
\newblock In \emph{Proceedings of the Workshop: Analyzing and Interpreting
  Neural Networks for NLP}, 2018.

\bibitem[Solomonoff(1964{\natexlab{a}})]{solomonoff1964formalI}
Ray~J. Solomonoff.
\newblock A formal theory of inductive inference. part {I}.
\newblock \emph{Inf. Control.}, 1964{\natexlab{a}}.

\bibitem[Solomonoff(1964{\natexlab{b}})]{solomonoff1964formalII}
Ray~J. Solomonoff.
\newblock A formal theory of inductive inference. part {II}.
\newblock \emph{Inf. Control.}, 1964{\natexlab{b}}.

\bibitem[Solomonoff(2009)]{solomonoff2009algorithmic}
Ray~J. Solomonoff.
\newblock Algorithmic probability: Theory and applications.
\newblock In \emph{Information Theory and Statistical Learning}. Springer US,
  2009.

\bibitem[Solomonoff(2010)]{solomonoff2010algorithmic}
Ray~J. Solomonoff.
\newblock Algorithmic probability, heuristic programming and agi.
\newblock In \emph{Proceedings of the 3d Conference on Artificial General
  Intelligence}, 2010.

\bibitem[Steijvers \& Gr{\"{u}}nwald(1996)Steijvers and
  Gr{\"{u}}nwald]{steijvers1996recurrent}
Mark Steijvers and Peter Gr{\"{u}}nwald.
\newblock A recurrent network that performs a context-sensitive prediction
  task.
\newblock In \emph{Eighteenth Annual Conference of the Cognitive Science
  Society}, 1996.

\bibitem[Stogin et~al.(2020)Stogin, Mali, and Giles]{stogin2020provably}
John Stogin, Ankur~Arjun Mali, and C.~Lee Giles.
\newblock Provably stable interpretable encodings of context free grammars in
  rnns with a differentiable stack.
\newblock \emph{CoRR}, 2020.

\bibitem[Su et~al.(2021)Su, Lu, Pan, Wen, and Liu]{su2021roformer}
Jianlin Su, Yu~Lu, Shengfeng Pan, Bo~Wen, and Yunfeng Liu.
\newblock Roformer: Enhanced transformer with rotary position embedding, 2021.

\bibitem[Sukhbaatar et~al.(2015)Sukhbaatar, Szlam, Weston, and
  Fergus]{sukhbaatar2015end}
Sainbayar Sukhbaatar, Arthur Szlam, Jason Weston, and Rob Fergus.
\newblock End-to-end memory networks.
\newblock In \emph{Advances in Neural Information Processing Systems 28}, 2015.

\bibitem[Sun et~al.(1993)Sun, Giles, Chen, and Lee]{sun1993neural}
G.~Z. Sun, C.~L. Giles, H.~H. Chen, and Y.~C. Lee.
\newblock The neural network pushdown automaton: model, stack and learning
  simulations.
\newblock Technical report, University of Maryland, 1993.

\bibitem[Sutskever et~al.(2014)Sutskever, Vinyals, and
  Le]{sutskever2014sequence}
Ilya Sutskever, Oriol Vinyals, and Quoc~V. Le.
\newblock Sequence to sequence learning with neural networks.
\newblock In \emph{Advances in Neural Information Processing Systems 27}, 2014.

\bibitem[Suzgun et~al.(2019{\natexlab{a}})Suzgun, Gehrmann, Belinkov, and
  Shieber]{suzgun2019lstm}
Mirac Suzgun, Sebastian Gehrmann, Yonatan Belinkov, and Stuart~M. Shieber.
\newblock {LSTM} networks can perform dynamic counting.
\newblock \emph{CoRR}, 2019{\natexlab{a}}.

\bibitem[Suzgun et~al.(2019{\natexlab{b}})Suzgun, Gehrmann, Belinkov, and
  Shieber]{suzgun2019memory}
Mirac Suzgun, Sebastian Gehrmann, Yonatan Belinkov, and Stuart~M. Shieber.
\newblock Memory-augmented recurrent neural networks can learn generalized dyck
  languages.
\newblock \emph{CoRR}, 2019{\natexlab{b}}.

\bibitem[Vapnik(1998)]{vapnik1998statistical}
Vladimir Vapnik.
\newblock \emph{Statistical learning theory}.
\newblock Wiley, 1998.

\bibitem[Vaswani et~al.(2017)Vaswani, Shazeer, Parmar, Uszkoreit, Jones, Gomez,
  Kaiser, and Polosukhin]{vaswani2017attention}
Ashish Vaswani, Noam Shazeer, Niki Parmar, Jakob Uszkoreit, Llion Jones,
  Aidan~N. Gomez, Lukasz Kaiser, and Illia Polosukhin.
\newblock Attention is all you need.
\newblock In \emph{Advances in Neural Information Processing Systems 30}, 2017.

\bibitem[Weiss et~al.(2018)Weiss, Goldberg, and Yahav]{weiss2018practical}
Gail Weiss, Yoav Goldberg, and Eran Yahav.
\newblock On the practical computational power of finite precision rnns for
  language recognition.
\newblock In \emph{Proceedings of the 56th Annual Meeting of the Association
  for Computational Linguistics}, 2018.

\bibitem[Weiss et~al.(2021)Weiss, Goldberg, and Yahav]{weiss2021thinking}
Gail Weiss, Yoav Goldberg, and Eran Yahav.
\newblock Thinking like transformers.
\newblock In \emph{Proceedings of the 38th International Conference on Machine
  Learning}, 2021.

\bibitem[Weston et~al.(2015)Weston, Chopra, and Bordes]{weston2014memory}
Jason Weston, Sumit Chopra, and Antoine Bordes.
\newblock Memory networks.
\newblock In \emph{3rd International Conference on Learning Representations},
  2015.

\bibitem[Wiles \& Elman(1995)Wiles and Elman]{wiles1995learning}
Janet Wiles and Jeff Elman.
\newblock Learning to count without a counter: A case study of dynamics and
  activation landscapes in recurrent networks.
\newblock In \emph{Seventeenth Annual Conference of the Cognitive Science
  Society}, 1995.

\bibitem[Wineberg \& Oppacher(1994)Wineberg and
  Oppacher]{wineberg1994representation}
Mark Wineberg and Franz Oppacher.
\newblock A representation scheme to perform program induction in a canonical
  genetic algorithm.
\newblock In \emph{Parallel Problem Solving from Nature - {PPSN} III,
  International Conference on Evolutionary Computation}, 1994.

\bibitem[Yang \& Rush(2017)Yang and Rush]{yang2017lie}
Greg Yang and Alexander~M. Rush.
\newblock Lie-access neural turing machines.
\newblock In \emph{5th International Conference on Learning Representations},
  2017.

\bibitem[Yogatama et~al.(2018)Yogatama, Miao, Melis, Ling, Kuncoro, Dyer, and
  Blunsom]{yogatama2018memory}
Dani Yogatama, Yishu Miao, G{\'{a}}bor Melis, Wang Ling, Adhiguna Kuncoro,
  Chris Dyer, and Phil Blunsom.
\newblock Memory architectures in recurrent neural network language models.
\newblock In \emph{6th International Conference on Learning Representations},
  2018.

\bibitem[Zaremba \& Sutskever(2015)Zaremba and
  Sutskever]{zaremba2015reinforcement}
Wojciech Zaremba and Ilya Sutskever.
\newblock Reinforcement learning neural turing machines.
\newblock \emph{CoRR}, 2015.

\bibitem[Zaremba et~al.(2016)Zaremba, Mikolov, Joulin, and
  Fergus]{zaremba2016learning}
Wojciech Zaremba, Tom{\'{a}}s Mikolov, Armand Joulin, and Rob Fergus.
\newblock Learning simple algorithms from examples.
\newblock In \emph{Proceedings of the 33nd International Conference on Machine
  Learning}, 2016.

\bibitem[Zhang et~al.(2022)Zhang, Backurs, Bubeck, Eldan, Gunasekar, and
  Wagner]{zhang2022unveiling}
Yi~Zhang, Arturs Backurs, S{\'{e}}bastien Bubeck, Ronen Eldan, Suriya
  Gunasekar, and Tal Wagner.
\newblock Unveiling transformers with {LEGO:} a synthetic reasoning task.
\newblock \emph{CoRR}, 2022.

\end{thebibliography}
\bibliographystyle{iclr2023_conference}

\clearpage

\appendix

\counterwithin{figure}{section}
\counterwithin{table}{section}
\counterwithin{algocf}{section}

\section{Experimental details}
\label{sec:experimental-details}

\subsection{Language recognition \versus transduction}
\label{sec:experimental-details:recognition-vs-transduction}

In practice, learning to recognize a formal language by training a classifier on words and non-words of a language is hard because there is, in general, no formal construction for generating a sufficient but finite set of negative examples.
Thus, prior work has relied on other approaches to evaluate language recognition capabilities.
These approaches are variants of language generation, which are as hard as their recognition counterpart (for a given language) and thus are good proxy tasks.

The first approach is autoregressive token prediction (standard in the machine learning community~\citep{bahdanau2015neural,vaswani2017attention}): Given a prefix $p$ of a word $w \in L$, predict the probability of its next token $t \in \Sigma$.
The main problem with this approach is that the probabilities depend on the length distribution of the training strings.
For instance, for the language of palindromes of length 2, with alphabet $\Sigma = \{a, b\}$, the output probabilities are $P(a | \mathrm{prefix} = a) = 1$ and $P(b | \mathrm{prefix} = b) = 1$.
However, for palindromes of length 3, the probabilities are very different, \ie $P(a | \mathrm{prefix} = a) = 0.5$ and $P(b | \mathrm{prefix} = b) = 0.5$.
Therefore, there is a conflict between the different distributions, and no model can solve the task without knowing the length beforehand.
Accordingly, increasing the length at test time without informing the model makes it impossible to find the correct new probabilities and, therefore, to generalize to inputs of unseen lengths.

The second approach is to predict the set of possible next tokens and not their probabilities~\citep{bhattamishra2020ability, suzgun2019lstm}.
In this case, the distribution of possible next tokens does not change with the length, and increasing the length at test time is possible.
However, for each prefix, the target will consists of a set of tokens, which is incompatible with the (standard sequence) prediction setting and thus of lesser interest to the wider machine learning community.

Our approach is similar to the second one, but we only consider prefixes for which there is only one possible (deterministic) continuation.
Thus, instead of language generation, we consider language transduction: The inputs and outputs form two individual languages, and the goal is to learn the deterministic function that maps a word from one to the other.
The automata solving these tasks are the classical automata from \cref{fig:overview} but with outputs attached to each transition and are denoted transducers.
Therefore, the hierarchy that we consider is a hierarchy over transducers and not over acceptors as detailed in \cref{sec:background}.
However, the two types of machines (acceptor automata and transducer automata) are interchangeable for the purposes of our investigation as they are both memory-augmented finite-state automata that reside on different levels of the Chomsky hierarchy depending on their type of external memory access (none, a stack, or a tape).

\subsection{Models}
\label{sec:experimental-details:models}

Here we describe our neural architectures in more detail.
All models have a final linear readout layer to produce the output logits.
We do not use any regularization, \eg weight decay, dropout (except for the Transformer models), etc.
For training of the memory-augmented models, we set the tape size to $256$ and the stack size to $128$ and manually increase their sizes when testing on longer sequences.
Thus, as these memory sizes are significantly larger than the training range $N$ (see \cref{sec:methods}), the finite memory models appear infinite for our neural architectures in the context of our experimental evaluation.
Moreover, increasing the memory size does not affect how the RNN controller takes actions, since it only has access to a local view of the external memory (\eg the top of the stack or the current memory cell).

\paragraph{RNN}

A vanilla single-layer RNN~\citep{elman1990finding} with hidden size of 256.

\paragraph{Stack-RNN}

A single-layer RNN controller of hidden size 256 with access to a differentiable stack~\citep{joulin2015inferring}.
The controller can perform any linear combination of \push{}, \pop{}, and \noop{} on the stack, with action weights given by a softmax over a linear readout of the RNN output. Each cell of the stack contains a real vector of dimension 8.

\paragraph{NDStack-RNN}

A single-layer RNN controller of hidden size 256 with access to a differentiable nondeterministic stack.
We consider the NDStack-RNN proposed by~\citep{dusell2020learning, dusell2022learning}, which simulates a nondeterministic stack via dynamic programming.
Concretely, we use normalized actions and nondeterministic state reading with 2 states and sweep over ${2, 4}$ symbols.
We did not use the unnormalized actions as suggested in~\citep{dusell2022learning}.

\paragraph{Tape-RNN}

A single-layer RNN controller of hidden size 256 with access to a differentiable tape, inspired by the Baby-NTM architecture~\citep{suzgun2019memory}.
The controller can perform any linear combination of \writeright{}, \writeleft{}, \writestay{}, \jumpleft{}, and \jumpright{} on the tape, with action weights given by a softmax.
The actions correspond to: writing at the current position and moving to the right (\writeright{}), writing at the current position and moving to the left (\writeleft{}), writing at the current position (\writestay{}), jumping $\ell$ steps to the right without writing (\jumpright{}), where $\ell$ is the length of the input, and jumping $\ell$ steps to the left without writing (\jumpleft{}).
Finally, to allow the Tape-RNN to perform memory operations without having to produce the output, we append ``computation'' tokens (different from the empty tokens $\varnothing$ used to produce the output, see \cref{sec:experimental-details:problem-setup} below) to the input sequence.
As in the Stack-RNN, each cell of the tape contains a real vector of dimension 8.

\paragraph{LSTM}

A vanilla single-layer LSTM~\cite{hochreiter1997long} of hidden size 256.

\paragraph{Stack-LSTM}

A single-layer LSTM controller of hidden size 256 with access to a differentiable stack~\cite{joulin2015inferring}.
The only difference to the Stack-RNN is that this model uses an LSTM as the controller.

\paragraph{Transformer encoder}

A vanilla Transformer encoder~\citep{vaswani2017attention}.
We do not use the autoregressive sequence-to-sequence model to produce the output but attend the whole input (including the empty tokens $\varnothing$).
We use five blocks with $d_{model} = 64$, where each block is composed of an attention layer, two dense layers, and a layer normalization.
We add a residual connections as in the original architecture~\citep{vaswani2017attention}.
We consider five different positional encodings: none, classical sin/cos~\citep{vaswani2017attention}, RoPE~\citep{su2021roformer}, ALiBi~\citep{press2021train}, and the relative positional encoding from Transformer-XL~\citep{dai2019transformer}.

\paragraph{Transformer (autoregressive)}

A vanilla Transformer~\citep{vaswani2017attention}.
We do not use empty tokens $\varnothing$ anymore, and the output is produced via the decoder (using causal masking).
We use the same hyperparameters and positional encodings as for the Transformer encoder above.

\paragraph{Convolutional Neural Network}

A vanilla CNN, consisting of 5 convolutional layers with 32 channels each and a final MLP that produces the task output(s).
We compute 2D convolutions over the time and (one-hot encoding) channel dimension.
That is, the first kernel will consider the entire one-hot encoding at once for every time step.
For all other kernel dimensions (time and filters), we sweep over the values $\{3, 5, 7, 11\}$.

\subsection{Tasks}

Here, we describe our tasks (listed in \cref{tab:tasks}) in more detail.
As mentioned in \cref{sec:methods}, all tasks consist of an input sequence $\bm{x} \in L_I$ from which the models have to predict the target sequence $\bm{y} \in L_O$.
We append $|\bm{y}|$ empty tokens $\varnothing$ to the input sequence, such that the models can process the entire input sequence before having to produce the output (\ie they know the input length before having to produce the output).
Note that half of our tasks have an output length of 1 and are thus just standard classification tasks with the set of classes being the alphabet of the output language. 
The networks still see one empty token $\varnothing$ at the end of the input string to signify that the input phase has finished and that the output must be computed.
We do not consider \gls*{ndcf} tasks since the distinction between \gls*{dcf} and \gls*{ndcf} is fairly technical and does not fundamentally change the type of memory structure required (\ie a stack).

\paragraph{\evenpairs{} (\gls*{regular})}

Given a binary sequence, \eg $\bm{x} = aabba$, compute if the number of $ab$s and $ba$s is even.
For $aabba$, we have one $ab$ and $ba$, meaning that the total number is 2 and thus even, \ie $\bm{y} = b$ (where we arbitrarily equate odd with $a$ and even with $b$).
This task is equivalent to computing whether the first and last character of the string are equal.
Therefore, the task is regular since we can solve it with a 2-state finite-state machine that looks at the first bit of the sequence and compares it with the last one.

\paragraph{\modulararithmeticsimple{} (\gls*{regular})}

Given a sequence of numbers in $\{0, 1, 2, 3, 4\}$ and operations in $\{+, -, \cdot\}$, compute the result modulo 5.
For example, $\bm{x} = 1 + 2 - 4$ evaluates to $\bm{y} = 4$.
Note that the length of the input sequences must be of odd.
Therefore, if we sample an even length during the training/evaluation process, we return an input sequence of this length plus 1.
This task is regular since we can solve it with a 5-state finite-state machine, which transitions for each new operation with the state being equal to the current result.

\paragraph{\paritycheck{} (\gls*{regular})}

Given a binary string, \eg $aaabba$, compute if the number of $b$s is even.
The sequence $\bm{x} = aaabba$ contains 2 $b$s, which is even, \ie $\bm{y} = b$ (where we arbitrarily equate odd with $a$ and even with $b$).
This task is regular since we can solve it with a 2-state finite-state machine, which transitions every time it sees a token different from its current state.

\paragraph{\cyclenavigation{} (\gls*{regular})}

Given a sequence of movements on a cycle of length 5, compute the end position.
The movements are \textsc{STAY}, \textsc{INCREASE}, \textsc{DECREASE} and are represented as $\{0, 1, 2\}$.
The agent always starts at position 0.
For example, $010211$ means the agent stops at position $0 + 0 + 1 + 0 - 1 + 1 + 1$ = $2$, and the output class is $2$.
This task is regular since we can solve it with a 5-state finite-state machine, similar to \modulararithmeticsimple{} (\gls*{regular}), as the task can be converted into a sum modulo 5.

\paragraph{\modulararithmeticbrackets{} (\gls*{dcf})}

Given a sequence of numbers in $\{0, 1, 2, 3, 4\}$, brackets, and operations in $\{+, -, \cdot\}$, compute the result modulo 5.
For example, the sequence $\bm{x} = -(1 - 2) \cdot (4 - 3 \cdot (-2))$ evaluates to $\bm{y} = 0$.
Note that we do not have the same problem of odd lengths as in \modulararithmeticsimple{} (\gls*{regular}), since we can use substrings of the form $(-z)$ with $z \in \{0, 1, 2, 3, 4\}$, which have length 4 and thus can be used to produce input strings of even length.
The task is \gls*{dcf} since we can solve it with a 5-state finite-state machine and an external stack.
That is, similar to the \modulararithmeticsimple{} (\gls*{regular}) task, the finite-state machine transitions for each new operation, with the state being equal to the current result.
Moreover, the stack is used to handle brackets: The machine pushes if the current input token is an opening bracket, pops if it is a closing bracket, and uses no-ops to let the finite-state machine work otherwise.

\paragraph{\reversestring{} (\gls*{dcf})}

Given a binary string, \eg $\bm{x} = aabba$, compute its reverse, \ie $\bm{y} = abbaa$.
The task is \gls*{dcf} since we can solve it by pushing the tokens of the input sequence on a stack until the first empty token $\varnothing$ appears and then popping the tokens from the stack, outputting the values one by one.
This task cannot be solved with a finite-state machine since the number of possible outputs is infinite.

\paragraph{\solveequation{} (\gls*{dcf})}

Given an equation consisting of numbers in $\{0, 1, 2, 3, 4\}$, brackets, operations in $\{+, -, \cdot\}$, and an unknown variable $z$, compute the value of $z$ such that the equation holds modulo 5.
For example, $\bm{x} = -(z - 2) \cdot (4 - 3 \cdot (-2)) = 0$ modulo 5 holds for $z = 1$, \ie $\bm{y} = 1$.
This task is \gls*{dcf} since we can solve it by simulating all possible values of $x$ in a nondeterministic finite-state machine and producing each result using the same algorithm as in \modulararithmeticbrackets{} (\gls*{dcf}).
Note that a nondeterministic finite-state machine can be simulated with a deterministic one, implying that this task is \gls*{dcf} and not \gls*{ndcf}.

\paragraph{\stackmanipulation{} (\gls*{dcf})}

Given a binary string representing a stack's content (given bottom-to-top) and a sequence of \pusha{}, \pushb{}, or \pop{} actions, execute the actions on the stack and return the final stack content (in a top-to-bottom manner, \ie as if popping the resulting stack).
For example, executing the sequence \pop{} \pusha{} \pop{} on the stack $abbaa$, \ie $\bm{x} = abbaa~\pop{}~\pusha{}~\pop{}$, results in $\bm{y} = abba$.
If a \pop{} action is called on an empty stack, the action is ignored.
This task requires stack manipulation by design and is therefore \gls*{dcf}.

\paragraph{\binaryaddition{} (\gls*{cs})}

Given two binary numbers, \eg $10010$ and  $101$, compute their sum in base 2, \ie $\bm{y} = 10111$ for $\bm{x} = 10010 + 101$.
The inputs and outputs are given in little-endian representation.
The task is \gls*{cs} since it exhibits cross-serial dependencies because the two numbers are provided one after the other (and not bit-by-bit at the same time).

\paragraph{\binarymultiplication{} (\gls*{cs})}

Given two binary numbers, \eg $100$ and  $10110$, compute their product in base 2, \ie $\bm{y} = 1011000$ for $\bm{x} = 100 * 10110$.
The inputs and outputs are given in little-endian representation.
The task is \gls*{cs} since it exhibits cross-serial dependencies because the two numbers are provided one after the other (and not bit-by-bit at the same time).

\paragraph{\computesqrt{} (\gls*{cs})}

Given a binary number, \eg $101001$, compute the floor of its square root, \ie $\bm{y} = \lfloor \sqrt{101001} \rfloor = 101$.
The inputs and outputs are given in little-endian representation. This task is \gls*{cs} since the digit-by-digit algorithms to compute the integer square root require many sequence manipulations, including binary multiplications, which is a CS task.

\paragraph{\duplicatestring{} (\gls*{cs})}

Given a binary string, \eg $\bm{x} = abaab$, output the string twice, \ie $\bm{y} = abaababaab$.
This task is \gls*{cs} since it corresponds to the well-known language $\{ww~|~w~\mathrm{is~a~word}\}$ (\cf \cref{sec:background}), which is \gls*{cs}.

\paragraph{\missingduplicate{} (\gls*{cs})}

The input is a binary string of the form $ww$ where $w$ is itself a binary string. One token in this string has been hidden, and the network must find out which one it is, by looking at the value at the same position but on the other side of the string. For instance, if $\bm{x} = ab\_aba$ (\ie $w = aba$), then $\bm{y} = a$.
This task is \gls*{cs} since it requires on the recognition of the well-known language $\{ww~|~w~\mathrm{is~a~word}\}$ (\cf \cref{sec:background}), which is \gls*{cs}.

\paragraph{\oddsfirst{} (\gls*{cs})}

Given a binary string $t_1 ... t_n$, output $t_1t_3t_5...t_2t_4t_6...$.
For example, the output corresponding to $\bm{x} = aaabaa$ is $\bm{y} = aaaaba$.
The task is \gls*{cs} because it exhibits cross-serial dependencies (the output is the interleaved input).

\paragraph{\bucketsort{} (\gls*{cs})}

Given a string over an alphabet of fixed size (5 in our case), return the sorted string.
Since the alphabet has a fixed size, the task can be solved via bucket sort, which only requires a finite amount of counters (\ie 5 counters in our case).
For example, the output corresponding to $\bm{x} = 421302214$ is $\bm{y} = 011222344$.
The task is \gls*{cs} because it requires multiple counters that keep track of the number of occurrences for every token.

\begin{table}
  \caption{
    Our tasks with their level in the Chomsky hierarchy and example input/output pairs.
    The $\dagger$ denotes permutation-invariant tasks; the $\star$ denotes counting tasks; the $\circ$ denotes tasks that require a nondeterministic controller; and the $\times$ denotes tasks that require superlinear running time in terms of the input length.
  }
  \label{tab:tasks}
  \begin{center}
    \resizebox{\columnwidth}{!}{\begin{tabular}{@{}llll@{}}
    \toprule
    \textbf{Level} & \textbf{Name} & \textbf{Example Input} & \textbf{Example Output} \\
    \midrule
    \multirow{4}{*}{\Gls*{regular}} 
    & \evenpairs & $aabba$ & True \\
    & \modulararithmeticsimple & $1+2-4$ & $4$ \\
    & \paritycheck$^\dagger$ & $aaabba$ & True \\
    & \cyclenavigation$^\dagger$ & $011210$ & $2$ \\
    \\
    \multirow{4}{*}{\Gls*{dcf}}
    & \stackmanipulation & $abbaa$ \pop{} \push{} $a$ \pop{} & $abba$ \\
    & \reversestring & $aabba$ & $abbaa$ \\
    & \modulararithmeticbrackets & $-(1-2)\cdot(4-3\cdot(-2))$ & $0$ \\
    & \solveequation$^\circ$ & $-(x-2)\cdot(4-3\cdot(-2))$ & $1$ \\
    \\
    \multirow{7}{*}{\Gls*{cs}}
    & \duplicatestring & $abaab$ & $abaababaab$ \\
    & \missingduplicate & $10011021$ & $0$ \\
    & \oddsfirst & $aaabaa$ & $aaaaba$ \\
    & \binaryaddition & $10010+101$ & $10111$ \\
    & \binarymultiplication$^\times$ & $10010*101$ & $1001000$ \\
    & \computesqrt & $100010$ & $110$ \\
    & \bucketsort$^{\dagger\bigstar}$ & $421302214$ & $011222344$ \\
    \bottomrule
\end{tabular}
 
}
  \end{center}
\end{table}

\subsection{Problem setup}
\label{sec:experimental-details:problem-setup}

We now provide the full formalization of our problem setup (described in \cref{sec:methods}).
\cref{alg:pipeline} illustrates our training procedure.
We train on sequences of length $\ell \sim \mathcal{U}(1, N)$, with $N = 40$ (using batches of size 128).
For testing, we consider sequences of length sampled from $\mathcal{U}(N + 1, M)$, with $M = 500$.
In general, we report the score, which we define as the per-sequence accuracy $A$ (see \cref{sec:methods}) averaged over all sequences of unseen length, \ie $\mathrm{score} := \frac{1}{M - N} \sum_{\ell = N + 1}^M A(\bm{x}, \bm{y})$.
We use the Adam optimizer~\citep{kingma2015adam} with default hyperparameters for \num{1000000} steps, which we have found to be sufficient to achieve a near-optimal training performance for all tasks and architectures.
As described in \cref{alg:pipeline}, we augment our input sequences $\bm{x}$ with $m$ empty tokens $\varnothing$, where $m$ is the length of the corresponding target sequence $\bm{y}$, so that the networks only need to start outputting after they have consumed the entire input sequence.
Moreover, we add computational tokens (which are different from the empty tokens $\varnothing$) to the input sequence $\bm{x}$ for the Tape-RNN to enable it to perform memory operations without having to output the result.
In general, we run all experiments with 10 different random seeds (used for network parameter initialization) and three learning rates (\num{1e-4}, \num{3e-4} and \num{5e-4}), and we report the result obtained by the hyperparameters with the maximum score (means and standard deviations in \cref{sec:additional-experiments}).
However, for Transformers, NDStack-RNNs, and Tape-RNNs we sweep over additional hyperparameters:
For Transformers we consider five different positional encodings, for the NDStack-RNN we optionally read from internal states of the nondeterministic stack (one of the improvements proposed by \citet{dusell2022learning}) and sweep over 2 and 4 symbols, and for the Tape-RNN we consider various numbers of computation tokens (0, $\ell$, or $2\ell$), and various numbers of tapes (1 and 4).

\begin{algorithm}
    \DontPrintSemicolon
    \KwInput{model $p_\theta(\cdot|\bm{x})$ with parameters $\parameters$, learning rate $\alpha$, number of training steps $S$}
    Initialize parameters $\parameters$ \\
    \For{$i\gets1$ \KwTo $S$}{
        Sample length $\ell$ from $\mathcal{U}(1, 40)$ \tcc*{$\ell = 3$}
        Sample sequence $\bm{x}$ of length $\ell$ from the task's input grammar \tcc*{$\bm{x} = 011$ for $\ell = 3$}
        Compute the corresponding output sequence $\bm{y}$ of length $m$ \tcc*{$\bm{y} = 110$ with $m = 3$}
        Pad $\bm{x}$ with $m$ empty tokens $\varnothing$ such that the model only needs to start outputting \emph{after} having consumed the entire input sequence, \ie $\bm{x} \gets \bm{x} \cdot \varnothing \ldots \varnothing$ \tcc*{$\bm{x} = 011\varnothing\varnothing\varnothing$}
        Set the output to empty sequence of length $m$, \ie $\bm{o} \gets \varnothing \ldots \varnothing$ \\
        \For{$t\gets 1$ \KwTo $m$}{
            Compute the output probability distribution $o_t \gets \model(\cdot|\bm{x}_{1:\ell + t})$ \tcc*{$o_t \in \mathbb{R}^2$}
        }
        Compute the cross-entropy loss (averaged over output tokens) $\mathcal{C} \gets -\frac{1}{m}\sum_{t=1}^m y_t^\top \log o_t$ \\
        Update the parameters with gradient descent $\theta \gets \theta - \alpha \nabla \mathcal{C}$ \\
        Compute the per-sequence accuracy $A \gets \frac{1}{m} \sum_{t=1}^m \mathds{1}\left[\argmax_j y_{tj} = \argmax_j o_{tj} \right]$
    }
    \caption{
        Training pipeline for our sequence prediction tasks.
        The comments (in blue) show an example output for the \reversestring{} (\gls*{dcf}) task.
    }
    \label{alg:pipeline}
\end{algorithm}

\subsection{Computational resources}
\label{sec:experimental-details:computational-resources}

We implemented our evaluation suite in JAX \citep{bradbury2018jax} using the DeepMind JAX ecosystem \citep{babuschkin2020deepmind, hessel2020optax, hennigan2020haiku}.
We make all of our code publicly available at \url{https://github.com/deepmind/neural_networks_chomsky_hierarchy}.
We ran each task-architecture-hyperparameter triplet on a single TPU on our internal cluster. 
For RNNs, Stack-RNNs, LSTMs and Stack-LSTMs, the hyperparameters we have used are learning rates and seeds, which gives $15~(\mathrm{tasks}) \cdot 4~(\mathrm{models}) \cdot 3~(\mathrm{learning~rates}) \cdot 10~(\mathrm{seeds}) = 1800$ TPU-units.
For Tape-RNNs, the hyperparameters we have used are the learning rate, the number of computation steps tokens, the number of tapes, and the random seeds, which yields $15~(\mathrm{tasks}) \cdot 3~(\mathrm{learning~rates}) \cdot 3~(\mathrm{computation~steps}) \cdot 2~(\mathrm{tapes}) \cdot 10~(\mathrm{seeds}) = 2700$ TPU-units.
For Transformers (encoder and autoregressive), the hyperparameters we have used are the learning rate, the positional encodings, and the random seeds, which gives $15~(\mathrm{tasks}) \cdot 2~(\mathrm{models}) \cdot 5~(\mathrm{positional~encodings}) \cdot 3~(\mathrm{learning~rates}) \cdot 10~(\mathrm{seeds}) = 4500$ TPU-units.
For the NDStack-RNN, the hyperparameters we have used are the learning rate, the number of stack symbols, and the random seeds, which gives $15~(\mathrm{tasks}) \cdot 2~(\mathrm{number~of~symbols}) \cdot 3~(\mathrm{learning~rates}) \cdot 10~(\mathrm{seeds}) = 900$ TPU-units.
Moreover, we used an additional $300$ TPU-units for the phase transition experiment (\cref{sec:additional-experiments:phase-transition}).
For the CNN experiment, we used $15~(\mathrm{tasks}) \cdot 4~(\mathrm{kernel~sizes}) \cdot 3~(\mathrm{learning~rates}) \cdot 10~(\mathrm{seeds}) = 1800$ TPU-units.
For the scaling laws experiment, we used $15~(\mathrm{tasks}) \cdot 3~(\mathrm{training~steps}) \cdot 5~(\mathrm{num~layers}) \cdot 3~(\mathrm{learning~rates}) \cdot 10~(\mathrm{seeds}) = 6750$ TPU-units.
Finally, for the autoregresive experiments (excluding the autoregressive Transformer), we used $8~(\mathrm{tasks}) \cdot 3~(\mathrm{learning~rates}) \cdot 10~(\mathrm{seeds}) = 240$ TPU-units for the RNNs, Stack-RNNs, and LSTMs, \ie $720$ TPU-units, and $8~(\mathrm{tasks}) \cdot 3~(\mathrm{computation~steps}) \cdot 2~(\mathrm{tapes}) \cdot 3~(\mathrm{learning~rates}) \cdot 10~(\mathrm{seeds}) = 1440$ TPU-units for the Tape-RNNs.
Thus, in total we used $\num{20910}$ TPU-units.
We report all the running times, for each task and architecture, in \cref{tab:running_times}.
Note that we aggregated the results over the hyperparameters, which explains the high standard deviation for Transformers (five different positional encodings) and Tape-RNN (three different numbers of computation steps tokens, two different numbers of tapes).

The NDStack-RNN is comparably slow to train. It is roughly four times slower than a classical
Stack-RNN, due to the computation of the big tensor $\gamma$, denoting the transition weights between
nodes. Furthermore, this tensor takes a lot of space in memory, \ie the number of values it
contains is, once unrolled on the whole trajectory, proportional to $(l+m)^3$, where l and m are the
length of the input and output sequences, respectively. Thus, for our training range of 40 and a task
where the output length is equal to the input length (\eg Reverse String (DCF)), the required
space is proportional to $(40 + 40)^3 = 512 000$. Therefore, with a batch size of 128 and 16 possible
actions (\ie state-symbol pairs), the tensor is roughly of size 33.5Gb using float32 representation,
which exceeds the memory size of our GPUs. Consequently, we reduced the batch size to 16 (\ie 8
times smaller than normal) for this architecture.

\begin{table}
  \caption{
    Mean and standard deviation of the running times (in hours) for all the architectures and tasks.
    The $\dagger$ denotes permutation-invariant tasks; the $\star$ denotes counting tasks; the $\circ$ denotes tasks that require a nondeterministic controller; and the $\times$ denotes tasks that require superlinear running time in terms of the input length.
  }
  \label{tab:running_times}
  \begin{center}
    \resizebox{\columnwidth}{!}{\begin{tabular}{@{}llrrrrrrrr@{}}
    \toprule
    \textbf{Level} & \textbf{Task}  & \textbf{RNN} & \textbf{Stack-RNN} & \textbf{NDStack-RNN} & \textbf{Tape-RNN} & \textbf{Transformer} & \textbf{LSTM} & \textbf{Stack-LSTM} & \textbf{Autoreg.\ Transformer} \\
    \midrule
    \multirow{4}{*}{\Gls*{regular}}
    & \evenpairs  & $0.37 \pm 0.01$ & $0.56 \pm 0.01$ & $8.50 \pm 0.16$ & $3.31 \pm 2.41$  & $1.19 \pm 0.73$ & $0.75 \pm 0.03$ & $2.97 \pm 0.01$ & $1.61 \pm 0.27$ \\
    & \modulararithmeticsimple  & $0.57 \pm 0.04$ & $0.92 \pm 0.06$ & $8.34 \pm 0.19$ & $3.44 \pm 2.48$  & $1.31 \pm 0.55$ & $0.62 \pm 0.00$ & $3.05 \pm 0.01$ & $1.51 \pm 0.24$ \\
    & \paritycheck$^\dagger$ & $0.59 \pm 0.02$  & $0.78 \pm 0.05$  & $8.54 \pm 0.12$ & $3.37 \pm 2.32$  & $1.23 \pm 0.58$ & $0.68 \pm 0.06$ & $2.96 \pm 0.01$ & $1.61 \pm 0.27$ \\
    & \cyclenavigation$^\dagger$  & $0.60 \pm 0.03$ & $0.57 \pm 0.07$  & $8.42 \pm 0.23$ & $3.26 \pm 2.25$  & $1.23 \pm 0.79$ & $0.76 \pm 0.03$ & $2.97 \pm 0.01$ & $1.66 \pm 0.45$ \\
    \\
    \multirow{4}{*}{\Gls*{dcf}}
    & \stackmanipulation  & $4.07 \pm 0.12$ & $4.69 \pm 0.85$ & $17.80 \pm 1.43$ & $8.27 \pm 2.93$  & $5.06 \pm 0.65$ & $4.27 \pm 0.10$ & $8.89 \pm 0.04$ & $7.25 \pm 2.24$ \\
    & \reversestring  & $0.50 \pm 0.02$ & $0.88 \pm 0.11$ & $16.66 \pm 0.99$ & $4.58 \pm 2.88$  & $1.62 \pm 0.78$ & $0.70 \pm 0.03$ & $5.48 \pm 0.01$ & $4.32 \pm 2.03$ \\
    & \modulararithmeticbrackets  & $2.70 \pm 0.08$ & $3.11 \pm 0.95$ & $9.04 \pm 0.78$ & $5.44 \pm 2.16$  & $3.59 \pm 0.65$ & $3.23 \pm 0.72$ & $5.21 \pm 0.04$ & $3.99 \pm 0.47$ \\
    & \solveequation$^\circ$  & $2.82 \pm 0.06$ & $3.24 \pm 0.79$ & $9.16 \pm 0.64$ & $5.69 \pm 2.30$  & $3.59 \pm 0.49$ & $2.96 \pm 0.10$ & $5.25 \pm 0.05$ & $3.91 \pm 0.28$ \\
    \\
    \multirow{7}{*}{\Gls*{cs}}
    & \duplicatestring  & $0.57 \pm 0.05$ & $1.33 \pm 0.13$ & $25.25 \pm 2.31$ & $5.90 \pm 3.66$  & $2.66 \pm 0.76$ & $0.92 \pm 0.05$ & $7.92 \pm 0.02$ & $5.02 \pm 0.93$ \\
    & \missingduplicate  & $0.49 \pm 0.04$ & $0.71 \pm 0.04$ & $8.49 \pm 0.27$ & $3.31 \pm 2.16$  & $1.24 \pm 0.35$ & $0.60 \pm 0.02$ & $3.05 \pm 0.01$ & $1.72 \pm 0.30$ \\
    & \oddsfirst  & $0.55 \pm 0.04$ & $0.99 \pm 0.05$ & $16.63 \pm 1.34$ & $4.49 \pm 2.81$  & $1.67 \pm 0.38$ & $0.80 \pm 0.04$ & $5.48 \pm 0.01$ & $4.57 \pm 1.72$ \\
    & \binaryaddition  & $1.33 \pm 0.05$ & $1.35 \pm 0.04$ & $17.36 \pm 0.45$ & $4.69 \pm 2.94$  & $2.29 \pm 0.75$ & $1.35 \pm 0.05$ & $5.62 \pm 0.01$ & $5.14 \pm 1.48$ \\
    & \binarymultiplication$^\times$  & $1.35 \pm 0.06$ & $1.33 \pm 0.04$ & $17.04 \pm 0.76$ & $4.53 \pm 2.76$  & $2.06 \pm 0.37$ & $1.34 \pm 0.04$ & $5.49 \pm 0.01$ & $4.64 \pm 1.95$ \\
    & \computesqrt  & $0.95 \pm 0.05$ & $1.05 \pm 0.05$ & $12.62 \pm 1.21$ & $3.82 \pm 2.45$  & $1.57 \pm 0.34$ & $0.97 \pm 0.04$ & $4.19 \pm 0.01$ & $3.30 \pm 1.27$ \\
    & \bucketsort$^{\dagger\bigstar}$ & $0.58 \pm 0.09$ & $0.98 \pm 0.08$ & $16.75 \pm 2.01$ & $3.54 \pm 2.24$ & $1.34 \pm 0.76$ & $0.79 \pm 0.09$ & $5.64 \pm 0.05$ & $4.52 \pm 1.77$ \\
    \bottomrule
\end{tabular}
 }
  \end{center}
\end{table}

\section{Additional Experiments}
\label{sec:additional-experiments}

\subsection{Main result}
\label{sec:additional-experiments:main-result}

\begin{figure}
    \begin{center}
        \begin{subfigure}[b]{0.31\textwidth}
            \begin{center}
                \includegraphics[width=\textwidth]{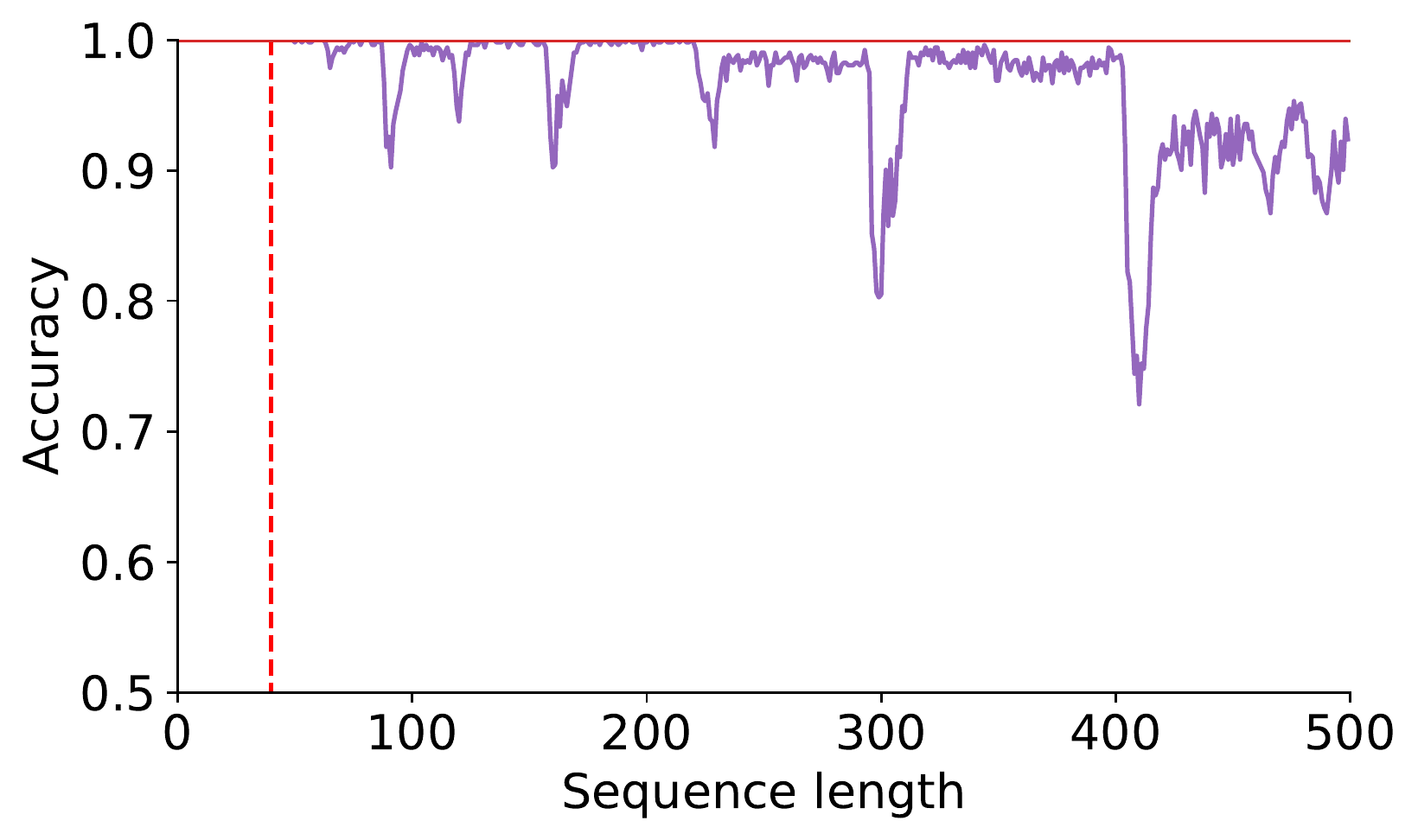}
            \end{center}
            \vspace{-0.25cm}
            \caption{\tiny{\evenpairs{} (\gls*{regular})}}
        \end{subfigure}
        \hspace{0.01\textwidth}
        \begin{subfigure}[b]{0.31\textwidth}
            \begin{center}
                \includegraphics[width=\textwidth]{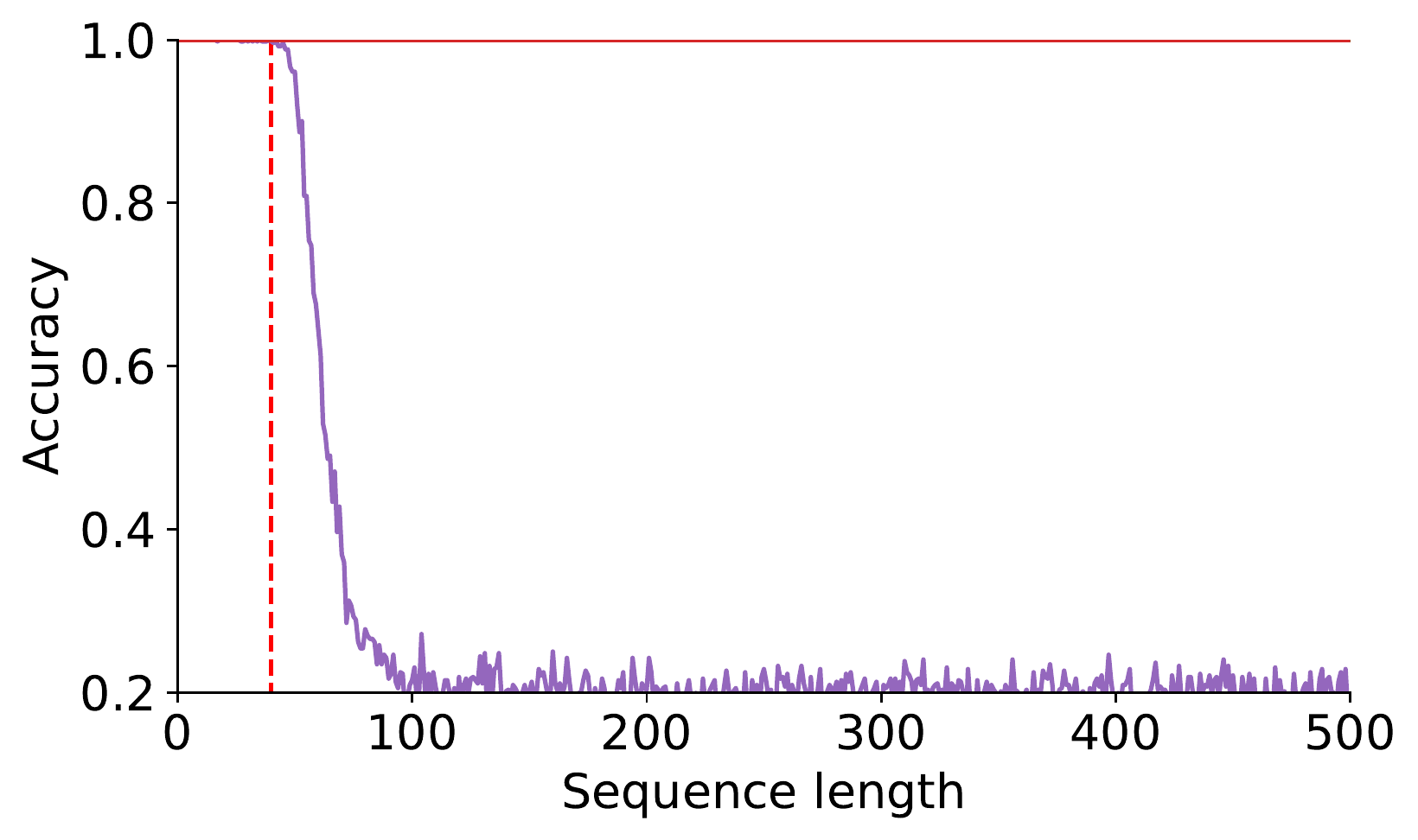}
            \end{center}
            \vspace{-0.25cm}
            \caption{\tiny{\modulararithmeticsimple{} (\gls*{regular})}}
        \end{subfigure}
        \hspace{0.01\textwidth}
        \begin{subfigure}[b]{0.31\textwidth}
            \begin{center}
                \includegraphics[width=\textwidth]{figures/parity_check_all_archis-eps-converted-to.pdf}
            \end{center}
            \vspace{-0.25cm}
            \caption{\tiny{\paritycheck{} (\gls*{regular})}}
        \end{subfigure}
        \vspace{0.5cm}
        
        \begin{subfigure}[b]{0.31\textwidth}
            \begin{center}
                \includegraphics[width=\textwidth]{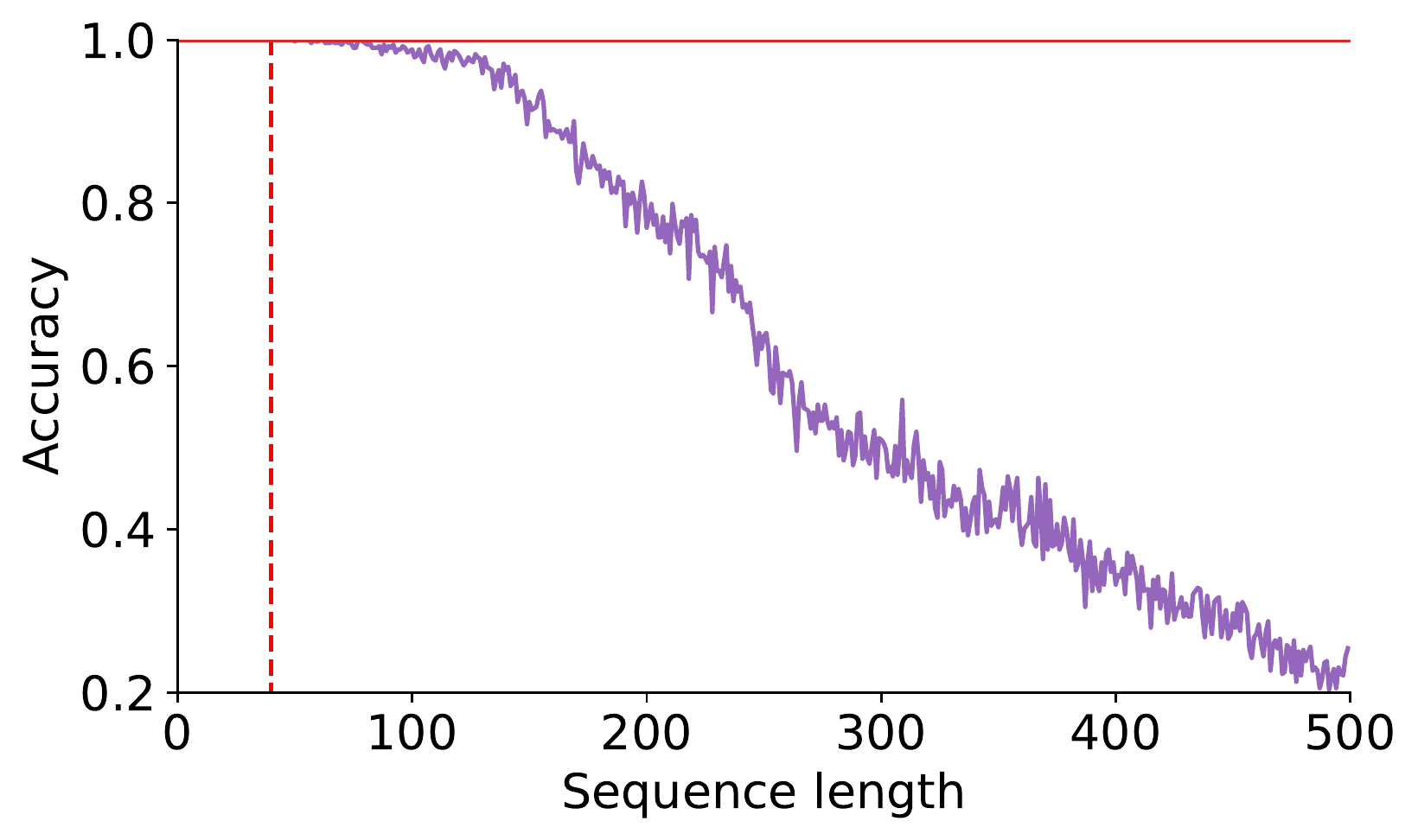}
            \end{center}
            \vspace{-0.25cm}
            \caption{\tiny{\cyclenavigation{} (\gls*{regular})}}
        \end{subfigure}
        \hspace{0.01\textwidth}
        \begin{subfigure}[b]{0.31\textwidth}
            \begin{center}
                \includegraphics[width=\textwidth]{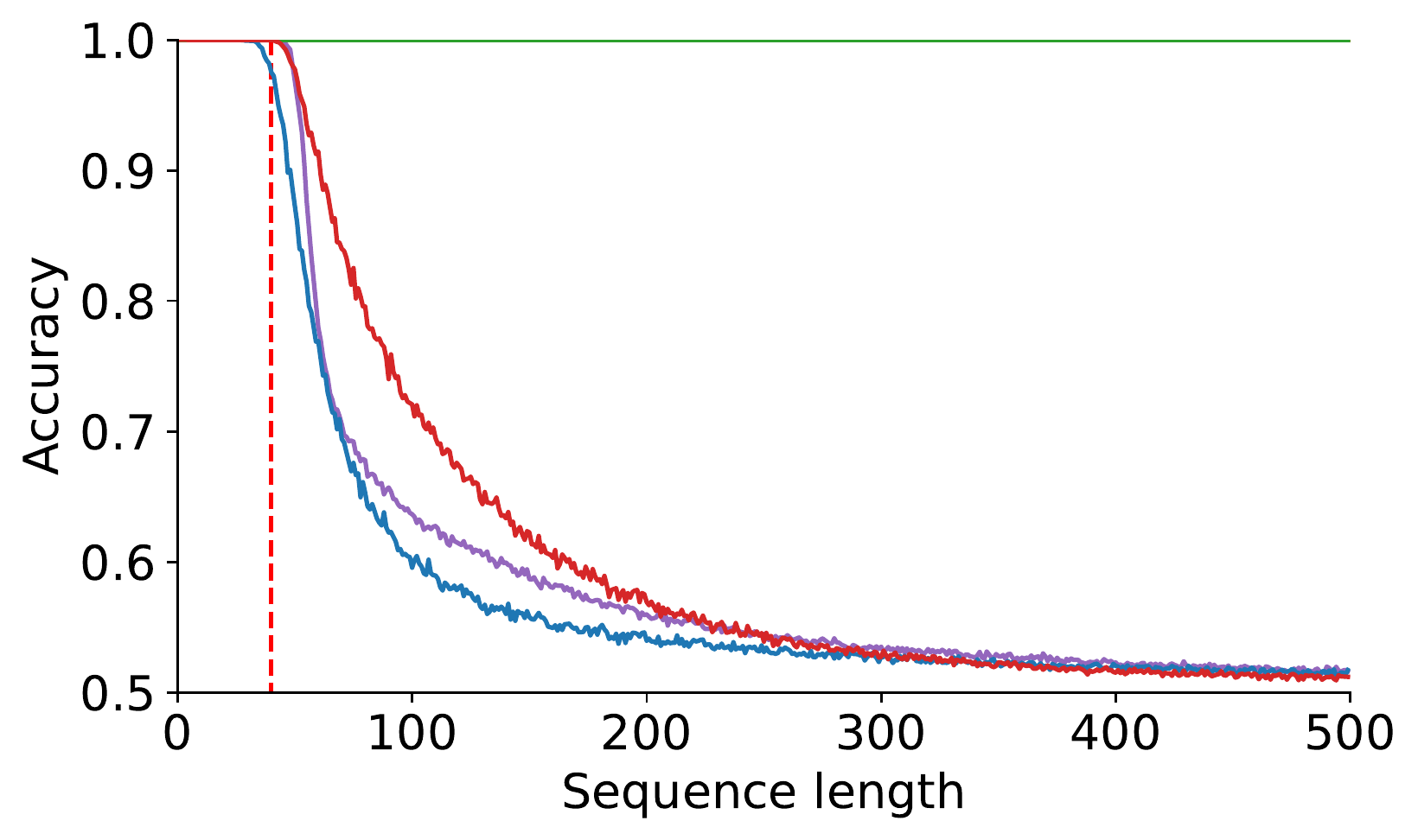}
            \end{center}
            \vspace{-0.25cm}
            \caption{\tiny{\stackmanipulation{} (\gls*{dcf})}}
        \end{subfigure}
        \hspace{0.01\textwidth}
        \begin{subfigure}[b]{0.31\textwidth}
            \begin{center}
                \includegraphics[width=\textwidth]{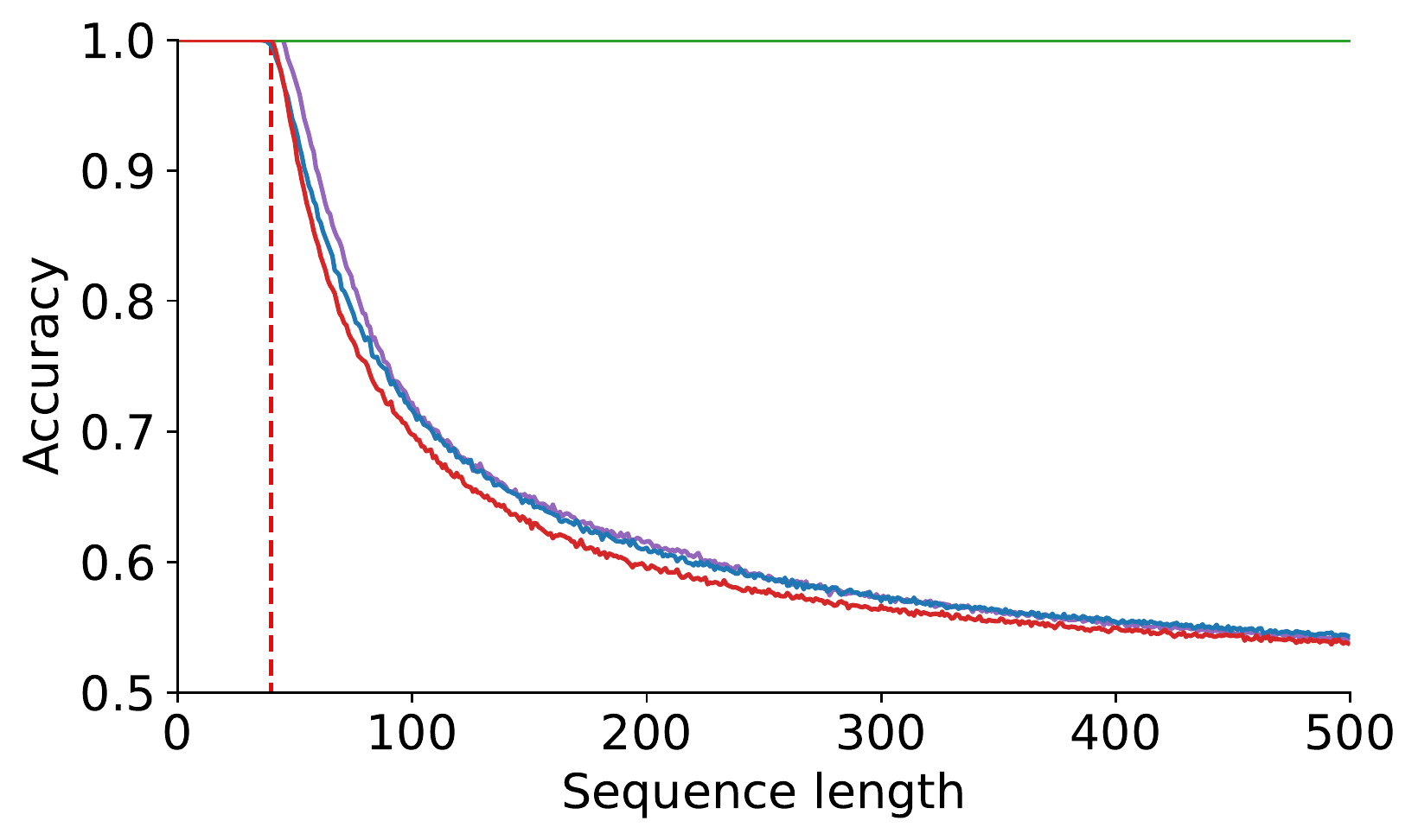}
            \end{center}
            \vspace{-0.25cm}
            \caption{\tiny{\reversestring{} (\gls*{dcf})}}
        \end{subfigure}
        \vspace{0.5cm}
        
        \begin{subfigure}[b]{0.31\textwidth}
            \begin{center}
                \includegraphics[width=\textwidth]{figures/modular_arithmetic_brackets_all_archis-eps-converted-to.pdf}
            \end{center}
            \vspace{-0.25cm}
            \caption{\tiny{\modulararithmeticbrackets{} (\gls*{dcf})}}
        \end{subfigure}
        \hspace{0.01\textwidth}
        \begin{subfigure}[b]{0.31\textwidth}
            \begin{center}
                \includegraphics[width=\textwidth]{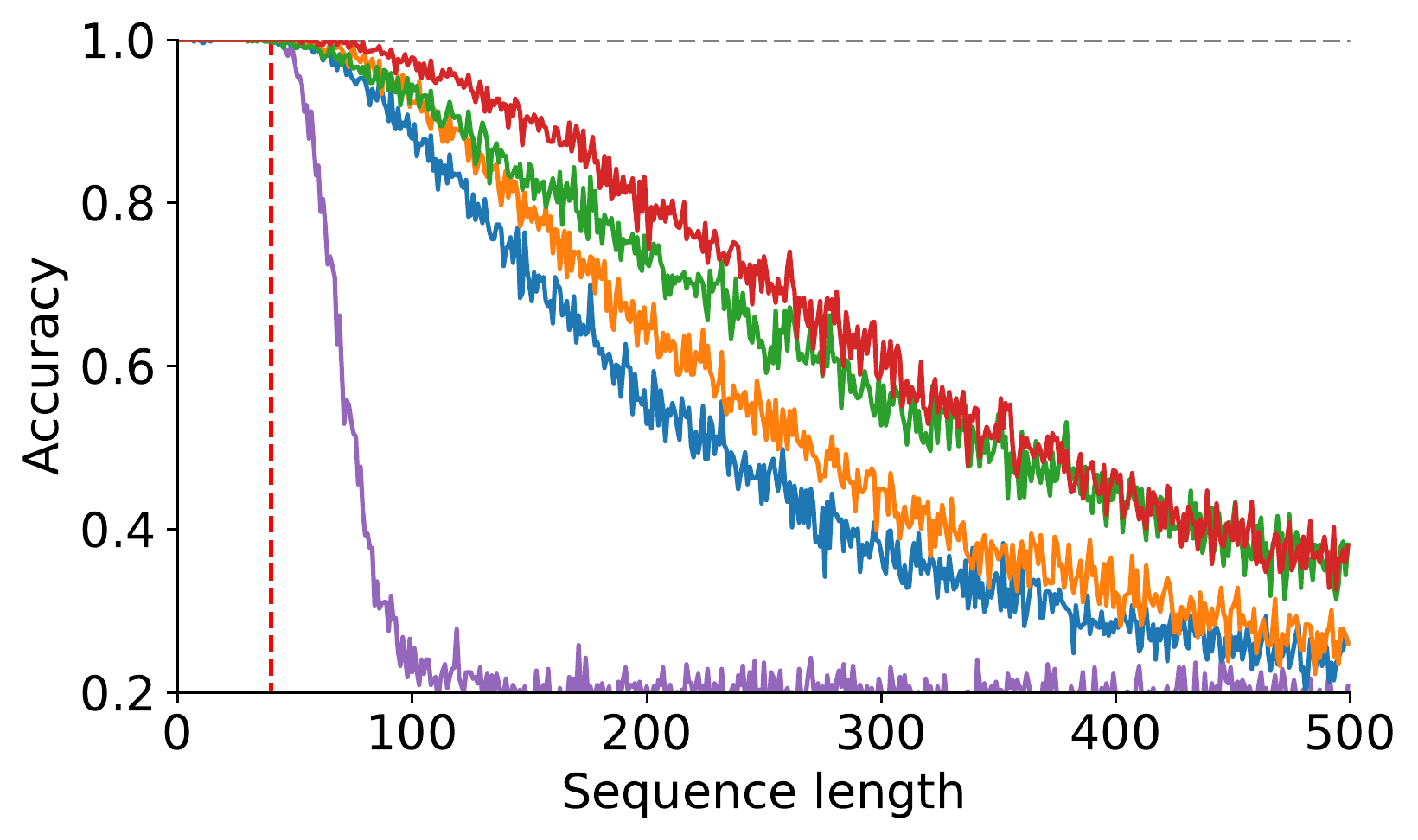}
            \end{center}
            \vspace{-0.25cm}
            \caption{\tiny{\solveequation{} (\gls*{dcf})}}
        \end{subfigure}
        \hspace{0.01\textwidth}
        \begin{subfigure}[b]{0.31\textwidth}
            \begin{center}
                \includegraphics[width=\textwidth]{figures/duplicate_string_all_archis-eps-converted-to.pdf}
            \end{center}
            \vspace{-0.25cm}
            \caption{\tiny{\duplicatestring{} (\gls*{cs})}}
        \end{subfigure}
        \hspace{0.01\textwidth}
        \vspace{0.5cm}
        
        \begin{subfigure}[b]{0.31\textwidth}
            \begin{center}
                \includegraphics[width=\textwidth]{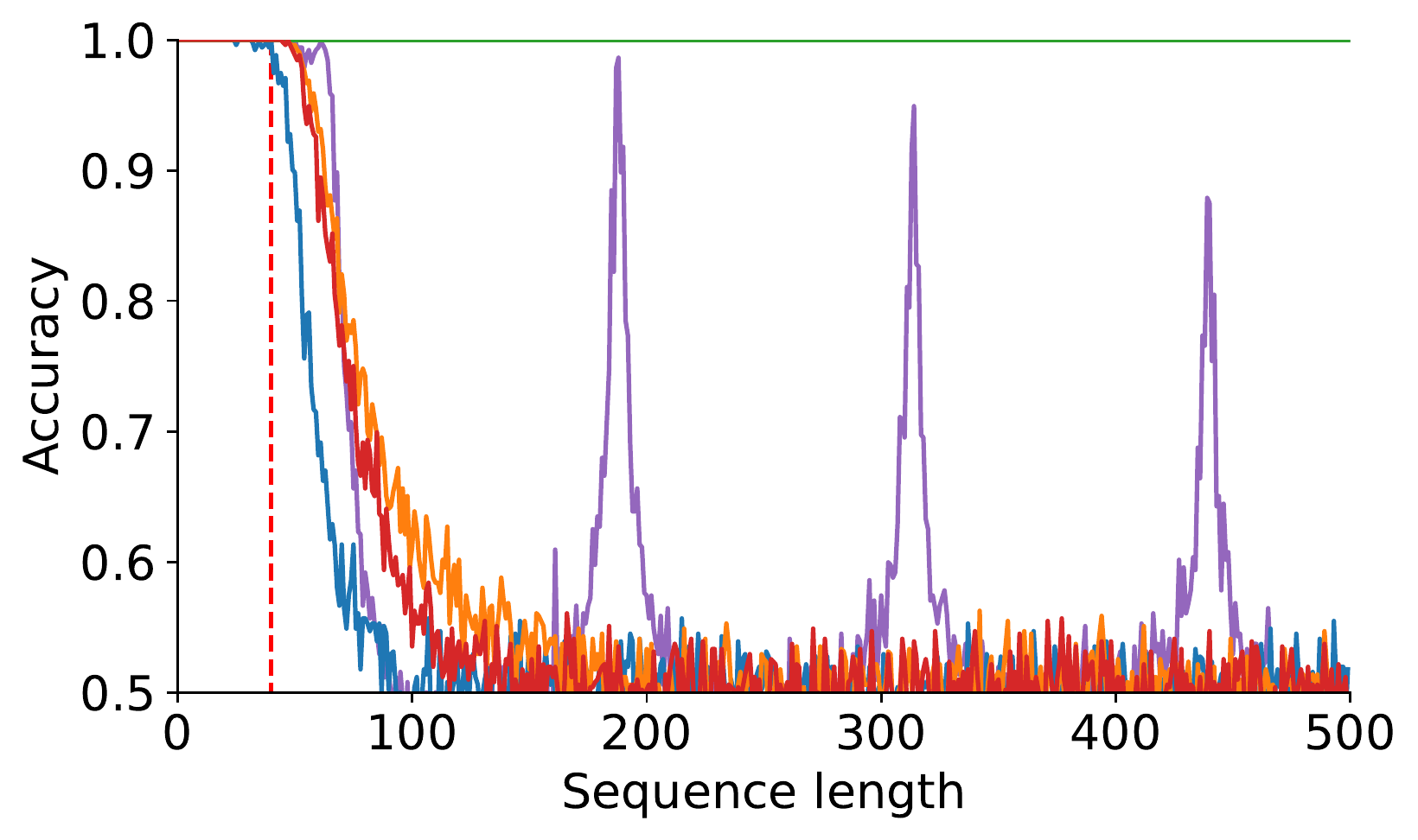}
            \end{center}
            \vspace{-0.25cm}
            \caption{\tiny{\missingduplicate{} (\gls*{cs})}}
        \end{subfigure}
        \hspace{0.01\textwidth}
        \begin{subfigure}[b]{0.31\textwidth}
            \begin{center}
                \includegraphics[width=\textwidth]{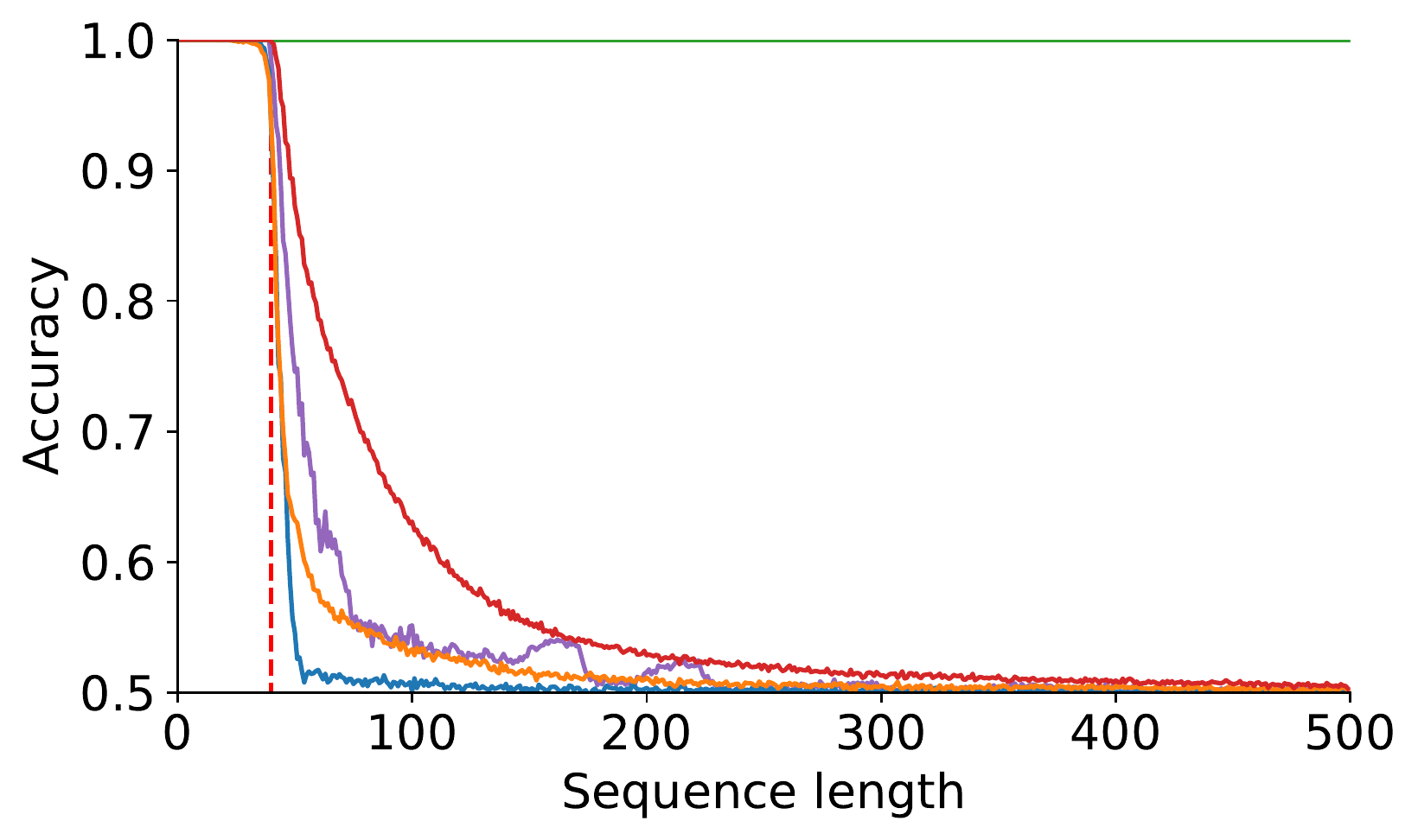}
            \end{center}
            \vspace{-0.25cm}
            \caption{\tiny{\oddsfirst{} (\gls*{cs})}}
        \end{subfigure}
        \hspace{0.01\textwidth}
        \begin{subfigure}[b]{0.31\textwidth}
            \begin{center}
                \includegraphics[width=\textwidth]{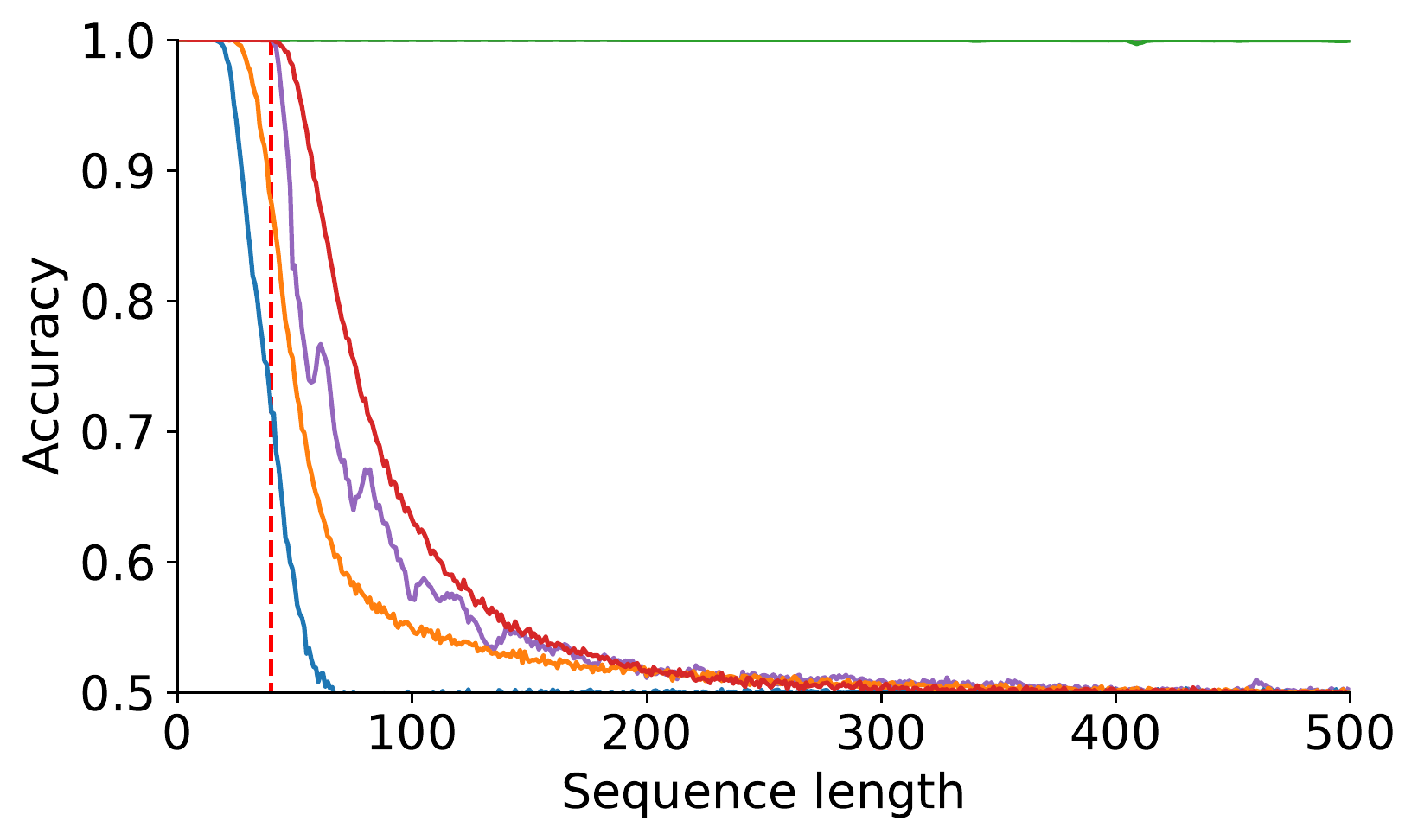}
            \end{center}
            \vspace{-0.25cm}
            \caption{\tiny{\binaryaddition{} (\gls*{cs})}}
        \end{subfigure}
        \vspace{0.5cm}
        
        \begin{subfigure}[b]{0.31\textwidth}
            \begin{center}
                \includegraphics[width=\textwidth]{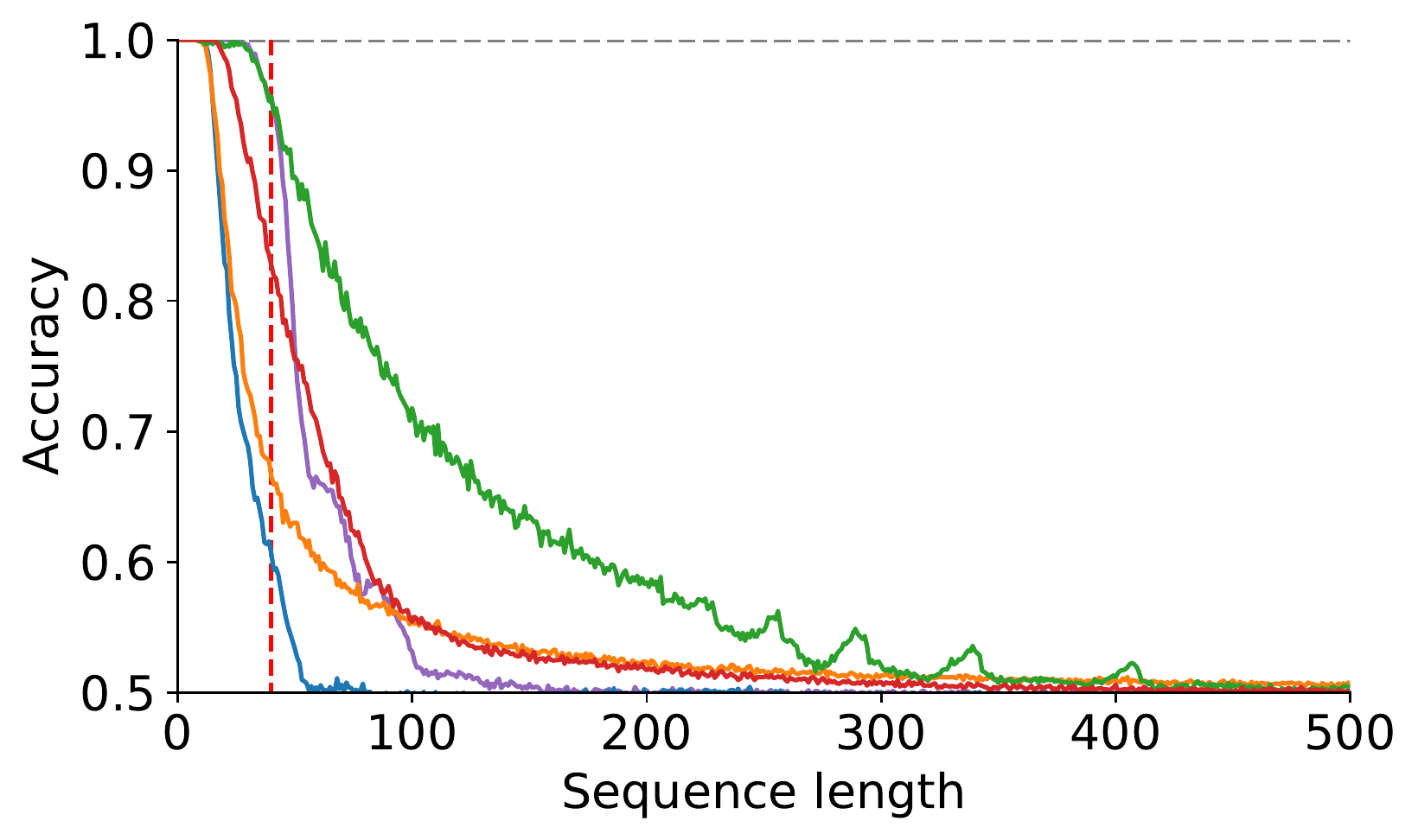}
            \end{center}
            \vspace{-0.25cm}
            \caption{\tiny{\binarymultiplication{} (\gls*{cs})}}
        \end{subfigure}
        \hspace{0.01\textwidth}
        \begin{subfigure}[b]{0.31\textwidth}
            \begin{center}
                \includegraphics[width=\textwidth]{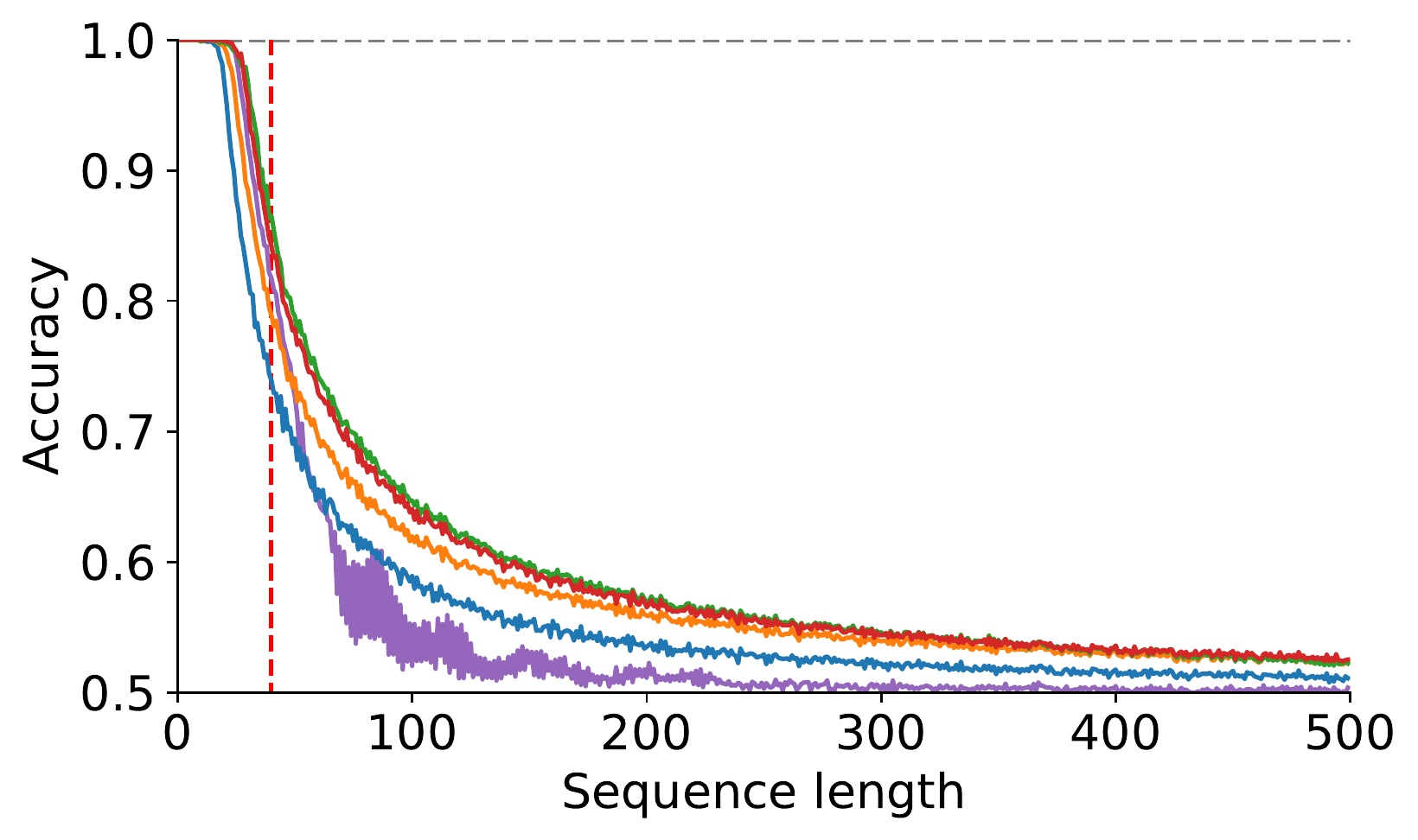}
            \end{center}
            \vspace{-0.25cm}
            \caption{\tiny{\computesqrt{} (\gls*{cs})}}
        \end{subfigure}
        \hspace{0.01\textwidth}
        \begin{subfigure}[b]{0.31\textwidth}
            \begin{center}
                \includegraphics[width=\textwidth]{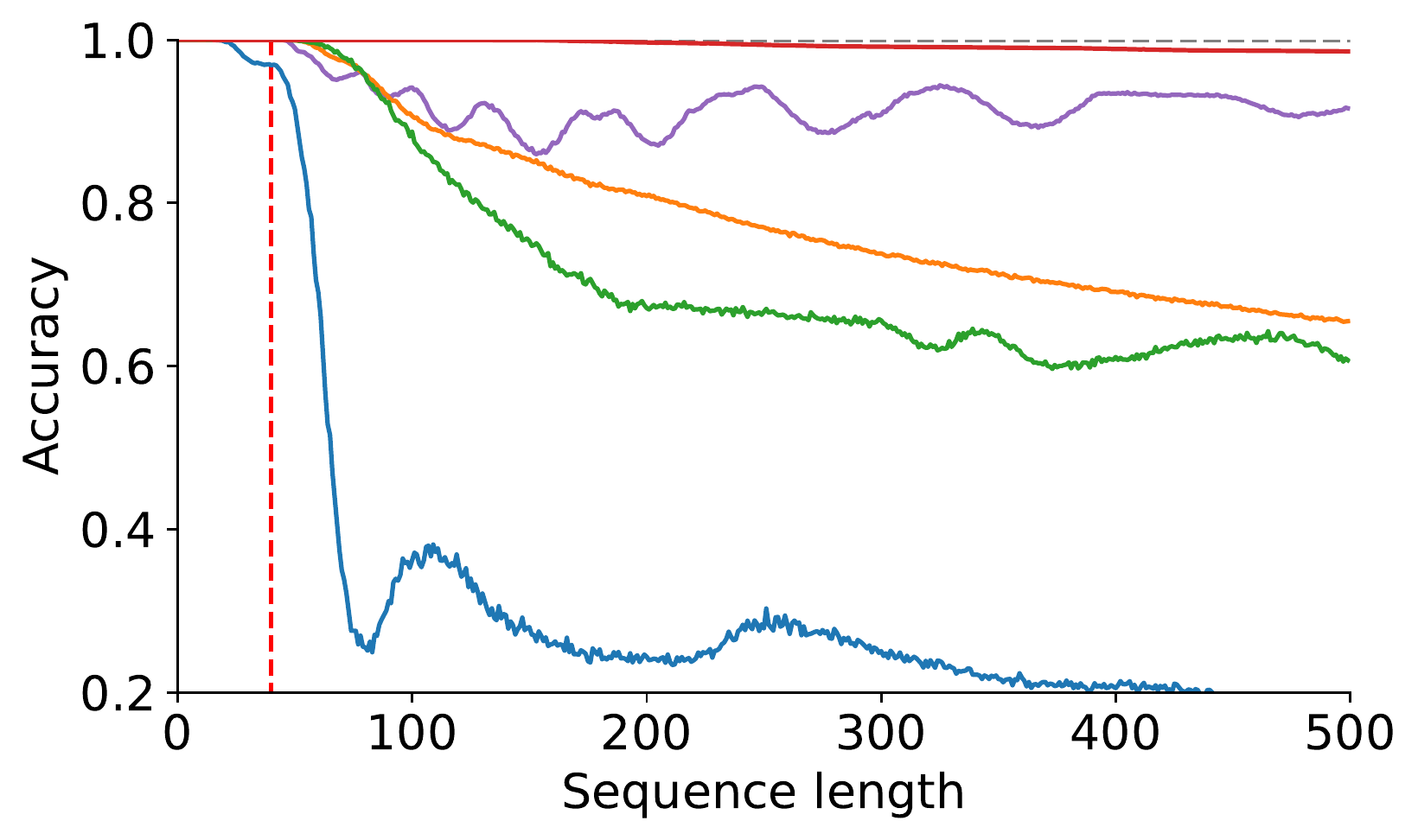}
            \end{center}
            \vspace{-0.25cm}
            \caption{\tiny{\bucketsort{} (\gls*{cs})}}
        \end{subfigure}
        \hspace{0.01\textwidth}
    \end{center}
    \caption{
        Performance curves on all tasks. For the Transformer encoder, we pick the best positional encoding for each task (it is considered as a hyperparameter).
        The dashed vertical red line is the training range, meaning that sequences to the right have not been seen during training and thus measure generalization.
    }
    \label{fig:scores-all}
\end{figure}

We visualize the performance curves for all tasks in \cref{fig:scores-all} and report the means and standard deviations of the test accuracies for all architectures on our tasks in \cref{tab:scores-means-variances}.
We observe that all the networks, except for Transformers, solve all regular tasks perfectly for all the seeds.
However, only a few tasks above beyond \acrlong*{regular} can be solved (defined by a score larger or equal to $90\%$) on average, but the hierarchy still persists.
For example, the Tape-RNN has much better scores overall, even on average, on the \gls*{cs} tasks compared to the other architectures.
However, even though the Tape-RNN can solve more tasks, it still fails on some of them.
We hypothesize that it is because the limited set of actions makes the right action trajectory long and difficult to find.
Nevertheless, the Tape-RNN is the closest we have to an architecture capable of solving all tasks from \acrlong*{regular} to \gls*{cs}.
Note that networks may be evaluated on unseen sequences even within the training range, as we are sampling from a training distribution and not a finite dataset (as discussed in \cref{sec:methods}).
Thus, it is possible that some models do not achieve a perfect accuracy within the training range $\mathcal{U}(1, N)$, such as in the \computesqrt{} (\gls*{cs}) task (\cf \cref{fig:scores-all}).

\begin{table}
  \caption{
  Means and standard deviations (computed over random seeds) of the score (average test accuracy, see \cref{sec:methods}) for the results of the main experiment (see \cref{tab:scores} and \cref{sec:results:main-result}).
  The $\dagger$ denotes permutation-invariant tasks; the $\star$ denotes counting tasks; the $\circ$ denotes tasks that require a nondeterministic controller; and the $\times$ denotes tasks that require superlinear running time in terms of the input length.
  }
  \label{tab:scores-means-variances}
  \begin{center}
    \resizebox{\columnwidth}{!}{\begin{tabular}{@{}llrrrrr@{}}
    \toprule
    \textbf{Level} & \textbf{Tasks} & \textbf{RNN} & \textbf{Stack-RNN} & \textbf{Tape-RNN} & \textbf{Transformer} & \textbf{LSTM} \\
    \midrule
    \multirow{4}{*}{\Gls*{regular}} & \evenpairs & $\textbf{100.0} \pm 0.0$ & $\textbf{100.0} \pm 0.0$ & $\textbf{100.0} \pm 0.0$ & $67.7 \pm 15.2$ & $\textbf{100.0} \pm 0.0$ \\
    & \modulararithmeticsimple & $\textbf{100.0} \pm 0.0$ & $\textbf{100.0} \pm 0.0$ & $\textbf{100.0} \pm 0.0$ & $23.1 \pm 0.8$ & $\textbf{100.0} \pm 0.0$ \\
    & \paritycheck$^\dagger$ & $\textbf{95.5} \pm 14.2$ & $\textbf{100.0} \pm 0.0$ & $\textbf{100.0} \pm 0.0$ & $50.4 \pm 0.7$ & $\textbf{100.0} \pm 0.0$ \\
    & \cyclenavigation$^\dagger$ & $\textbf{100.0} \pm 0.0$ & $\textbf{96.8} \pm 6.6$ & $\textbf{100.0} \pm 0.0$ & $33.9 \pm 10.5$ & $\textbf{90.0} \pm 10.8$ \\
    \\
    \multirow{4}{*}{\Gls*{dcf}} 
    & \stackmanipulation & $54.5 \pm 1.1$ & $\textbf{93.8} \pm 15.1$ & $66.9 \pm 22.6$ & $51.1 \pm 8.1$ & $58.1 \pm 0.7$ \\
    & \reversestring & $60.4 \pm 2.3$ & $\textbf{96.2} \pm 12.1$ & $\textbf{90.7} \pm 15.6$ & $56.1 \pm 2.7$ & $59.5 \pm 1.0$ \\
    & \modulararithmeticbrackets & $34.6 \pm 4.3$ & $75.8 \pm 11.3$ & $60.9 \pm 20.7$ & $28.2 \pm 3.4$ & $53.1 \pm 4.3$ \\
    & \solveequation$^\circ$ & $31.3 \pm 8.7$ & $46.1 \pm 7.3$ & $37.7 \pm 12.7$ & $23.5 \pm 1.6$ & $37.1 \pm 14.0$ \\
    \\
    \multirow{7}{*}{\Gls*{cs}} 
    & \duplicatestring & $50.2 \pm 0.0$ & $51.6 \pm 0.6$ & $\textbf{90.0} \pm 16.3$ & $52.8 \pm 0.0$ & $56.5 \pm 0.9$ \\
    & \missingduplicate & $ 51.7 \pm 0.2 $ & $ 53.6 \pm 0.7$ & $66.1 \pm 23.3$ & $53.8 \pm 1.5$ & $53.1 \pm 0.6$ \\
    & \oddsfirst & $50.4 \pm 0.2$ & $51.5 \pm 0.1$ & $76.2 \pm 16.6$ & $52.7 \pm 0.0$ & $54.6 \pm 0.8$ \\
    & \binaryaddition & $49.5 \pm 0.3$ & $50.9 \pm 0.9$ & $74.3 \pm 10.6$ & $51.7 \pm 1.3$ & $54.8 \pm 0.4$ \\
    & \binarymultiplication$^\times$ & $49.6 \pm 0.2$ & $51.4 \pm 0.7$ & $54.5 \pm 2.2$ & $50.4 \pm 3.6$ & $52.9 \pm 0.2$ \\
    & \computesqrt & $53.9 \pm 0.3$ & $56.1 \pm 0.3$ & $55.6 \pm 1.4$ & $51.5 \pm 0.4$ & $57.3 \pm 0.2$ \\
    & \bucketsort$^{\dagger\bigstar}$ & $22.5 \pm 3.6$ & $56.6 \pm 16.2$ & $33.6 \pm 12.8$ & $35.7 \pm 23.1$ & $79.4 \pm 21.9$ \\
    \bottomrule
\end{tabular}
 }
  \end{center}
\end{table}

\subsection{Phase transition under increasing training ranges}
\label{sec:additional-experiments:phase-transition}

\begin{figure}
    \begin{center}
        \begin{subfigure}[b]{0.49\textwidth}
            \begin{center}
                \includegraphics[width=0.75\textwidth]{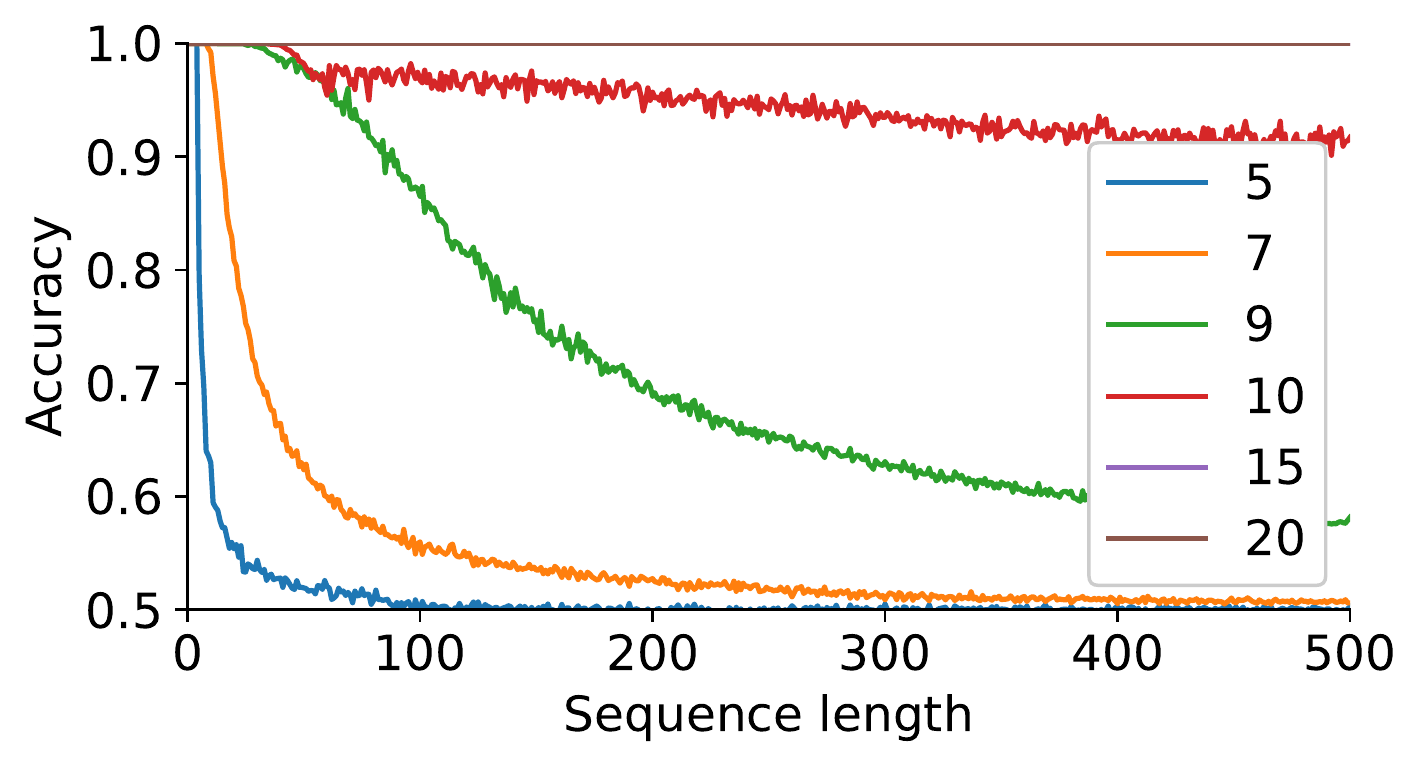}
            \end{center}
            \caption{Maximum accuracy (over 10 different random seeds) per length for different training ranges (in color).} 
            \label{fig:phase-transition:maximum}
        \end{subfigure}
        \hspace{0.005\textwidth}
        \begin{subfigure}[b]{0.49\textwidth}
            \begin{center}
                \includegraphics[width=0.75\textwidth]{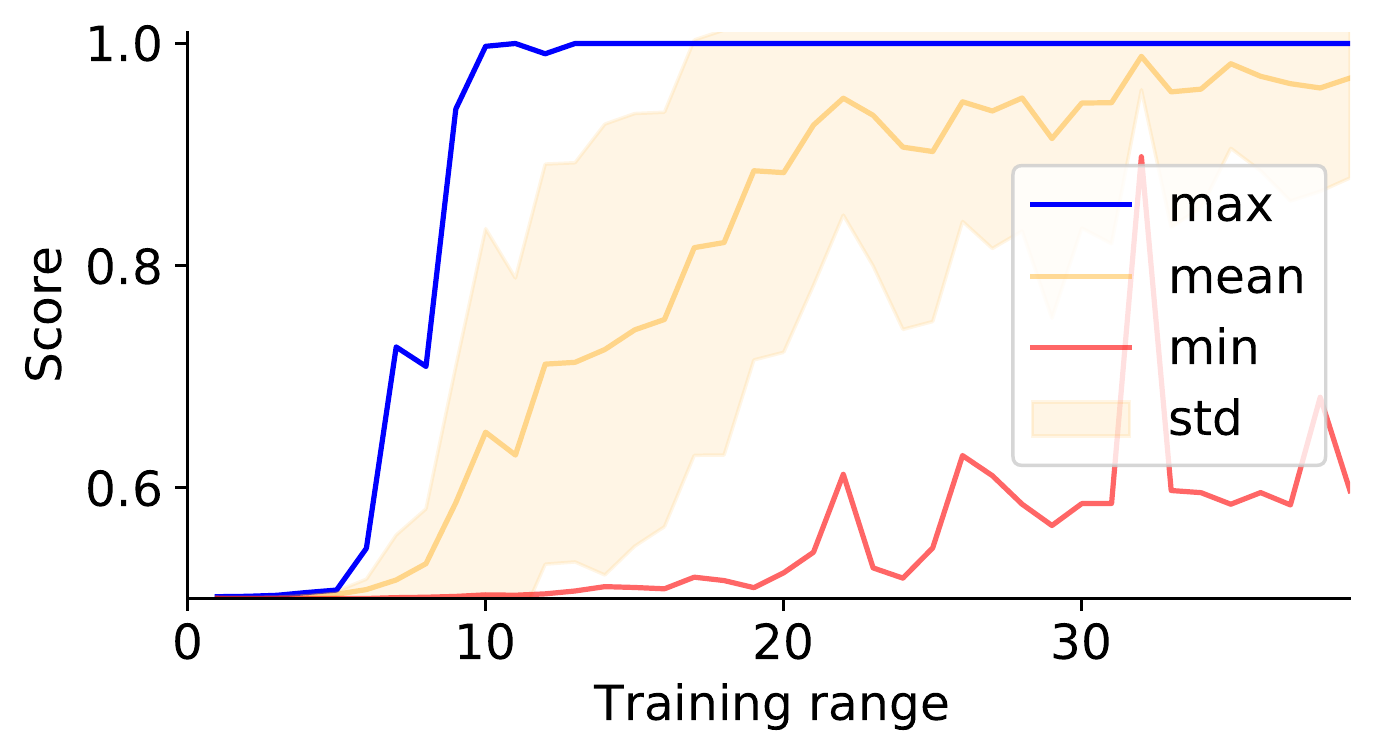}
            \end{center}
            \caption{Score per training range (min, max, and mean taken over 10 random seeds).}
            \label{fig:phase-transition:average}
        \end{subfigure}
    \end{center}
    \caption{
        Phase transition effect for the Stack-RNN on \reversestring{} (\gls*{dcf}).
        For training ranges $N < 10$, the model fails to generalize to longer sequences as it memorizes the training distribution.
        The score in \cref{fig:phase-transition:average} is the accuracy shown in \cref{fig:phase-transition:maximum} averaged over sequence lengths (x-axis).
    }
    \label{fig:phase-transition}
\end{figure}

In this section, we investigate how much training data is needed to learn the data-generating grammars.
For certain tasks and architectures we observe a substantial ``phase transition'' of the test accuracy for different training ranges.
For example, \cref{fig:phase-transition} shows the generalization performance of Stack-RNNs on the \reversestring{} (\gls*{dcf}) task for different training ranges.
Clearly, the models fail to learn the data-generating algorithm when only exposed to sequences of length smaller than 10. 
We hypothesize that for small training ranges the networks overfit to the training data, essentially learning a lookup table to recognize the small finite language.
In contrast, larger training ranges encourage the networks to learn the data-generating algorithm that generalizes to sequences of unseen length.

\subsection{Analysis of state- and memory-dynamics}
\label{sec:additional-experiments:analysis}

\subsubsection{Stack-RNN program analysis}
\label{sec:stack-rnn-sanity}

\begin{figure}
 \begin{center}
     \begin{subfigure}[b]{0.315\textwidth}
        \begin{center}
            \includegraphics[width=\textwidth]{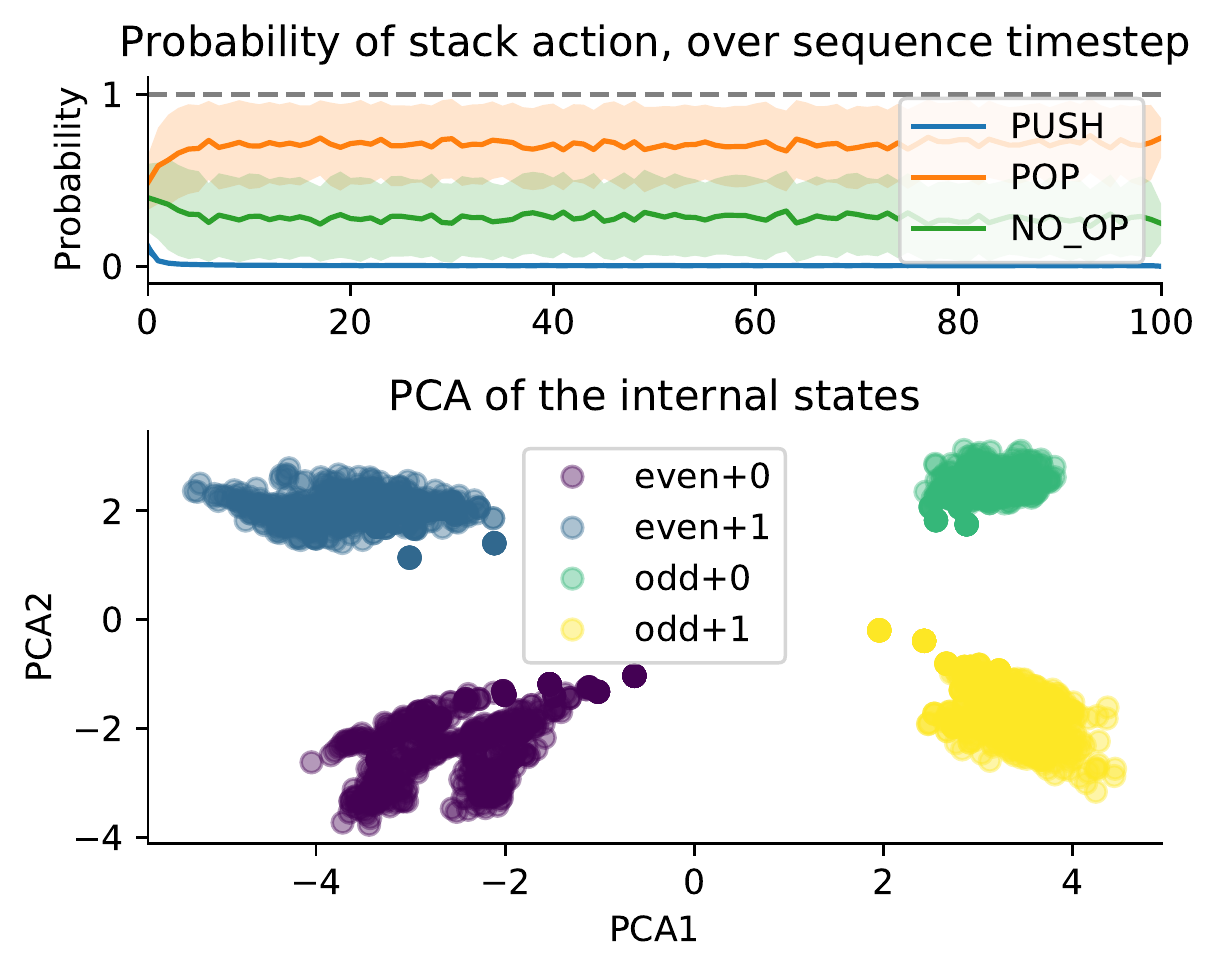}
        \end{center}
        \caption{\tiny{Stack-RNN on \paritycheck{} (\gls*{regular}).}}
        \label{fig:analysis-stack-rnn:paritycheck}
     \end{subfigure}
     \hspace{0.005\textwidth}
     \begin{subfigure}[b]{0.315\textwidth}
        \begin{center}
            \includegraphics[width=\textwidth]{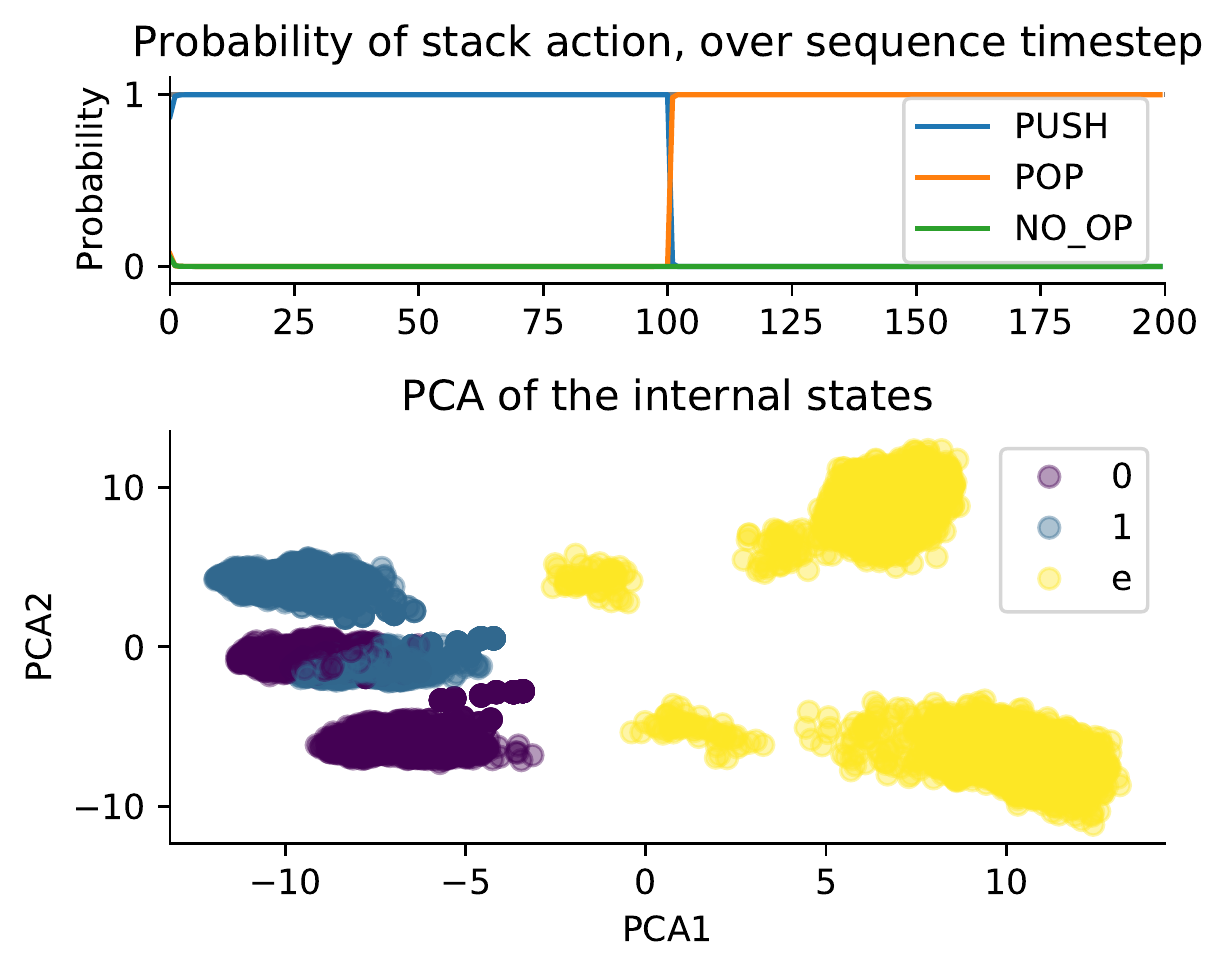}
        \end{center}
        \caption{\tiny{Stack-RNN on \reversestring{} (\gls*{dcf}).}}
        \label{fig:analysis-stack-rnn:reversestring}
     \end{subfigure}
     \begin{subfigure}[b]{0.315\textwidth}
        \begin{center}
            \includegraphics[width=\textwidth]{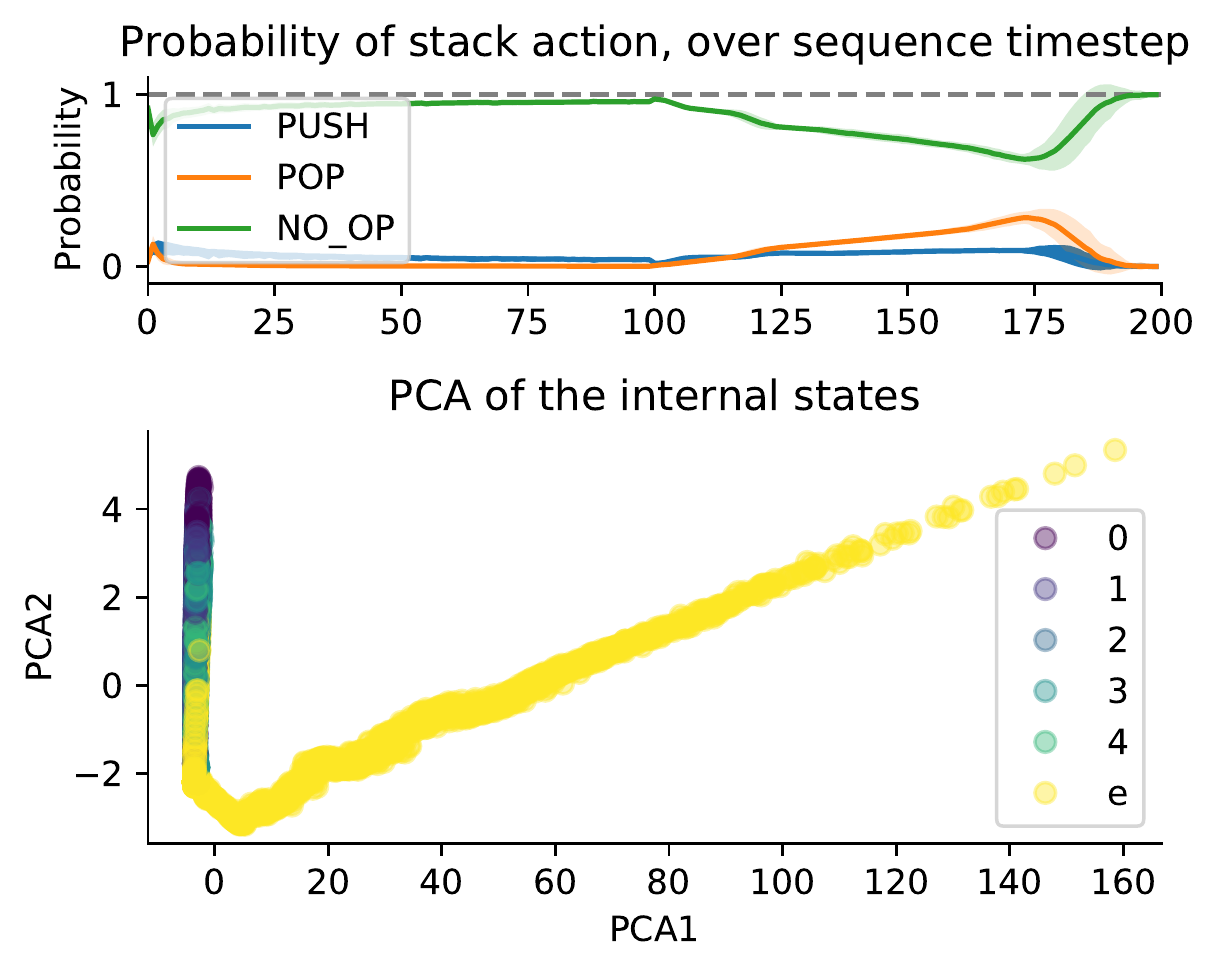}
        \end{center}
        \caption{\tiny{Stack-RNN on \bucketsort{} (\gls*{cs}).}}
        \label{fig:analysis-stack-rnn:bucketsort}
     \end{subfigure}

 \end{center}
 \caption{
    Analysis of the Stack-RNN on the \paritycheck{} (\gls*{regular}) and \reversestring{} (\gls*{dcf}) tasks.
    \emph{Left:} The network does not use the stack on \paritycheck{} and has the same state structure as a classical RNN.
    \emph{Right:} The network uses the stack on \reversestring{} and the PCA of the states shows that they are clustered according to the last input token only.
  }
 \label{fig:analysis-stack-rnn}
\end{figure}

Here, we check that the Stack-RNN, which is our simplest memory augmented network, performs as expected on some of our simple tasks.
As mentioned in \cref{sec:results:analysis}, \cref{fig:analysis-stack-rnn:paritycheck} shows that on the \paritycheck{} (\gls*{regular}) task the network ignores the stack by only applying \noop{} and \pop{} actions (visualized on an unseen sequence of length 100).
We recall that \cref{fig:analysis-rnn:states} also showed that the PCA of the states is identical to the one of a simple RNN as it only depends on the last result and the last input token.
Moreover, \cref{fig:analysis-stack-rnn:reversestring} demonstrates that on the \reversestring{} task the network always uses the \push{} action until the first empty token is reached, whereupon it starts performing \pop{} actions until the end, which exactly matches the algorithms we would expected a \gls*{pda} to implement.
Note that in this task the number of empty tokens $\varnothing$ we append to the input sequence is equal to the length of the input sequence, and thus the total sequence is of length 200.
Finally, \cref{fig:analysis-stack-rnn:bucketsort} shows that the Stack-RNN can leverage its probabilistic actions to implement multiple counters and consequently solve (counting) tasks that are beyond \acrlong*{dcf}.
Concretely, the Stack-RNN uses a mixture of \push{} and \noop{} actions to simultaneously push incremented counters while keeping the previous counter values unchanged.
Once the Stack-RNN has consumed the entire input sequence, it switches to a mixture of \pop{} and \noop{} to decrement the counters for the different tokens.
Interestingly, once the final token is reached, the Stack-RNN relies mostly on \noop{} actions, since it has learned that the remainder of the output string will contain the token of highest value.

\subsubsection{Transformer program analysis}

\paragraph{Setup}

\begin{figure}
 \begin{center}
    \includegraphics[width=\textwidth]{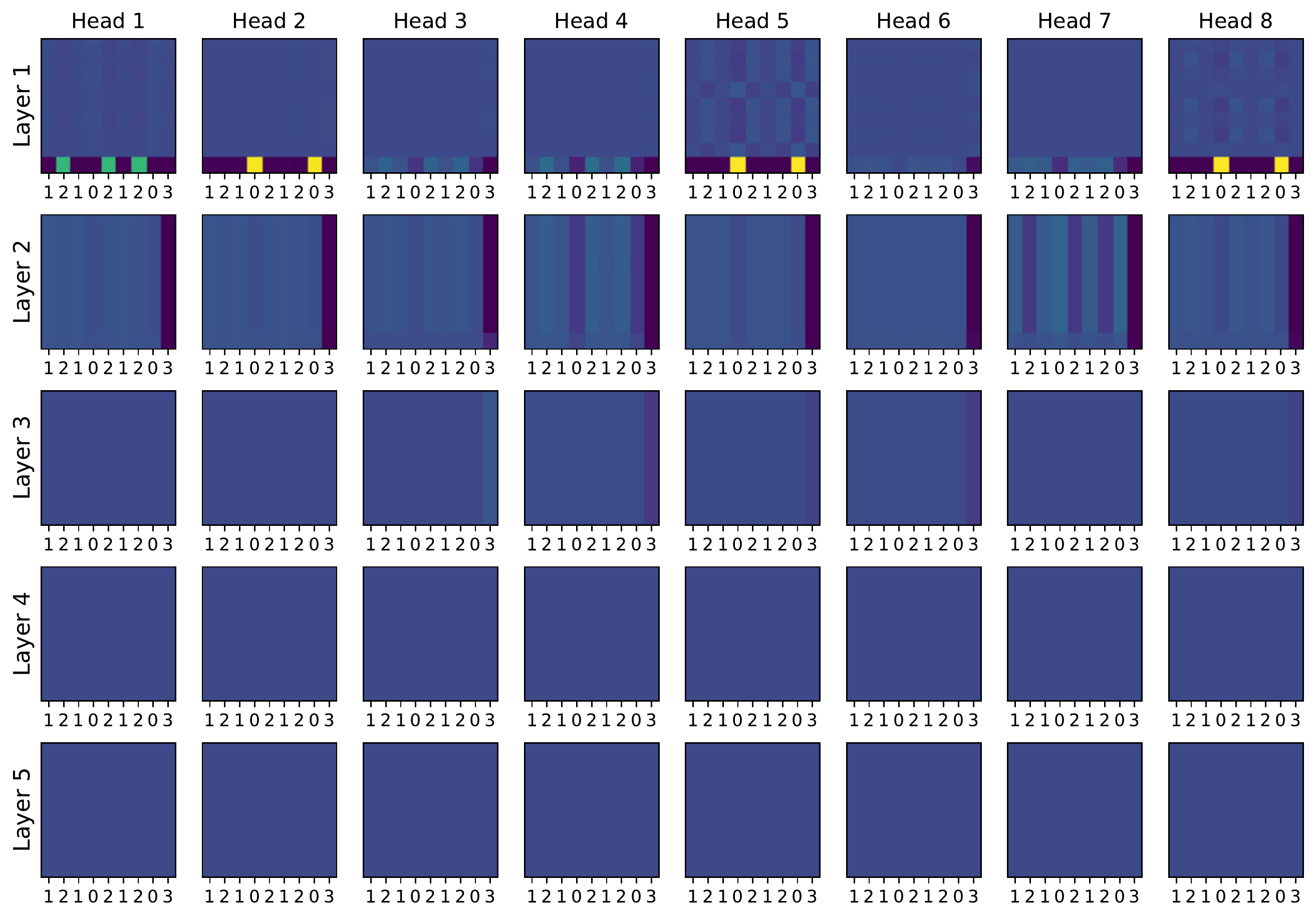}
 \end{center}
 \caption{
    Analysis of the Transformer attention matrices on the \cyclenavigation{} (\gls*{regular}) task, for an input sequence of length 8, used as ticks on the x-axis.
  }
 \label{fig:analysis-transformer-cycle-attention}
\end{figure}

The transformer achieves a non-trivial generalization score ($61.9\%$ with baseline $20\%$) on the \cyclenavigation{}~(\gls*{regular}) task but it does not `solve' the task (in terms of our definition of score $\geq 90\%$).
Therefore, we analyze the network's internals, in particular the attention matrices, to understand which algorithm the network has learned and why it fails on very long sequences.
We visualize the attention matrices in \cref{fig:analysis-transformer-cycle-attention}.
The color scale is dark blue for $0.0$ and yellow for $0.5$ (not $1.0$ to increase readability).
Our transformer networks consist of 5 layers with 8 heads, and we show all of the attention matrices (\ie $5 \cdot 8$) for a short input sequence of length 8.
Note that the actual input to the network is of length 9 since we append an empty token to the input sequence, from which we retrieve the output (see \cref{sec:methods} and \cref{sec:experimental-details}).

\paragraph{Interpretation of the learned algorithm}

\begin{figure}
 \begin{center}
    \includegraphics[width=\textwidth]{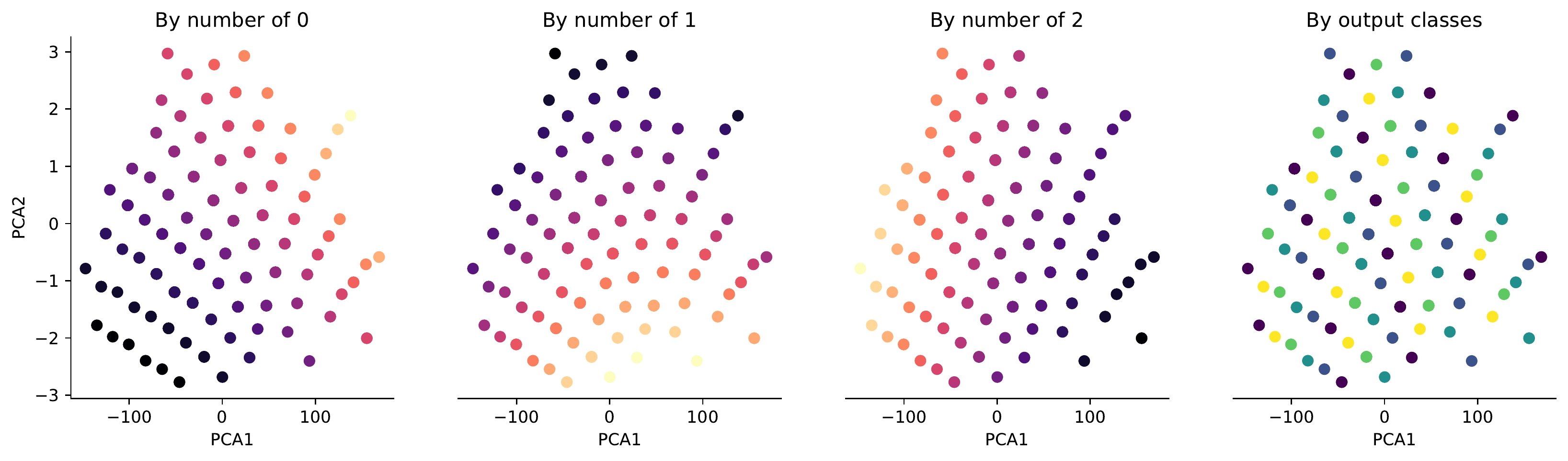}
 \end{center}
 \caption{
    Analysis of the Transformer activations after the first layer on the \cyclenavigation{} (\gls*{regular}) task, for an input sequence of length 8. We report only the first two principal components of 512 sequences. We clearly see that the network learns to count the number of 0s, 1s and 2s and already separates the output classes very well after this first layer.
  }
 \label{fig:analysis-transformer-cycle-activations-first}
\end{figure}

On the first layer, the model uses only the last empty token and matches it with the tokens 0 and 2 (clearly seen on matrices (1, 1) and (1, 2)).
Note that the task is to count the number of 0s, the number of 1s, and to do a subtraction modulo 5.
This first layer is able to separate the different tokens; the output of the last token for each head is a weighted sum (softmax) of a single value (the token which is attended to), and the weights are all equal to a constant, which means the output of the last token is just the value of the token which is attended to.
Then the other tokens have a constant weight over the whole sequence, which means that the output is a weighted sum of all the tokens' values, with the weights also being constant and proportional to the number of occurrence of each token.
This will then allow the network to compute the occurrence of each token individually in the linear layer to merge the heads.
\cref{fig:analysis-transformer-cycle-activations-first} shows a scatter plot of the activations of the network after the first layer for sequences of length 20, which confirms this claim.

\paragraph{Failure mode for long sequences}

\begin{figure}
 \begin{center}
    \includegraphics[width=\textwidth]{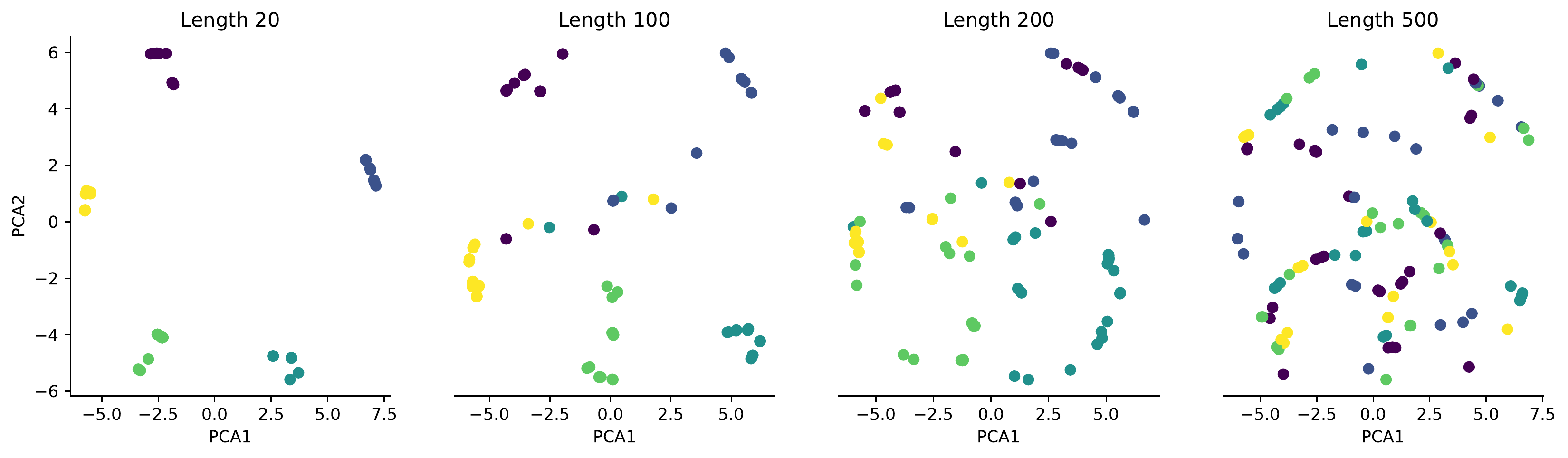}
 \end{center}
 \caption{
    Analysis of the Transformer activations after the last layer, before the final linear logit layer, on the \cyclenavigation{} (\gls*{regular}) task, for an input sequence of length 8. We report only the first two principal components of 512 sequences. We see that the activations form 5 clear clusters for short sequences (length 20), and slowly decay into  the center, mixing the clusters together.
  }
 \label{fig:analysis-transformer-cycle-activations-last}
\end{figure}

The counting mechanism, which consists in having constant weights over the whole sequence, does not return true counts but frequencies, since they are multiplied by $1/N$ where $N$ is the sequence length (since we are using a softmax).
This explains why the network can reasonably generalize to longer lengths, but not perfectly.
Looking at \cref{fig:analysis-transformer-cycle-activations-first}, it means that the structure we see is conserved for longer sequences, but the points become closer and closer, which makes further differentiation into classes more difficult. 
\cref{fig:analysis-transformer-cycle-activations-last} shows the evolution of the activations at the last layer for increasingly long sequences.
The classification becomes more challenging because the input activations do not form 5 clear clusters anymore.

\subsection{Convolutional Neural Networks}

\Citet{bansal2022endtoend} showed that fully convolutional networks achieve near-perfect length generalization on the prefix sum task in the length generalization setting.
At the same time, \citet{merrill2019sequential} theoretically showed that one-layer convolutional networks with maxpool activation functions are incapable of recognizing all regular languages.
Therefore, we investigate the length generalization performance of a simple CNN on our benchmark and visualize the maximum and average accuracies over unseen test lengths in \cref{tab:scores-other-models,tab:scores-means-variances-other-models}, respectively.
We observe that our simple architecture with kernel sizes in $\{3, 5, 7, 11\}$ (see \cref{sec:experimental-details:models}) generally fails to solve even regular tasks, achieving the random baseline accuracy in most cases.
This confirms the theoretical derivations by \citet{merrill2019sequential} but contrasts the results presented by \citet{bansal2022endtoend}.
However, the convolutional architecture that \citet{bansal2022endtoend} consider relies on an adaptive number of layers depending on the sequence length (similar to adaptive computation time \citep{graves2016adaptive}) and thus crucially differs from a classical CNN (such as the one we study).

\subsection{Stack-LSTMs}

We evaluate the experiment suite from \cref{sec:results:main-result} for the Stack-LSTMs and compare their performance with that of Stack-RNNs in \cref{tab:scores-other-models}.
We observe that Stack-LSTMs and Stack-RNNs perform roughly equally well when considering the score (maximum accuracy over the 10 initialization seeds), which is not surprising given that both networks are theoretically located on the same level of the Chomsky hierarchy.
However, when considering the average Stack-LSTMs perform worse, \ie our training protocol (gradient descent) is less stable for this architecture, confirming the results of prior work~\citep{hao2018context}.
We hypothesize that this is due to the fact that LSTMs are more powerful than RNNs (as they can implement k-counter machines) and thus do not always simulate a \acrlong*{fsa}.
This can be an advantage for more difficult tasks, \eg counting tasks such as \bucketsort{} (\gls*{cs}), but also a disadvantage for simpler tasks that only need a \acrlong*{fsa} controller to interact with the memory structure.
Thus, if the controller is not a \acrlong*{fsa}, then its internal state does not interact well with the stack content as it is not part of a finite set, \ie the interactions are potentially infinite and thus the network does not generalize well beyond the training range.

\subsection{NDStack-RNN}

We also compared the Stack-RNN and the Stack-LSTM to their non-deterministic counterpart, the NDStack-RNN (which stands for Non Deterministic Stack-RNN) \citep{dusell2020learning, dusell2022learning}. We observe that this architecture performs well on regular tasks, as the other memory-augmented RNNs. However, it fails on most tasks beyond
regular, potentially due to its large action space (24 actions) and the difficulty of training a tensor
that tracks all possible stack values. We also don't use unnormalized actions as it requires computing all the probabilities in log-space, and unfortunately JAX \citep{bradbury2018jax} does not support complex tensor multiplications like $einsum$ in log-space. Surprisingly, the architecture solves the \missingduplicate{}~(\gls*{cs}) task perfectly. We tried some ablations and reduced the stack size to only 1: the model still solves the task perfectly. Also, the actions taken by the model on the stack are not binary, leading to mixed computation. We conjecture that the network uses this mixed embedding of values on the stack (reduced to a single cell) efficiently as a memory, on top of the memory of the RNN controller.

\subsection{Autoregressive Models}
\label{sec:additional-results:autoregressive}

In this paper, we chose to predict the output sequences from empty tokens, but we could also have chosen to predict the sequences autoregressively.
Here, we conduct the same set of experiments but in the autoregressive setting to confirm that this aspect of our experimental setup does not impact the architectures' positions in the Chomsky hierarchy.
To that end, we use the true output sequence as the input to the model at train time and autoregressive sampling at test time, \ie we use the predicted output token as the input for the next token until we reach the desired length (more details below for the different architectures).
We show the results in \cref{tab:scores-other-models} with the corresponding means and standard deviations in \cref{tab:scores-means-variances-other-models}.

\subsubsection{Memory-augmented recurrent neural networks}

For the (memory-augmented) RNNs, \ie RNN, LSTM, Stack-RNN and Tape-RNN, we replace the empty tokens $\varnothing$ with the true output sequence during training.
However, to avoid predicting the input, we append an extra empty token between the input sequence (which could contain computation steps) and the true output sequence.
We compute the loss with an offset of 1, so the network should predict the first token of the output sequence for the extra empty token, then the second token of the output sequence for the first token, passed now as input, etc.
We observe in \cref{tab:scores-other-models} that these models trained with the auto-regressive system perform very similarly to the ones trained with the empty tokens system (see \cref{tab:scores} for comparison).
That is, this part of the training scheme does not affect the length generalization scores, and therefore does not change where the networks lie on the Chomsky hierarchy.

\subsubsection{Transformer}

In \cref{sec:results}, we compared the Transformer encoder (used in a non-autoregressive manner) to other sequence prediction models since this allowed for a fair comparison in terms of the experiment setup (all models had to predict the output from empty tokens $\varnothing$).
However, the original Transformer architecture~\citep{vaswani2017attention} also consists of a decoder and processes sequences autoregressively, which is what we evaluate in this section.
Similar to memory-augmented RNNs, we observe that the autoregressive model performs roughly as well as the Transformer encoder, except that it fails to solve (or show strong performance) on the permutation-invariant task \bucketsort{}~(\gls*{cs}).
This is likely due to the causal masking in the decoder attention, which eliminates permutation-invariance.
However, the autoregressive model is capable of solving the \duplicatestring{}~(\gls*{cs}) perfectly.
We hypothesize that this is due to the fact that the relevant input and output tokens are at a fixed length throughout the task and thus the model manages to align the input and output embeddings (using relative positional encodings).

\begin{table}
  \caption{
    Score (in percentage, see \cref{sec:methods}), \ie accuracy averaged over all test lengths and maximized over 10 random seeds (and other hyperparameters, see \cref{sec:experimental-details}), for the CNN, Stack-LSTM, and NDStack-RNN.
    The results for the Stack-RNN are identical to \cref{tab:scores} and are included only for easier comparison with the Stack-LSTM.
    We consider a model to generalize successfully (bold) if its $\mathrm{score} \geq 90\%$.
    The random accuracy is $50\%$ (except for \cyclenavigation{}~(\gls*{regular}), \bucketsort{} (\gls*{cs}), and the two modular arithmetic tasks where it is $20\%$).
    We denote permutation-invariant tasks with $\dagger$, counting tasks with $\star$, tasks requiring a nondeterministic controller with $\circ$, and tasks requiring superlinear running time with $\times$.
}
  \label{tab:scores-other-models}
  \begin{center}
    \resizebox{\columnwidth}{!}{\begin{tabular}{@{}llrrrr@{}}
    \toprule
    \textbf{Level} & \textbf{Tasks} & \textbf{CNN} & \textbf{Stack-RNN} & \textbf{Stack-LSTM} & \textbf{NDStack-RNN} \\
    \midrule
    \multirow{4}{*}{\Gls*{regular}}
    & \evenpairs                        & 50.1 & \textbf{100.0} & \textbf{100.0} & \textbf{100.0} \\
    & \modulararithmeticsimple          & 20.1 & \textbf{100.0} & \textbf{100.0} & \textbf{100.0} \\
    & \paritycheck$^\dagger$            & 50.0 & \textbf{100.0} & \textbf{100.0} & \textbf{100.0} \\
    & \cyclenavigation$^\dagger$        & 20.1 & \textbf{100.0} & \textbf{100.0} & \textbf{100.0} \\
    \\
    \multirow{4}{*}{\Gls*{dcf}}
    & \stackmanipulation                & 52.4 & \textbf{100.0} & \textbf{100.0} & 59.2 \\
    & \reversestring                    & 53.0 & \textbf{100.0} & \textbf{100.0} & 62.4 \\
    & \modulararithmeticbrackets        & 31.2 & \textbf{96.1} & 61.0 & 35.8 \\
    & \solveequation$^\circ$            & 20.1 & 56.2 & 67.5 & 47.5 \\
    \\
    \multirow{7}{*}{\Gls*{cs}}
    & \duplicatestring                  & 50.0 & 52.8 & 59.0 & 51.0 \\
    & \missingduplicate                 & 51.0 & 55.2 & 55.1 & \textbf{100.0} \\
    & \oddsfirst                        & 50.0 & 51.9 & 55.5 & 53.1 \\
    & \binaryaddition                   & 49.8 & 52.7 & 56.0 & 53.5 \\
    & \binarymultiplication$^\times$    & 49.9 & 52.7 & 53.3 & 50.9 \\
    & \computesqrt                      & 50.2 & 56.5 & 57.6 & 55.0 \\
    & \bucketsort$^{\dagger\bigstar}$   & 29.1 & 78.1 & \textbf{99.2} & 41.7 \\
    \bottomrule
\end{tabular}
}
  \end{center}
\end{table}

\begin{table}
  \caption{
    Means and standard deviations (computed over random seeds) of the score (average test accuracy, see \cref{sec:methods}) for the results of \cref{tab:scores-other-models}.
    The results for the Stack-RNN are identical to \cref{tab:scores-means-variances} and are included only for easier comparison with the Stack-LSTM.
    The $\dagger$ denotes permutation-invariant tasks; the $\star$ denotes counting tasks; the $\circ$ denotes tasks that require a nondeterministic controller; and the $\times$ denotes tasks that require superlinear running time in terms of the input length.
  }
  \label{tab:scores-means-variances-other-models}
  \begin{center}
    \resizebox{\columnwidth}{!}{\begin{tabular}{@{}llrrrr@{}}
    \toprule
    \textbf{Level} & \textbf{Tasks} & \textbf{CNN} & \textbf{Stack-RNN} & \textbf{Stack-LSTM} & \textbf{NDStack-RNN} \\
    \midrule
    \multirow{4}{*}{\Gls*{regular}}
    & \evenpairs                        & $50.1 \pm 0.0$ & $\textbf{100.0} \pm 0.0$ & $\textbf{100.0} \pm 0.0$ & $98.0 \pm 5.7$ \\ 
    & \modulararithmeticsimple          & $20.0 \pm 0.1$ & $\textbf{100.0} \pm 0.0$ & $\textbf{100.0} \pm 0.0$ & $78.4 \pm 28.8$ \\
    & \paritycheck$^\dagger$            & $50.0 \pm 0.0$ & $\textbf{100.0} \pm 0.0$ & $\textbf{100.0} \pm 0.0$ & $95.6 \pm 13.6$ \\
    & \cyclenavigation$^\dagger$        & $20.0 \pm 0.1$ & $\textbf{96.8} \pm 6.6$ & $77.7 \pm 9.5$ & $95.0 \pm 9.1$ \\
    \\
    \multirow{4}{*}{\Gls*{dcf}}
    & \stackmanipulation                & $52.4 \pm 0.0$ & $\textbf{93.8} \pm 15.1$ & $66.2 \pm 17.4$ & $55.1 \pm 2.5$ \\
    & \reversestring                    & $53.0 \pm 0.1$ & $\textbf{96.2} \pm 12.1$ & $\textbf{94.0} \pm 12.4$ & $61.6 \pm 0.6$ \\
    & \modulararithmeticbrackets        & $30.7 \pm 0.5$ & $75.8 \pm 11.3$ & $51.2 \pm 5.0$ & $28.4 \pm 4.1$ \\
    & \solveequation$^\circ$            & $19.9 \pm 0.1$ & $46.1 \pm 7.3$ & $42.7 \pm 16.0$ & $34.4 \pm 9.7$ \\
    \\
    \multirow{7}{*}{\Gls*{cs}}
    & \duplicatestring                  & $50.0 \pm 0.0$ & $51.6 \pm 0.6$ & $56.3 \pm 1.3$ & $50.6 \pm 0.2$ \\
    & \missingduplicate                 & $50.6 \pm 0.1$ & $53.6 \pm 0.7$ & $53.3 \pm 0.7$ & $82.0 \pm 23.3$ \\
    & \oddsfirst                        & $50.0 \pm 0.0$ & $51.5 \pm 0.1$ & $54.4 \pm 0.7$ & $52.2 \pm 0.8$ \\
    & \binaryaddition                   & $49.7 \pm 0.2$ & $50.9 \pm 0.9$ & $55.1 \pm 0.6$ & $50.4 \pm 1.3$ \\
    & \binarymultiplication$^\times$    & $49.8 \pm 0.0$ & $51.4 \pm 0.7$ & $52.9 \pm 0.3$ & $50.0 \pm 0.4$ \\
    & \computesqrt                      & $50.2 \pm 0.0$ & $56.1 \pm 0.3$ & $57.3 \pm 0.2$ & $54.3 \pm 0.4$ \\
    & \bucketsort$^{\dagger\bigstar}$   & $28.4 \pm 0.6$ & $56.6 \pm 16.2$ & $81.7 \pm 19.9$ & $25.9 \pm 7.2$ \\
    \bottomrule
\end{tabular}
}
  \end{center}
\end{table}

\begin{table}
  \caption{
    Score (in percentage, see \cref{sec:methods}), \ie accuracy averaged over all test lengths and maximized over 10 random seeds (and other hyperparameters, see \cref{sec:experimental-details}), for autoregressive setting.
    We only include the tasks with output length larger than 1, since the results are identical to those in \cref{tab:scores} otherwise.
    We consider a model to generalize successfully (bold) if its $\mathrm{score} \geq 90\%$.
    The random accuracy is $50\%$ (except for \cyclenavigation{}~(\gls*{regular}), \bucketsort{} (\gls*{cs}), and the two modular arithmetic tasks where it is $20\%$).
    We denote permutation-invariant tasks with $\dagger$, counting tasks with $\star$, tasks requiring a nondeterministic controller with $\circ$, and tasks requiring superlinear running time with $\times$.
}
  \label{tab:scores-autoregressive}
  \begin{center}
    \resizebox{0.85\columnwidth}{!}{\begin{tabular}{@{}llrrrr@{}}
    \toprule
    \textbf{Level} & \textbf{Tasks} & \textbf{RNN} & \textbf{Stack-RNN} & \textbf{Transformer}  & \textbf{LSTM} \\
    \midrule
    \multirow{2}{*}{\Gls*{dcf}}
    & \stackmanipulation                & 58.0 & \textbf{100.0} & 53.2 & 56.9 \\
    & \reversestring                    & 63.1 & \textbf{100.0} & 53.5 & 59.4 \\
    \\
    \multirow{6}{*}{\Gls*{cs}}
    & \duplicatestring                  & 50.3 & 50.7 & \textbf{100.0} & 51.7 \\
    & \oddsfirst                        & 50.5 & 51.4 & 54.7 & 53.6 \\
    & \binaryaddition                   & 49.8 & 51.7 & 69.0 & 53.9 \\
    & \binarymultiplication$^\times$    & 49.5 & 52.0 & 52.2 & 53.3 \\
    & \computesqrt                      & 54.1 & 57.8 & 52.4 & 58.0 \\
    & \bucketsort$^{\dagger\bigstar}$   & 28.9 & 52.8 & 40.4 & \textbf{92.9} \\
    \bottomrule
\end{tabular}
}
  \end{center}
\end{table}

\begin{table}
  \caption{
    Means and standard deviations (computed over random seeds) of the score (average test accuracy, see \cref{sec:methods}) for the results in the autoregressive setting (\ie \cref{tab:scores-autoregressive}).
    We only include the tasks with output length larger than 1, since the results are identical to those in \cref{tab:scores} otherwise.
    The $\dagger$ denotes permutation-invariant tasks; the $\star$ denotes counting tasks; the $\circ$ denotes tasks that require a nondeterministic controller; and the $\times$ denotes tasks that require superlinear running time in terms of the input length.
  }
  \label{tab:scores-means-variances-autoregressive}
  \begin{center}
    \resizebox{0.9\columnwidth}{!}{\begin{tabular}{@{}llrrrr@{}}
    \toprule
    \textbf{Level} & \textbf{Tasks} & \textbf{RNN} & \textbf{Stack-RNN} & \textbf{Transformer}  & \textbf{LSTM} \\
    \midrule
    \multirow{2}{*}{\Gls*{dcf}}
    & \stackmanipulation                & $56.7 \pm 0.2$ & $\bm{100.0} \pm 0.00$ & $51.7 \pm 1.3$ & $55.8 \pm 0.1$ \\
    & \reversestring                    & $62.4 \pm 0.1$ & $\bm{99.3} \pm 0.2$ & $52.3 \pm 1.6$ & $58.1 \pm 0.1$ \\
    \\
    \multirow{6}{*}{\Gls*{cs}}
    & \duplicatestring                  & $50.2 \pm 0.1$ & $50.5 \pm 0.1$ & $\bm{92.7} \pm 4.7$ & $51.4 \pm 0.1$ \\
    & \oddsfirst                        & $50.4 \pm 0.1$ & $51.0 \pm 0.02$ & $52.3 \pm 1.4$ & $52.9 \pm 0.4$ \\
    & \binaryaddition                   & $49.4 \pm 0.2$ & $50.5 \pm 0.5$ & $60.2 \pm 4.2$ & $52.9 \pm 0.6$ \\
    & \binarymultiplication$^\times$    & $47.5 \pm 0.5$ & $50.5 \pm 1.3$ & $50.8 \pm 1.3$ & $52.5 \pm 0.5$ \\
    & \computesqrt                      & $53.7 \pm 0.3$ & $57.5 \pm 0.3$ & $54.5 \pm 0.5$ & $57.7 \pm 0.5$ \\
    & \bucketsort$^{\dagger\bigstar}$   & $21.8 \pm 3.4$ & $35.6 \pm 8.7$ & $32.1 \pm 4.3$ & $\bm{93.0} \pm 6.7$ \\
    \bottomrule
\end{tabular}
}
  \end{center}
\end{table}

\begin{figure}
\begin{center}
        \begin{subfigure}[b]{0.31\textwidth}
            \begin{center}
                \includegraphics[width=\textwidth]{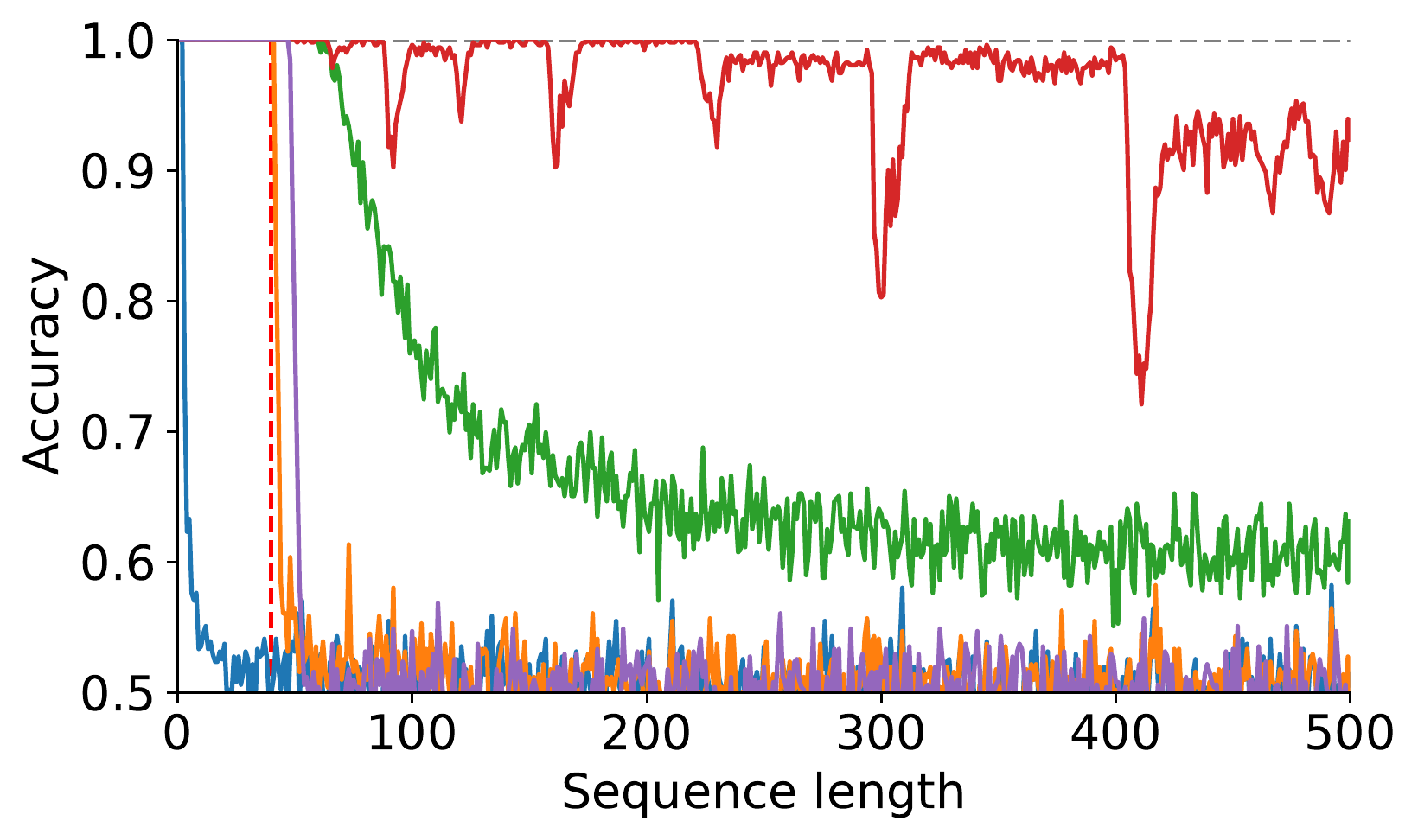}
            \end{center}
            \vspace{-0.25cm}
            \caption{\tiny{\evenpairs{} (\gls*{regular})}}
        \end{subfigure}
        \hspace{0.01\textwidth}
        \begin{subfigure}[b]{0.31\textwidth}
            \begin{center}
                \includegraphics[width=\textwidth]{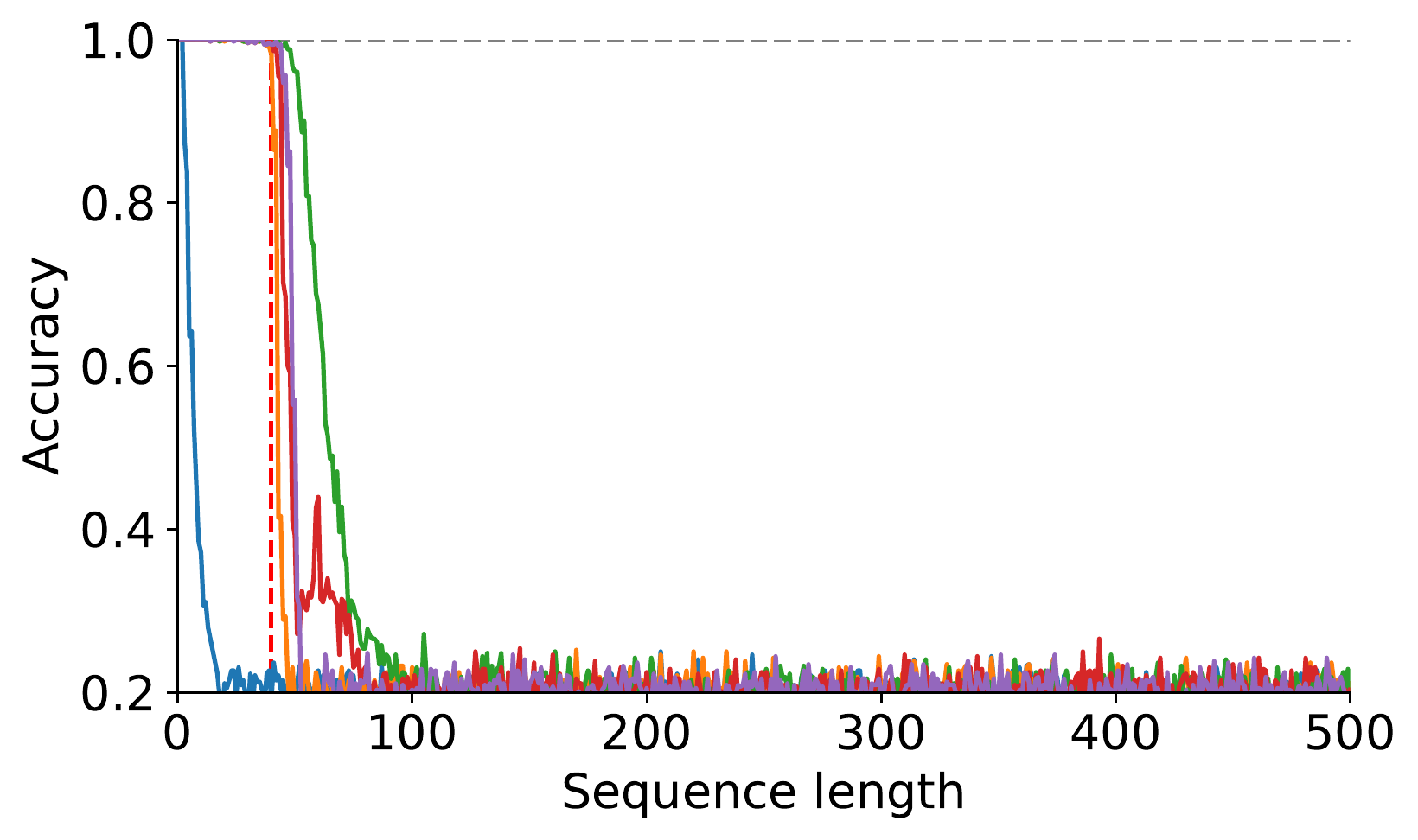}
            \end{center}
            \vspace{-0.25cm}
            \caption{\tiny{\modulararithmeticsimple{} (\gls*{regular})}}
        \end{subfigure}
        \hspace{0.01\textwidth}
        \begin{subfigure}[b]{0.31\textwidth}
            \begin{center}
                \includegraphics[width=\textwidth]{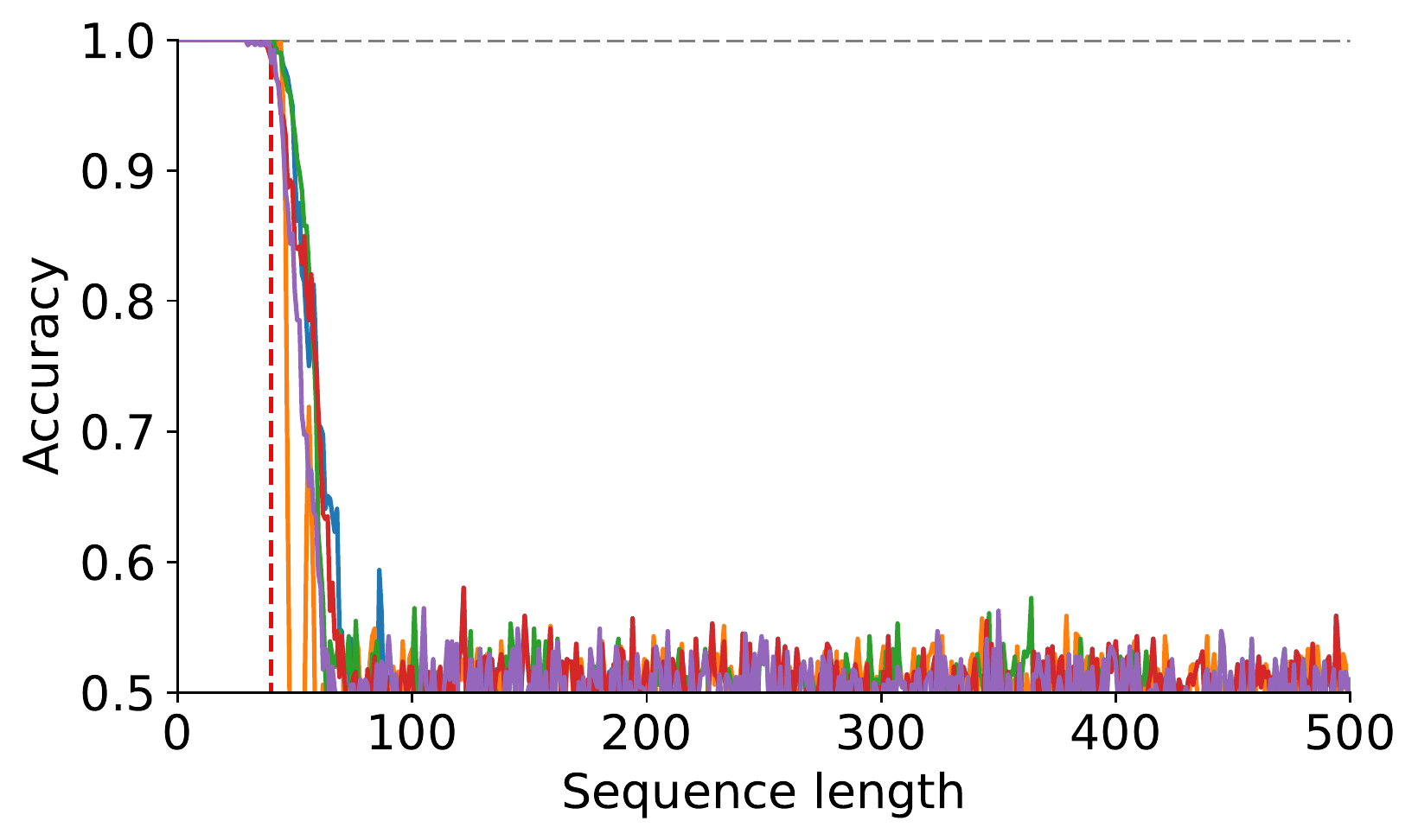}
            \end{center}
            \vspace{-0.25cm}
            \caption{\tiny{\paritycheck{} (\gls*{regular})}}
        \end{subfigure}
        \vspace{0.5cm}
        
        \begin{subfigure}[b]{0.31\textwidth}
            \begin{center}
                \includegraphics[width=\textwidth]{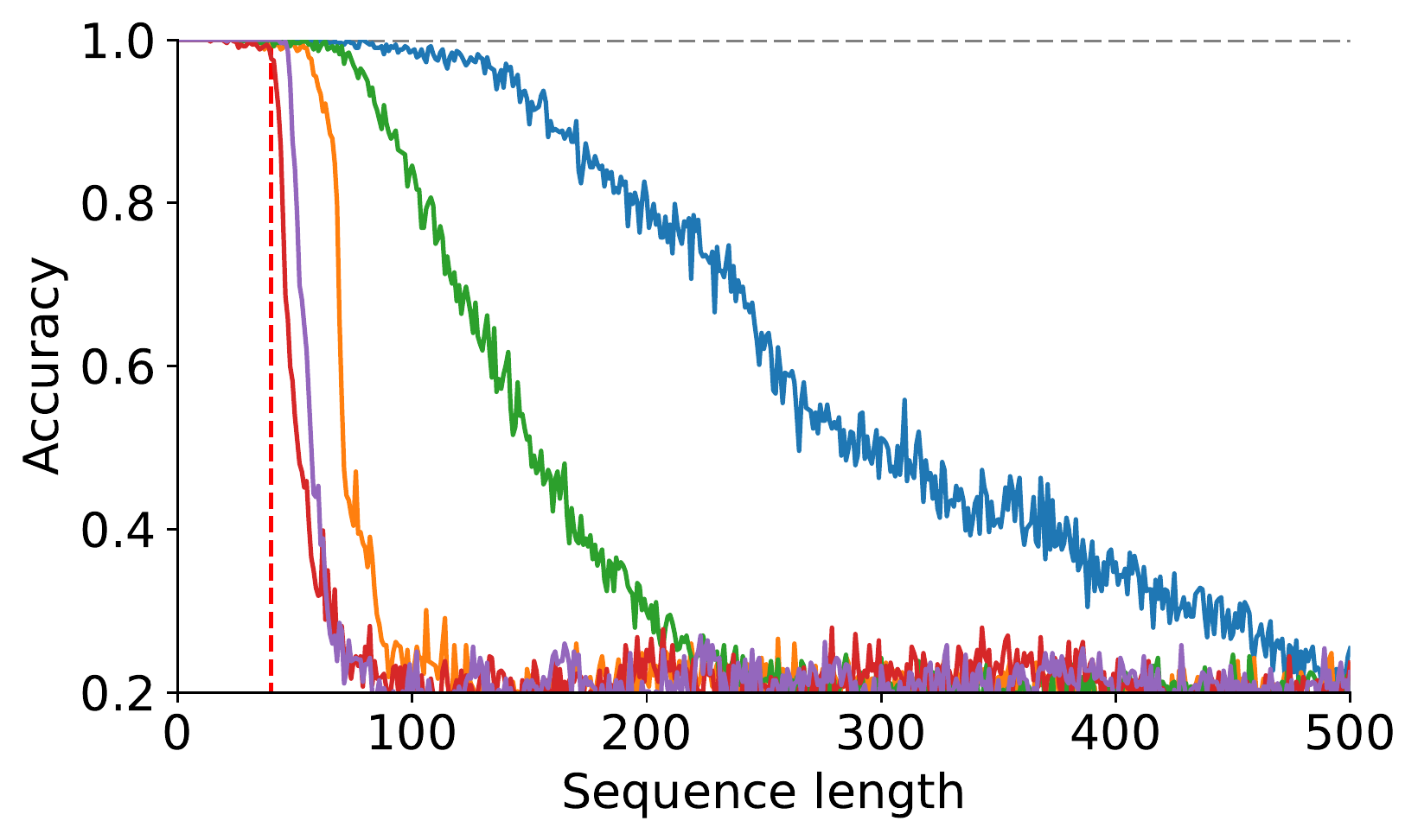}
            \end{center}
            \vspace{-0.25cm}
            \caption{\tiny{\cyclenavigation{} (\gls*{regular})}}
        \end{subfigure}
        \hspace{0.01\textwidth}
        \begin{subfigure}[b]{0.31\textwidth}
            \begin{center}
                \includegraphics[width=\textwidth]{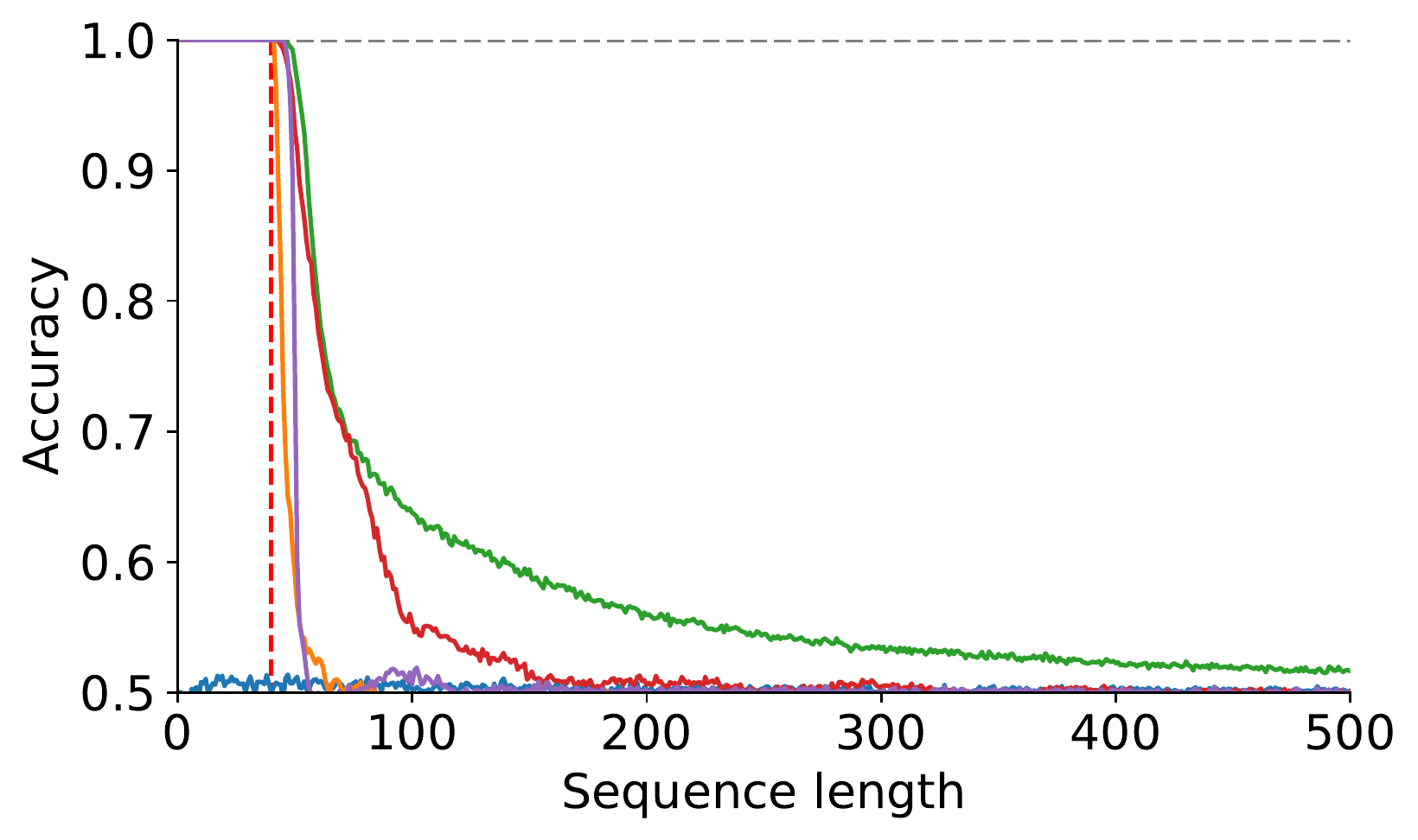}
            \end{center}
            \vspace{-0.25cm}
            \caption{\tiny{\stackmanipulation{} (\gls*{dcf})}}
        \end{subfigure}
        \hspace{0.01\textwidth}
        \begin{subfigure}[b]{0.31\textwidth}
            \begin{center}
                \includegraphics[width=\textwidth]{figures/pos_enc_reverse_string-eps-converted-to.pdf}
            \end{center}
            \vspace{-0.25cm}
            \caption{\tiny{\reversestring{} (\gls*{dcf})}}
        \end{subfigure}
        \vspace{0.5cm}
        
        \begin{subfigure}[b]{0.31\textwidth}
            \begin{center}
                \includegraphics[width=\textwidth]{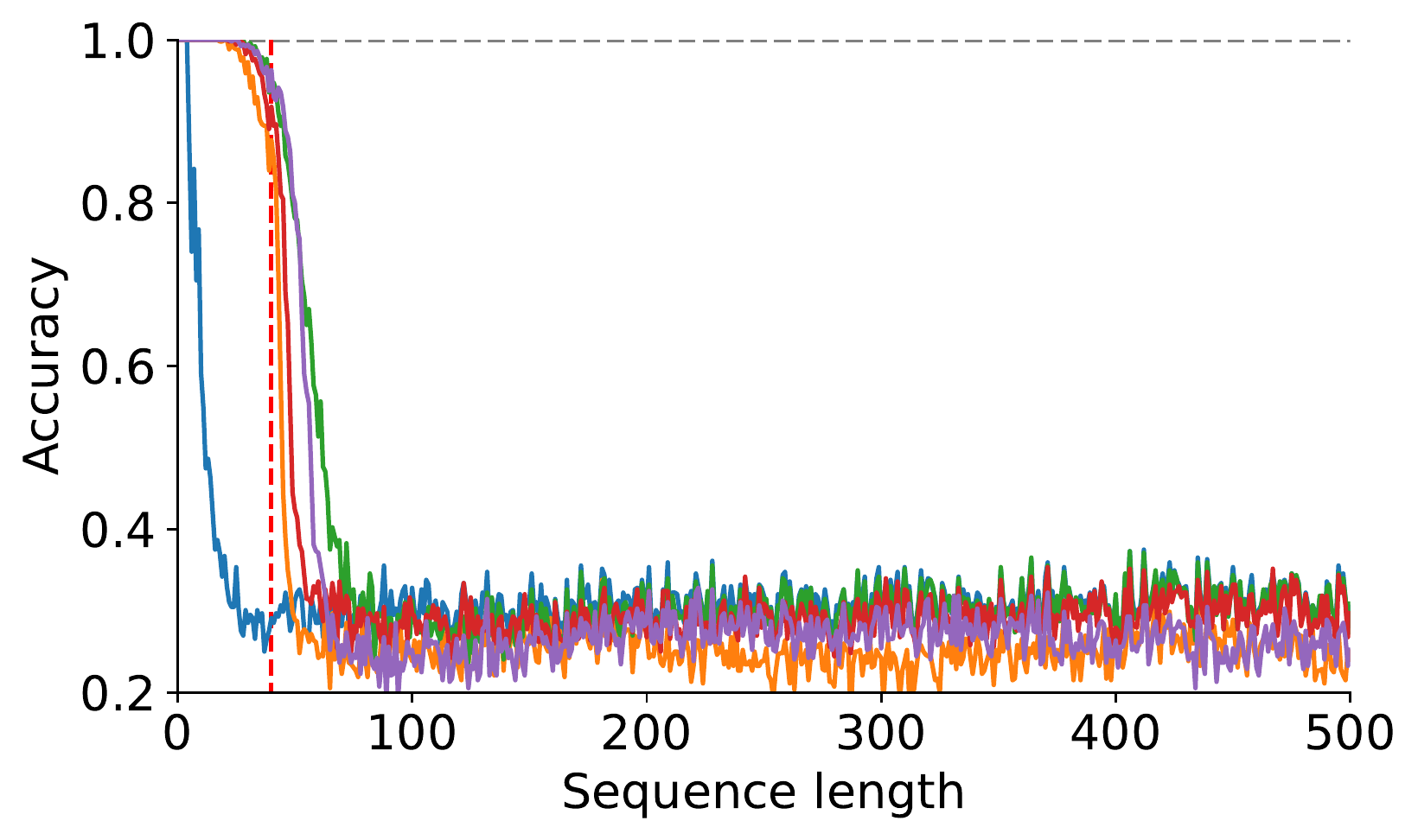}
            \end{center}
            \vspace{-0.25cm}
            \caption{\tiny{\modulararithmeticbrackets{} (\gls*{dcf})}}
        \end{subfigure}
        \hspace{0.01\textwidth}
        \begin{subfigure}[b]{0.31\textwidth}
            \begin{center}
                \includegraphics[width=\textwidth]{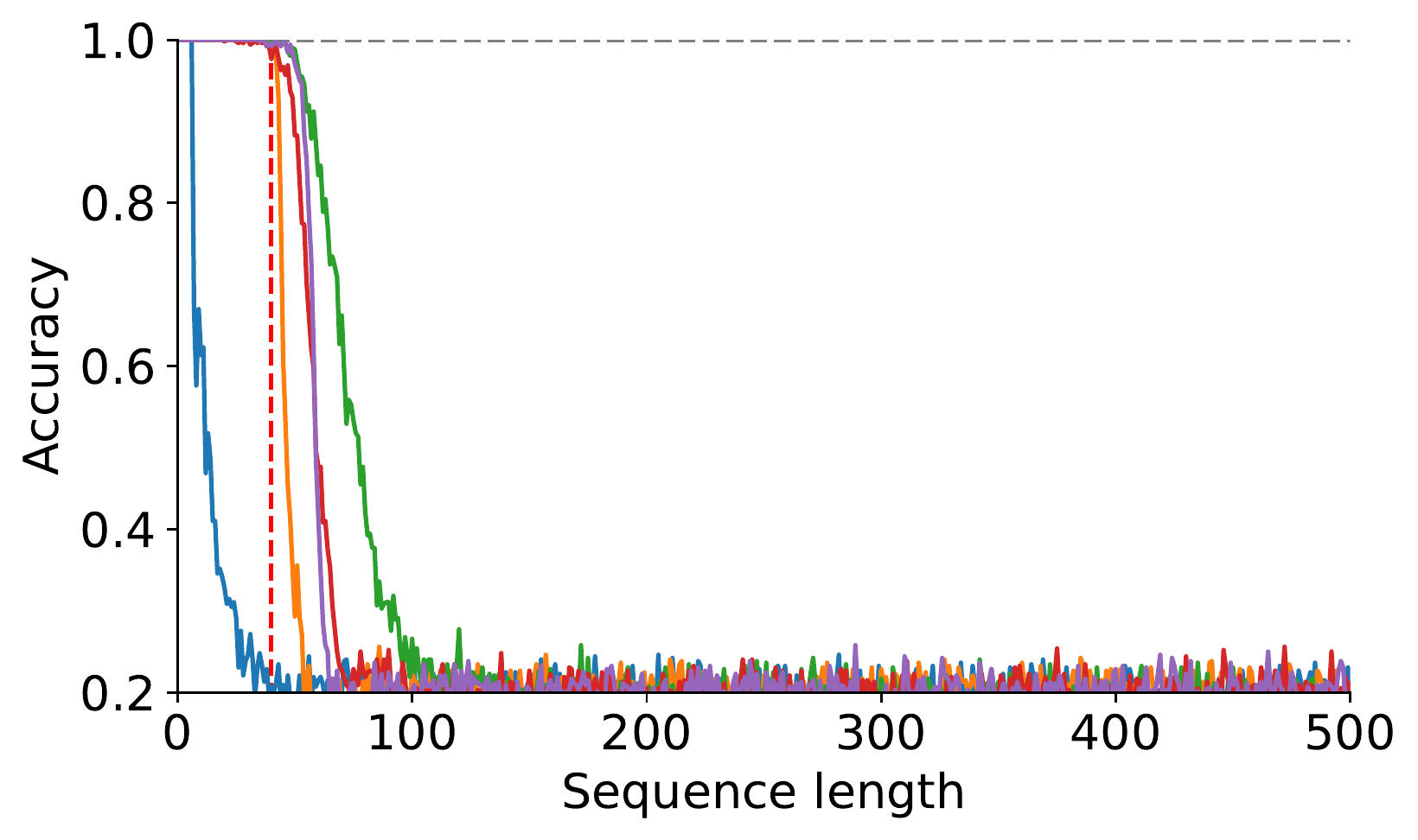}
            \end{center}
            \vspace{-0.25cm}
            \caption{\tiny{\solveequation{} (\gls*{dcf})}}
        \end{subfigure}
        \hspace{0.01\textwidth}
        \begin{subfigure}[b]{0.31\textwidth}
            \begin{center}
                \includegraphics[width=\textwidth]{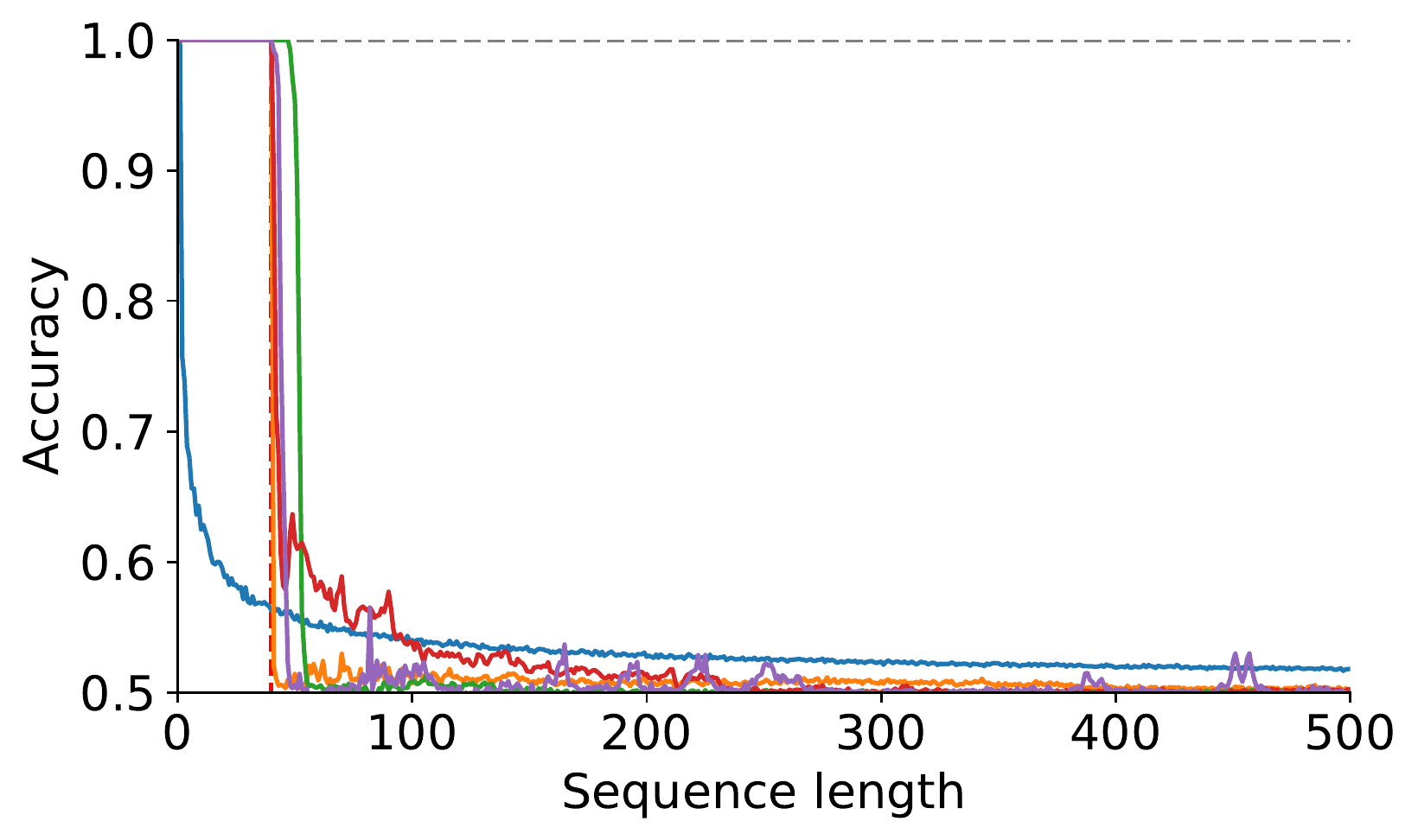}
            \end{center}
            \vspace{-0.25cm}
            \caption{\tiny{\duplicatestring{} (\gls*{cs})}}
        \end{subfigure}
        \hspace{0.01\textwidth}
        \vspace{0.5cm}
        
        \begin{subfigure}[b]{0.31\textwidth}
            \begin{center}
                \includegraphics[width=\textwidth]{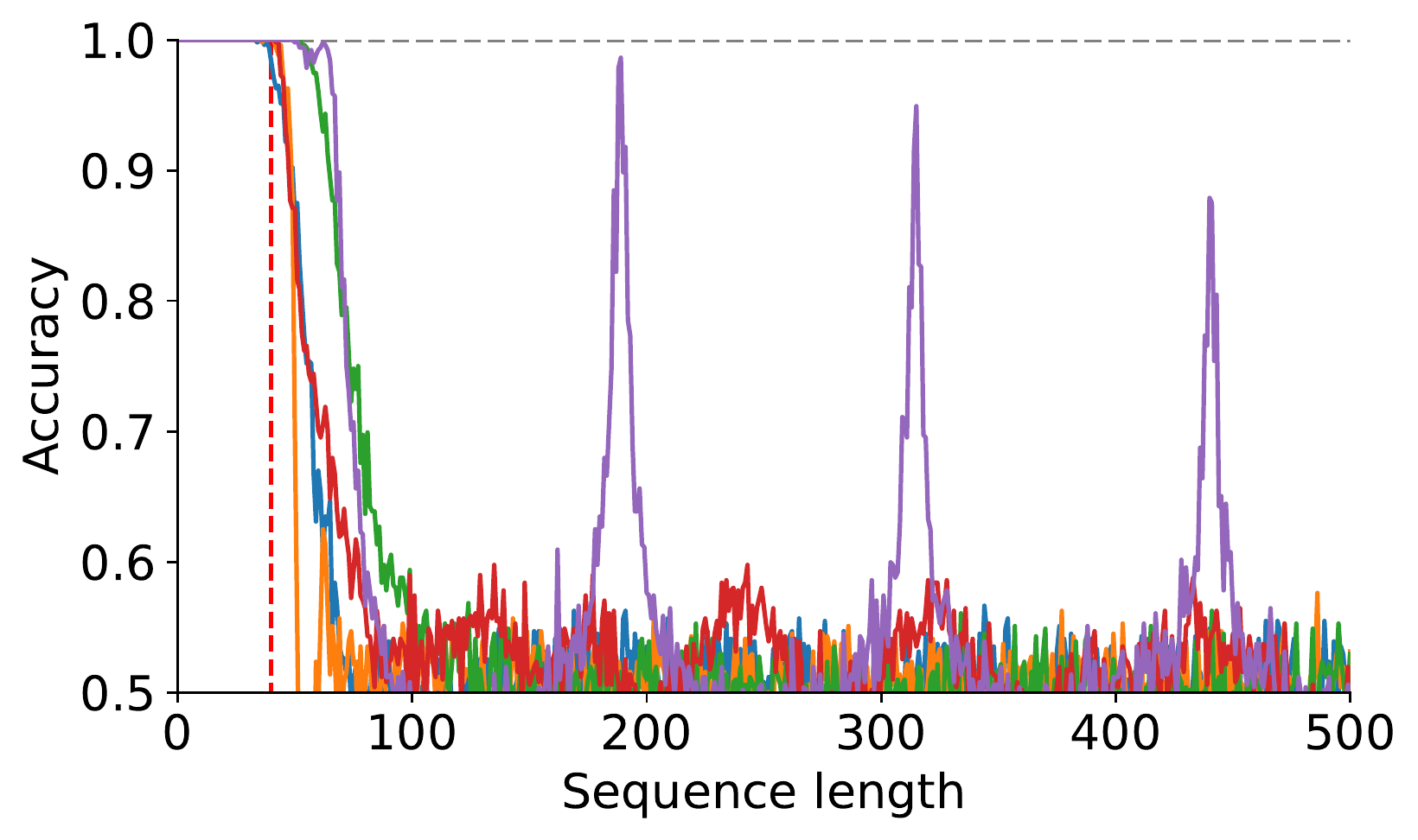}
            \end{center}
            \vspace{-0.25cm}
            \caption{\tiny{\missingduplicate{} (\gls*{cs})}}
        \end{subfigure}
        \hspace{0.01\textwidth}
        \begin{subfigure}[b]{0.31\textwidth}
            \begin{center}
                \includegraphics[width=\textwidth]{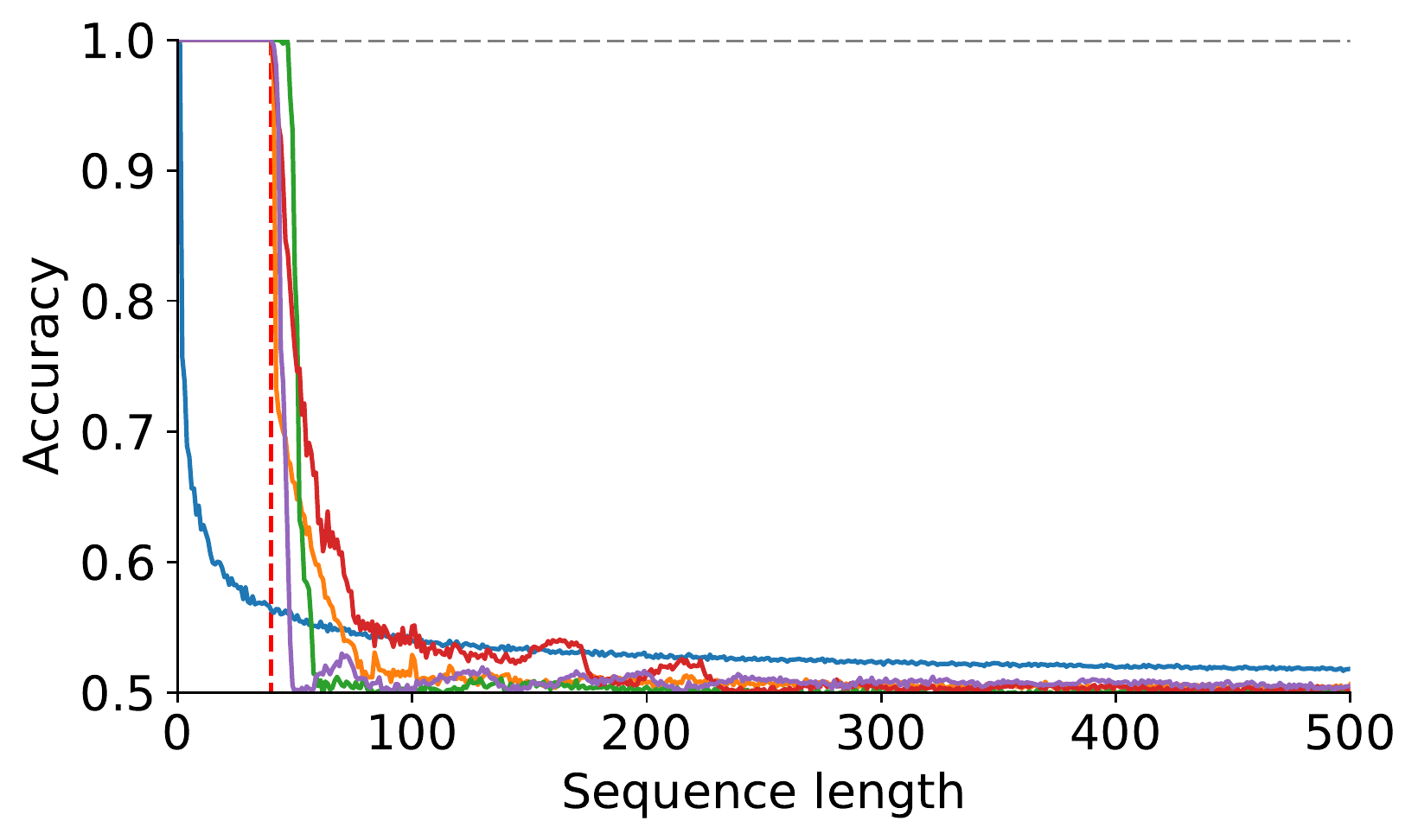}
            \end{center}
            \vspace{-0.25cm}
            \caption{\tiny{\oddsfirst{} (\gls*{cs})}}
        \end{subfigure}
        \hspace{0.01\textwidth}
        \begin{subfigure}[b]{0.31\textwidth}
            \begin{center}
                \includegraphics[width=\textwidth]{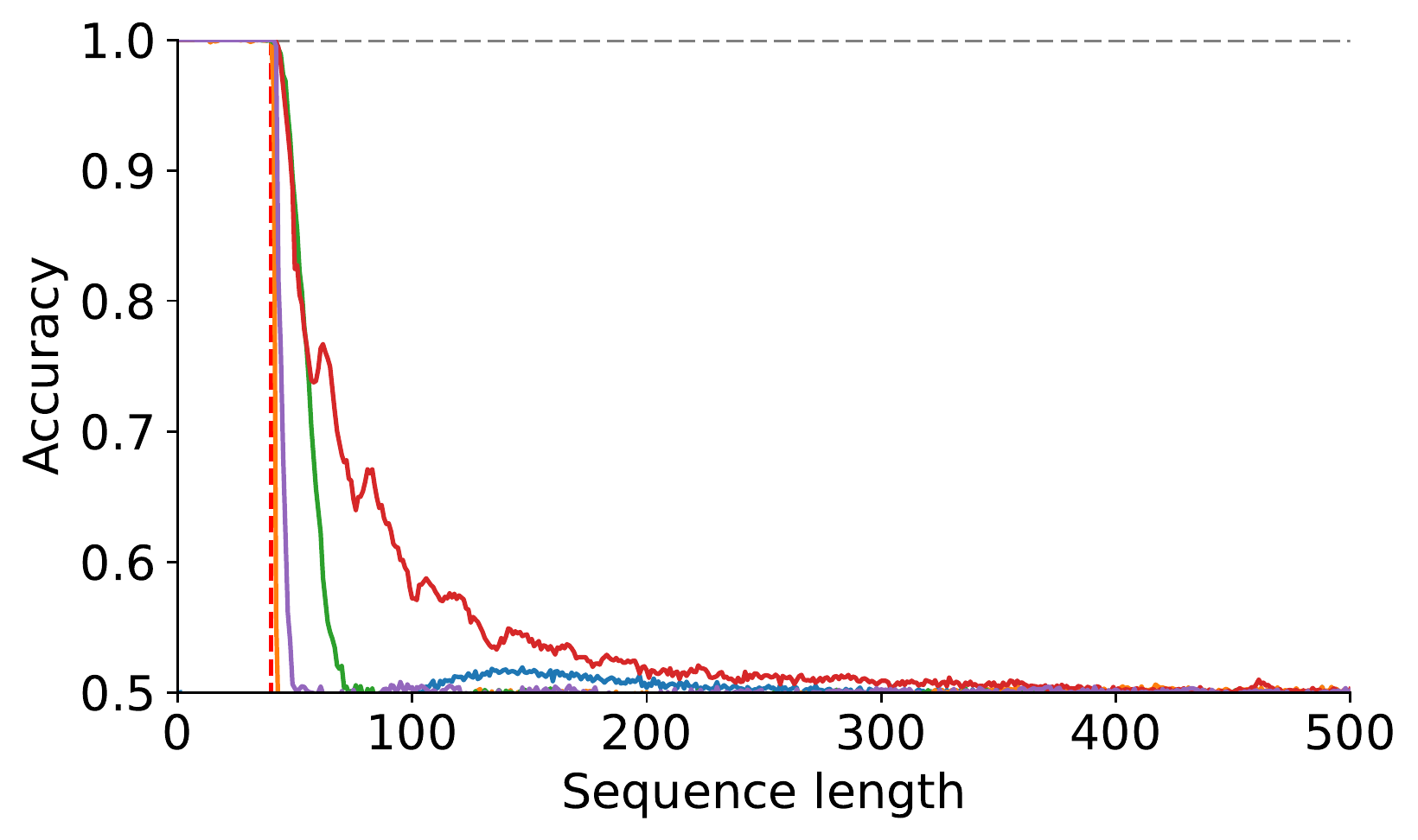}
            \end{center}
            \vspace{-0.25cm}
            \caption{\tiny{\binaryaddition{} (\gls*{cs})}}
        \end{subfigure}
        \vspace{0.5cm}
        
        \begin{subfigure}[b]{0.31\textwidth}
            \begin{center}
                \includegraphics[width=\textwidth]{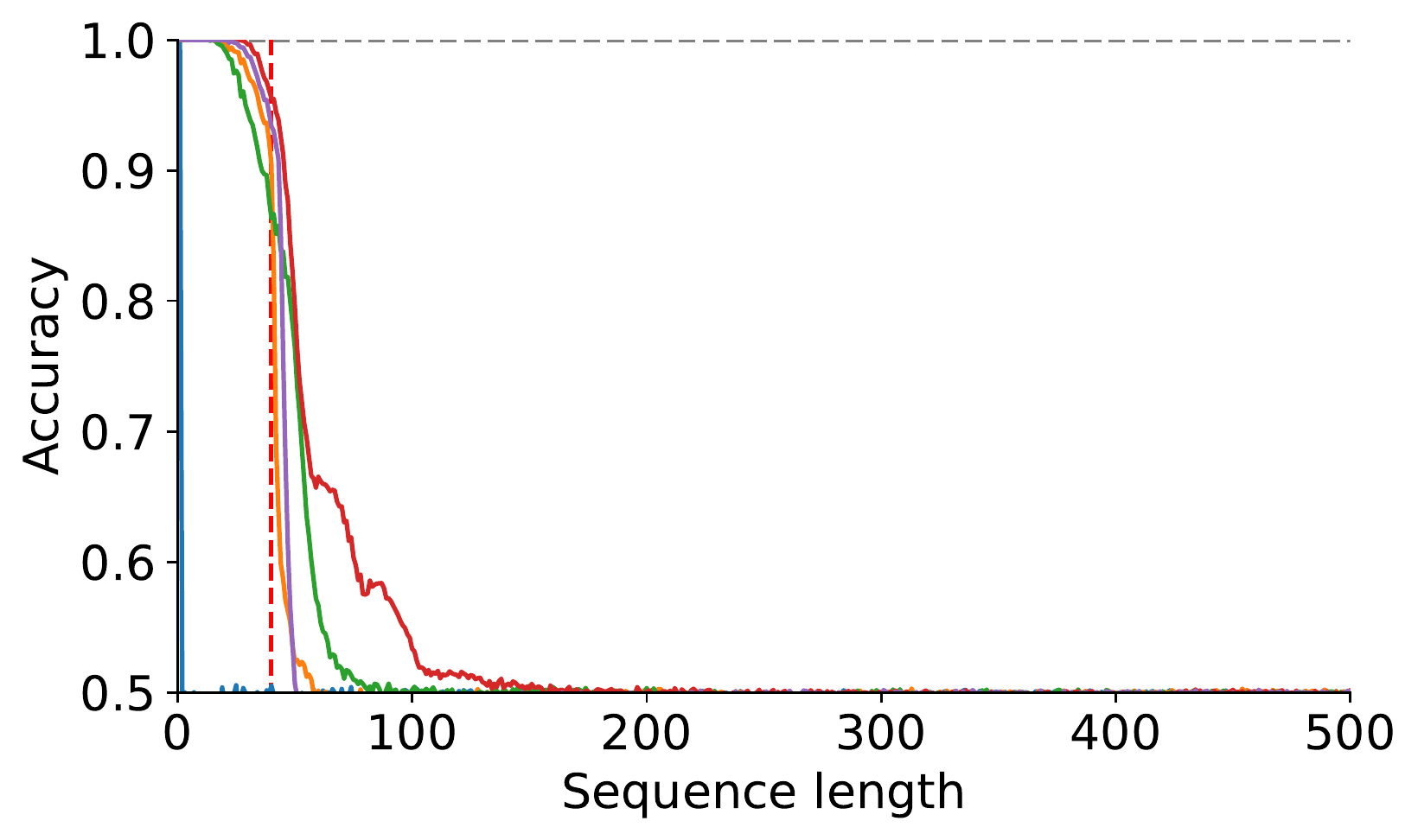}
            \end{center}
            \vspace{-0.25cm}
            \caption{\tiny{\binarymultiplication{} (\gls*{cs})}}
        \end{subfigure}
        \hspace{0.01\textwidth}
        \begin{subfigure}[b]{0.31\textwidth}
            \begin{center}
                \includegraphics[width=\textwidth]{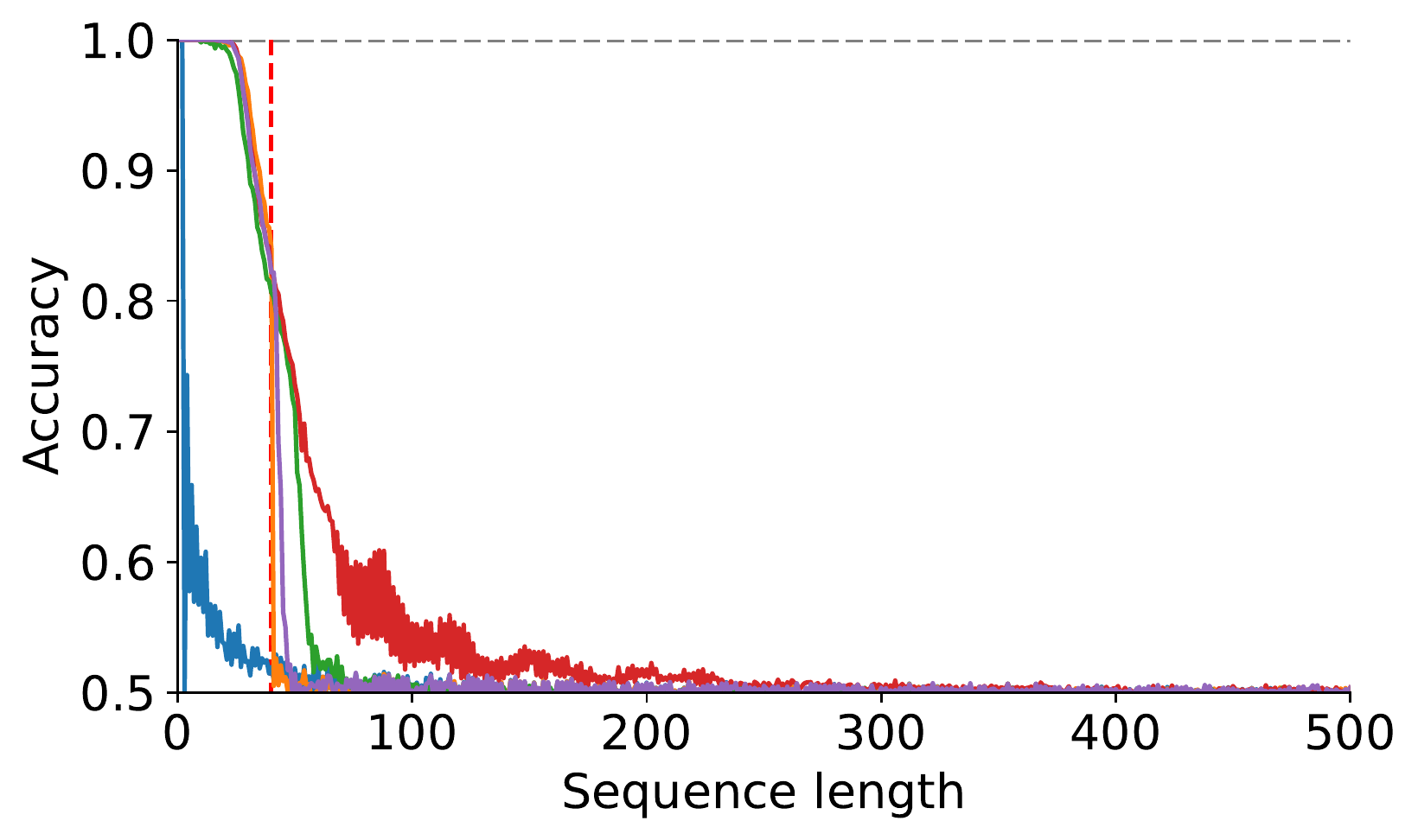}
            \end{center}
            \vspace{-0.25cm}
            \caption{\tiny{\computesqrt{} (\gls*{cs})}}
        \end{subfigure}
        \hspace{0.01\textwidth}
        \begin{subfigure}[b]{0.31\textwidth}
            \begin{center}
                \includegraphics[width=\textwidth]{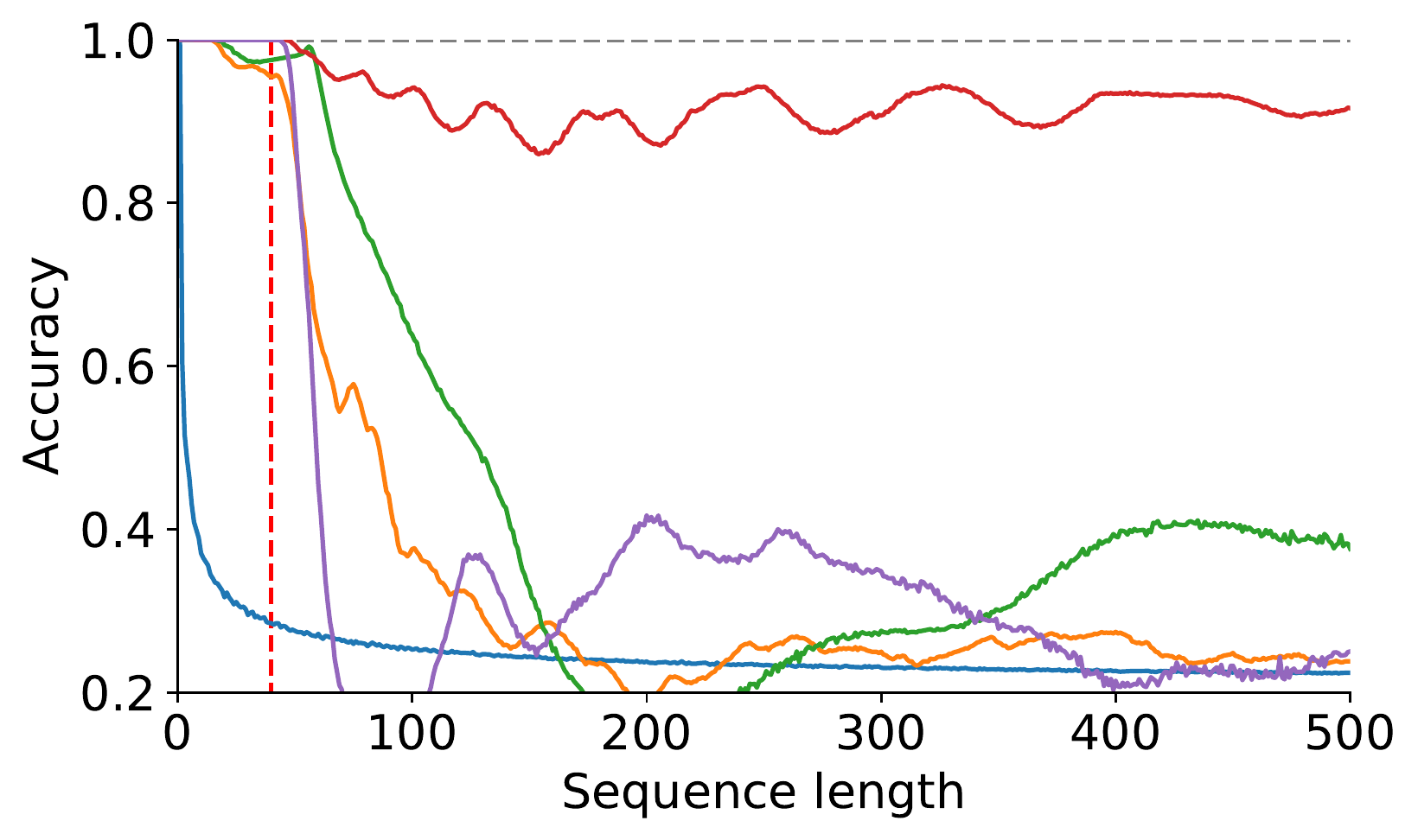}
            \end{center}
            \vspace{-0.25cm}
            \caption{\tiny{\bucketsort{} (\gls*{cs})}}
        \end{subfigure}
        \hspace{0.01\textwidth}
    \end{center}
    \caption{
        Performance curves on all tasks for the Transformer encoder architecture, for all the positional encodings we used.
        The dashed vertical red line is the training range, meaning that sequences to the right have not been seen during training and thus measure generalization.
    }
    \label{fig:scores-transformer-encodings}
\end{figure}

\subsection{Scaling laws and the Chomsky hierarchy}
\label{sec:additional-results:scaling-laws}

Suppose a neural architecture, \eg a transformer, is limited to a particular algorithmic complexity class (level on the Chomsky hierarchy).
In that case, no amount of scaling of the training time, training data, or the number of model parameters can overcome this limitation, thus putting a hard limit on what is possible via scaling.
Specific neural architectures might simply be unable to implement certain classes of algorithms in practice (\ie when considering the training protocol).
While we cannot rule out that excessive training, hyper-parameter tuning, or some very delayed grokking effect~\citep{power2022grokking} would lead to the architecture eventually solving the task, we can show that at least under standard training conditions, we can surface some systematic limitations that are not overcome with more training data or larger models but can be overcome with architectural extensions in line with the theory of computation (\ie adding a stack or a tape to an RNN).

To provide a more detailed empirical result, we train the transformer encoder with relative positional encodings for increasing numbers of in-distribution data points (\num{100000}, \num{1000000}, and \num{10000000} training steps) and increasing numbers of layers (5, 10, 15, 20, 25).
\Cref{tab:scaling-laws} shows that the score, \ie the accuracy averaged over all (unseen) test lengths and maximized over 10 seeds and 3 learning rates, generally does not increase with more training data or more parameters. Thus, in this empirical setup, the Chomsky hierarchy implies a hard limit on scaling laws~\citep{kaplan2020scaling}.

\begin{table}
  \caption{
    Score (see the caption of \cref{tab:scores}) for the Transformer with relative positional encodings and different numbers of layers and training steps (\ie training data).
    We observe that the Chomsky hierarchy implies a hard limit on scaling laws~\citep{kaplan2020scaling}, \ie more in-distribution data and/or more parameters do not lead to better length generalization performance.
  }
  \label{tab:scaling-laws}
  \begin{center}
    \resizebox{\columnwidth}{!}{\begin{tabular}{@{}lllrrrrr@{}}
\toprule
        &       &                   & \multicolumn{5}{c}{Layers} \\
\cmidrule{4-8}
Level   & Task  & Training Steps    & 5 & 10 & 15 & 20 & 25 \\
\midrule
\multirow{13}{*}{\Gls*{regular}}
& \multirow{3}{*}{\evenpairs}                         & \num{100000}      &     \textbf{90.2} & 84.7 & 74.8 & 50.1 & 50.1 \\
&                                                     & \num{1000000}     &     \textbf{90.0} & 78.6 & 78.7 & 50.1 & 50.1 \\
&                                                     & \num{10000000}    &     \textbf{97.5} & 89.7 & 89.5 & 50.1 & 50.1 \\
\cmidrule{3-8}
& \multirow{3}{*}{\modulararithmeticsimple}           & \num{100000}      &     21.7 & 21.6 & 21.5 & 20.5 & 19.9 \\
&                                                     & \num{1000000}     &     21.2 & 21.6 & 22.6 & 22.1 & 20.0 \\
&                                                     & \num{10000000}    &     21.3 & 22.1 & 24.8 & 24.4 & 20.0 \\
\cmidrule{3-8}
& \multirow{3}{*}{\paritycheck$^\dagger$}             & \num{100000}      &     50.3 & 50.4 & 50.1 & 50.1 & 49.9 \\
&                                                     & \num{1000000}     &     51.5 & 50.3 & 50.1 & 50.1 & 50.1 \\
&                                                     & \num{10000000}    &     52.0 & 50.5 & 50.1 & 50.1 & 50.1 \\
\cmidrule{3-8}
& \multirow{3}{*}{\cyclenavigation$^\dagger$}         & \num{100000}      &     24.5 & 25.4 & 23.7 & 22.9 & 23.2 \\
&                                                     & \num{1000000}     &     22.9 & 22.5 & 23.6 & 22.0 & 23.7 \\
&                                                     & \num{10000000}    &     22.8 & 22.9 & 25.9 & 23.2 & 23.3 \\
\midrule
\multirow{13}{*}{\Gls*{dcf}}
& \multirow{3}{*}{\stackmanipulation}               & \num{100000}      &     54.9 & 55.4 & 55.3 & 54.5 & 54.9 \\
&                                                   & \num{1000000}     &     54.0 & 54.4 & 53.7 & 54.9 & 56.2 \\
&                                                   & \num{10000000}    &     52.8 & 51.9 & 53.0 & 52.9 & 49.6 \\
\cmidrule{3-8}
& \multirow{3}{*}{\reversestring}                   & \num{100000}      &     61.6 & 61.2 & 58.0 & 56.1 & 56.8 \\
&                                                   & \num{1000000}     &     59.8 & 60.6 & 61.0 & 60.1 & 58.8 \\
&                                                   & \num{10000000}    &     56.5 & 58.7 & 57.2 & 52.1 & 50.0 \\
\cmidrule{3-8}
& \multirow{3}{*}{\modulararithmeticbrackets}       & \num{100000}      &     31.4 & 31.7 & 31.9 & 31.5 & 31.6 \\
&                                                   & \num{1000000}     &     28.9 & 27.9 & 31.0 & 31.0 & 31.0 \\
&                                                   & \num{10000000}    &     27.5 & 30.8 & 31.0 & 31.0 & 31.0 \\
\cmidrule{3-8}
& \multirow{3}{*}{\solveequation$^\circ$}           & \num{100000}      &     22.4 & 21.3 & 21.5 & 22.2 & 19.9 \\

&                                                   & \num{1000000}     &     23.0 & 22.4 & 21.6 & 21.8 & 19.9 \\
&                                                   & \num{10000000}    &     22.8 & 21.6 & 21.9 & 23.7 & 20.1 \\
\midrule
\multirow{24}{*}{\Gls*{cs}}
& \multirow{3}{*}{\duplicatestring}                 & \num{100000}      &     52.5 & 53.6 & 53.1 & 55.3 & 53.6 \\
&                                                   & \num{1000000}     &     51.9 & 51.5 & 53.1 & 53.7 & 54.2 \\
&                                                   & \num{10000000}    &     51.9 & 51.1 & 51.7 & 51.3 & 50.0 \\
\cmidrule{3-8}
& \multirow{3}{*}{\missingduplicate}                & \num{100000}      &     53.9 & 52.7 & 52.6 & 52.1 & 50.0 \\
&                                                   & \num{1000000}     &     53.2 & 53.0 & 50.0 & 50.0 & 50.0 \\
&                                                   & \num{10000000}    &     52.1 & 52.5 & 50.0 & 50.0 & 50.0 \\
\cmidrule{3-8}
& \multirow{3}{*}{\oddsfirst}                       & \num{100000}      &     52.5 & 53.4 & 52.9 & 52.2 & 53.2 \\
&                                                   & \num{1000000}     &     52.5 & 53.4 & 52.9 & 52.7 & 53.5 \\
&                                                   & \num{10000000}    &     51.8 & 52.8 & 52.0 & 50.0 & 50.0 \\
\cmidrule{3-8}
& \multirow{3}{*}{\binaryaddition}                  & \num{100000}      &     52.3 & 54.7 & 51.7 & 51.2 & 50.8 \\
&                                                   & \num{1000000}     &     53.3 & 55.9 & 54.8 & 54.8 & 53.8 \\
&                                                   & \num{10000000}    &     52.9 & 52.9 & 51.8 & 54.4 & 51.1 \\
\cmidrule{3-8}
& \multirow{3}{*}{\binarymultiplication$^\times$}   & \num{100000}      &     52.3 & 52.3 & 51.9 & 51.2 & 51.0 \\
&                                                   & \num{1000000}     &     52.5 & 53.1 & 53.4 & 53.4 & 54.0 \\
&                                                   & \num{10000000}    &     51.8 & 53.0 & 52.0 & 52.8 & 50.6 \\
\cmidrule{3-8}
& \multirow{3}{*}{\computesqrt}                     & \num{100000}      &     52.8 & 52.7 & 52.6 & 52.0 & 52.7 \\
&                                                   & \num{1000000}     &     52.3 & 53.4 & 53.7 & 54.0 & 53.5 \\
&                                                   & \num{10000000}    &     51.6 & 52.7 & 53.4 & 53.3 & 53.0 \\
\cmidrule{3-8}
& \multirow{3}{*}{\bucketsort$^{\dagger\bigstar}$}  & \num{100000}      &     \textbf{90.9} & \textbf{93.9} & \textbf{93.0} & \textbf{91.4} & 87.5 \\
&                                                   & \num{1000000}     &     \textbf{93.1} & \textbf{91.6} & \textbf{92.1} & \textbf{93.7} & \textbf{93.0} \\
&                                                   & \num{10000000}    &     84.3 & 81.9 & 80.4 & 89.5 & 65.1 \\
\bottomrule
\end{tabular}
}
  \end{center}
\end{table}

\end{document}